\documentclass[preprint,review,10pt]{elsarticle}

\pdfoutput=1
\usepackage{amssymb}
\usepackage{tikz}
\usepackage{colortbl}
\usepackage{fancyhdr}
\usepackage{atbegshi}
\usepackage[greek, british]{babel}
\usepackage{alphabeta}
\usepackage{libertine}
\usepackage{csquotes}
\usepackage{hyperref}
\usepackage{float}
\usepackage{amsmath}
\graphicspath{ {./images/} }
\usepackage[a4paper,margin=1in,footskip=0.25in]{geometry}
\usepackage{etoolbox}
\usepackage{svg}
\usepackage{threeparttable}

\makeatletter
\def\ps@pprintTitle{%
    \let\@oddhead\@empty
    \let\@evenhead\@empty
    \def\@oddfoot{\footnotesize\itshape
         {© 2024. This manuscript version is made available under the \href{https://creativecommons.org/licenses/by-nc-nd/4.0}{CC-BY-NC-ND 4.0 license}. DOI: 10.1016/j.engappai.2024.108231.\hfill}}%
    \let\@evenfoot\@oddfoot
    }
\makeatother

\newcommand{\xmark}{%
\tikz[scale=0.23] {
    \draw[line width=0.7,line cap=round] (0,0) to [bend left=6] (1,1);
    \draw[line width=0.7,line cap=round] (0.2,0.95) to [bend right=3] (0.8,0.05);
}}

\journal{Engineering Applications of Artificial Intelligence}

\fancypagestyle{customheader}{%
  \fancyhf{}% Clear header/footer
  \fancyhead[C]{Article published by Elsevier's Engineering Applications of Artificial Intelligence}% Customize this line with your custom header content
}

\AtBeginShipout{\thispagestyle{customheader}}

\begin{document}

\begin{frontmatter}

\title{Neural Natural Language Processing for Long Texts: A Survey on Classification and Summarization}

\author{Dimitrios Tsirmpas\textsuperscript{$\dagger$\textsection}, Ioannis Gkionis\textsuperscript{$\dagger$\textsection}, Georgios Th. Papadopoulos\textsuperscript{\textparagraph}, Ioannis Mademlis\textsuperscript{*}\textsuperscript{\textsection}\textsuperscript{\textparagraph}}
\address{\textsuperscript{\textsection} Department of Informatics, Athens University of Economics and Business}
\address{\textsuperscript{\textparagraph} Department of Informatics and Telematics, Harokopio University of Athens}
\fntext[myfootnote1]{\textsuperscript{$\dagger$} The first two authors contributed equally and are joint first authors.}
\fntext[myfootnote2]{\textsuperscript{*} Ioannis Mademlis is the corresponding author.}

\begin{abstract}
The adoption of Deep Neural Networks (DNNs) has greatly benefited Natural Language Processing (NLP) during the past decade. However, the demands of long document analysis are quite different from those of shorter texts, while the ever increasing size of documents uploaded online renders automated understanding of lengthy texts a critical issue. Relevant applications include automated Web mining, legal document review, medical records analysis, financial reports analysis, contract management, environmental impact assessment, news aggregation, etc. Despite the relatively recent development of efficient algorithms for analyzing long documents, practical tools in this field are currently flourishing. This article serves as an entry point into this dynamic domain and aims to achieve two objectives. First of all, it provides an introductory overview of the relevant neural building blocks, serving as a concise tutorial for the field. Secondly, it offers a brief examination of the current state-of-the-art in two key long document analysis tasks: document classification and document summarization. Sentiment analysis for long texts is also covered, since it is typically treated as a particular case of document classification. Consequently, this article presents an introductory exploration of document-level analysis, addressing the primary challenges, concerns, and existing solutions. Finally, it offers a concise definition of ``long text/document", presents an original overarching taxonomy of common deep neural methods for long document analysis and lists publicly available annotated datasets that can facilitate further research in this area.
\end{abstract}

%%Graphical abstract
%\begin{graphicalabstract}
%\includegraphics{grabs}
%\end{graphicalabstract}

%%Research highlights
% \begin{highlights}
% \item Research highlight 1
% \item Research highlight 2
% \end{highlights}


\end{frontmatter}

\section{Introduction}
Understanding written text has always drawn significant interest within the Artificial Intelligence (AI) community. Nowadays, it also enjoys increasingly many commercial applications: successfully parsing and analyzing texts expressed in natural language is crucial for a variety of practical tasks traditionally performed by humans, which require the extraction of sentiments, meaning or themes. Books \cite{worsham, worsham_book, brazil, jose}, academic papers \cite{gales}, technical manuals \cite{nabizadeh}, news articles \cite{shuo}, legal documents \cite{lulu, dale, merchant} and many other types of long texts can be the target of such analysis. Common application domains for existing, real-world practical systems range from the automated processing of legal, medical, scientific and journalistic documents, to more niche areas such as intelligence gathering \cite{mademlis2023invisible}.

% Natural Language Processing (NLP) is the field of AI dedicated to developing algorithms for the semantic understanding of written and spoken language. NLP methods can be differentiated by the level of granularity they operate on. \textit{Sentence-level} NLP examines individual sentences and their structure, grammar, and meaning. This type of analysis is useful for tasks such as sentiment analysis or named entity recognition, which can be performed on a sentence-by-sentence basis \cite{borgir}. \textit{Paragraph-level} NLP analysis involves examining the larger context in which sentences are used. This type of analysis can help identify the topic or theme of a paragraph, as well as the relationships between sentences within the paragraph, and is usually used as an intermediate stage between sentence-level and document-level analysis \cite{andrew} \cite{guo} \cite{jiwei}. \textit{Document-level} NLP analysis involves analyzing an entire document, such as a book, article, or email. This type of analysis can provide insights into the overall content, sentiment, and style of the document \cite{wei}\cite{timothy}. Document-level analysis can also involve tasks such as document classification or summarization, which require understanding the content of the entire document. 

Similarly to the more common case of short text analysis, long document Natural Language Processing (NLP) has been revolutionized during the past decade by Deep Neural Networks (DNNs), which greatly surpassed traditional statistical and machine learning approaches in accuracy and abilities. Nevertheless, even DNNs face severe challenges when analyzing long documents, due to a higher chance of ambiguities, varying context, potentially lower coherence, etc. \cite{hold}. Despite such limitations though, long document NLP has already become very important in the industry. It can be used to extract medical categories from electronic health records, enabling better patient care and treatment plans \cite{vithya}, or for automatically classifying lengthy legal documents \cite{lulu} \cite{bambroo}. In fact, the legal domain is currently one of the major application areas of long document NLP, with relevant algorithms exploited for desk research, electronic discovery, contract review, document automation, and even legal advice \cite{dale}.

This survey focuses on two key NLP tasks that present peculiarities for the long document case: \textit{document classification} and \textit{document summarization}. The first one involves categorizing entire documents into predefined classes, based on their content. This enables efficient organization and retrieval of information. The second one aims to generate concise and coherent summaries of longer documents. It involves distilling the key information and main ideas from a document, while preserving its meaning. Additionally, \textit{sentiment analysis} is also investigated as a particular variant of document classification. The task consists in determining the sentiment or emotional tone expressed in a piece of text, such as positive, negative, or neutral. It involves analyzing subjective opinions, attitudes, and emotions expressed in reviews, social media posts, or customer feedback.

Although other important tasks are also relevant to long documents (e.g., question answering, machine translation, information retrieval, topic modelling, etc.), the selected three tasks encompass a wide range of document analysis needs, play fundamental roles in handling large volumes of textual data and have direct applicability to diverse industries/domains, while they have been extensively studied and complement each other in terms of practical use. Overall, they jointly provide a balance between technical feasibility, practical applicability and user-centric impact, thus serving as suitable demonstrators for state-of-the-art (SoA) neural architectures designed for long document analysis.

% Before the rise of DNNs, traditional machine learning methods were typically used for executing NLP tasks. For instance, Latent Dirichlet Allocation (LDA) \cite{dirichet} is a popular unsupervised learning method that categorizes documents into $k$ topics, according to their \textit{tokens}. Tokens in this case would be individual words, but for other algorithms may be sentences, expressions, or subwords \cite{tokens}. The concept of tokens serves as a building block in defining more complex textual structures, such as \textit{n-grams} \cite{ngrams}. These are continuous sequences of $n$ tokens, which can be exploited by NLP algorithms to capture the context and/or the sequential dependencies within text. For a comprehensive coverage of relevant traditional approaches, the reader is referred to \cite{korde}.

Despite the presence of review papers discussing the development, methodology and applications of NLP tasks in general \cite{kowasari} \cite{torfi} \cite{chai}, or in specific thematic fields \cite{nielker} \cite{qian} \cite{gupta} \cite{hussein}, there are currently no survey/review articles discussing and aggregating recent DNN solutions for long document analysis across tasks. To remedy such gaps in existing literature, this article specifically focuses on document-level analysis for the three selected NLP tasks (classification, summarization, sentiment analysis). Unlike other review papers, particular emphasis is given to how the relevant neural building blocks and architectures operate, equipping this article with an additional tutorial value. From the survey point-of-view, existing solutions to the three NLP tasks under examination are systematically reviewed, while current challenges are subsequently identified and discussed along with common ways to circumvent them. The methods and algorithms that have been selected tackle the peculiarities of long documents; not text classification, summarization or sentiment analysis in general. Overall, the goal of this article is two-fold: i) to make the barrier of entry for this relatively young section of active research more accessible, and ii) to aggregate common issues and solutions across multiple document-level long text NLP tasks, thus encouraging cross-pollination of research and ideas.

The remainder of this article is organized in the following manner. Section \ref{sec::DNNs} is a brief tutorial of basic neural architectures and building blocks that are essential for modern document analysis. Section \ref{sec::LongDocumentAnalysis} offers a definition of long texts/documents, along with a list of relevant challenges, a timeline and a taxonomy of relevant deep neural methods, irrespectively of any specific task. Sections \ref{sec::Classification}, \ref{sec::Summarization} and \ref{sec::Sentiment} detail the challenges and proposed solutions to the document classification, summarization and sentiment analysis tasks, respectively, emphasizing methods designed specifically for long documents. To better focus this overview, these sections only examine dedicated algorithms and pretraining/finetuning approaches. Zero/few-shot prompting alternatives based on generic pretrained Large Language Models (LLMs) are not covered, since they fall outside the scope of the article. Section \ref{sec::Datasets} presents publicly available, annotated long document datasets, which can be utilized for relevant research. Section \ref{sec::Conclusions} summarizes and organizes the presented methods and challenges, discusses current trends and identifies open issues. Additionally, it draws conclusions from the preceding discussion, regarding the current state and future of long document analysis. The Appendix details previous recent surveys/reviews that overlap with this article, highlighting the main differences.
    
\section{Deep Neural Networks Used in Document Analysis}
\label{sec::DNNs}
NLP tasks place specific demands on DNNs: for instance, a DNN model needs to uncover on its own the structure, context and the meaning of a single word. This calls for specialized architectures whose properties allow the DNN to navigate through the complexities of human-generated natural text. Long documents exacerbate these challenges, since: i) the context, the tone and the theme may change repeatedly as the text progresses, and ii) the DNN may be required to correlate semantic cues which are far apart to each other within the text, in order to successfully execute the desired task.

Since the SoA in NLP has been revolving around DNNs for the last several years, this section briefly presents the most widespread and effective neural architectures used for NLP. The complexities and the individual variations of each architecture are out of the scope of this article, since its focus is on DNN-enabled document analysis and not on DNNs themselves. All architectures presented in this section are typically trained with variants of error back-propagation and stochastic gradient descent \cite{rumelhart} \cite{jordan}, but only the inference stage is described here.

The following presentation is split into two subsections. Initially, generic neural architectures common in the relevant SoA are briefly described for reference purposes. Subsequently, specific NLP-oriented neural architectures commonly used for analyzing texts are detailed.

\subsection{General Neural Architectures}

\subsubsection{Encoder-Decoder Architectures and the Attention Mechanism}
Neural NLP initially utilized traditional DNN architectures, such as MultiLayer Perceptrons (MLPs) \cite{rumelhart}, Convolutional Neural Networks (CNNs) \cite{lenet}\cite{alexnet} or Recurrent Neural Networks \cite{siegelman}, such as Long Short-Term Memory networks (LSTMs) \cite{sepp} or Gated Recurrent Units (GRUs) \cite{Cho2014}, with RNNs having the ability to analyze sequences by unfolding through time across discrete time steps. In NLP, typically each time step corresponds to an input token. However, these architectures face limitations: e.g., CNNs have computational overhead issues and RNNs/LSTMs struggle with long-term dependencies in long texts. The vast majority of contemporary DNNs for document analysis rely on the more recent Transformer architecture, which itself is built around the so-called \textit{neural attention mechanism}.

Modern neural attention emerged out of a specific setup of using RNNs, namely Encoder-Decoder. Concretely, any RNN/LSTM of one or more layers which unfolds across $K$ time steps accepts a sequence of $K$ input vectors in order to finally produce a corresponding sequence of $K$ output vectors. However, in a wide range of practical tasks the temporal lengths of the input/output sequences are not identical. If the length of one is $K$ while the other's is 1, the problem is trivially solved by unfolding the network across $K$ time steps and considering only the final activation of the $K$-th time step as the network prediction. However, \textit{sequence-to-sequence} mapping tasks between sequences of different lengths other than 1 are more complicated.

Typically, RNNs are adapted to solve such problems (e.g., machine translation) using a mixed Encoder-Decoder scheme. The input sequence is fed to a recurrent coding network which processes it over $K$ time instants (where $K$ is, e.g., the length of the input sequence), but only the final network output from the $K$-th time step is stored. Thus, it constitutes a coded vector (of fixed dimensionality equal to the number of the Encoder's output neurons, independent of $K$) representing the entire input sequence. This vector is then fed to a recurrent decoding network unfolding across $L$ time steps ($L$ can be arbitrary and not necessarily equal to $K$ or 1). At the first time step of the Decoder's inference stage it is given as input the coded vector. During the following time steps, its processing is based only on the stored internal state. The $L$ Decoder outputs constitute the final output sequence. The overall network is trained uniformly by BPTT.

As an improvement over this basic idea, the neural attention mechanism was introduced in \cite{bahdanau} to allow the Decoder of a mixed Encoder-Decoder RNN architecture to consider all the outputs of the Encoder, and not just that of the last encoding time step, with a different weighting factor. These weighting coefficients, known as \textit{attention weights}, are given by an attention distribution that the Decoder produces at each time step of its inference stage, in order to read selectively and at will the Encoder's outputs according to its own current internal state. This is achieved by the attention mechanism: a differentiable and continuous way to access a sequence of stored discrete \textit{key vectors}. Thus, the mechanism allows a neural network to learn suitable access patterns on its own from proper training on a dataset, through usual back-propagation and gradient descent optimization.

Attention operates by generating a continuous probability distribution defined over the entries of the key sequence. This distribution is produced in the following way. First, the network generates a \textit{query vector} that essentially needs to be searched for in the key sequence. Then, the dot product of that query with each key vector is separately computed, with the ordered results essentially forming a new sequence of cosine similarities between the query and each key vector. This similarity sequence is converted to a probability distribution by passing it through a softmax function. Then, a weighted sum of all key vectors is generated, using the attention weights as coefficients. Optionally, this weighted sum can be computed on a different set of \textit{value vectors}, instead of the key vectors through which the attention weights were estimated. Queries, vectors and keys can be derived by the DNN during inference through projecting given vectors (linearly or non-linearly) via learned separate transformations. These operations are differentiable and their parameters are learned while training the overall network with a typical, task-specific cost function. The process of constructing the weighted vector sum using the attention weights is referred to as \textit{the query attending to the keys and/or values}.

In the case of Encoder-Decoder RNN architectures, the key/value sequence is the temporal sequence of Encoder internal states/outputs across $K$ time steps, respectively, while the $L$ separate queries are the Decoder's internal states at each time step of its inference stage. Alternatively, the key and the value sequence may coincide. This setup allows the Decoder to read at each time step an aggregate \textit{context vector} before it generates its current output: such a context is a parameterized summary of all $K$ Encoder outputs, adjusted according to the Decoder's own current state.

\subsubsection{Transformers}
\label{ssec::Transformers}

The Transformer architecture \cite{ilia} was an attempt to replace RNNs for sequence-to-sequence mapping tasks, while simultaneously maintaining the mixed Encoder-Decoder scheme. Both the Encoder and the Decoder consist of $N$ consecutive macrolayers each. The Decoder operates at discrete time steps, with its final layer outputting at the end of each step the next element of the requested output sequence. Its first layer's input at the beginning of each step is the output of its final layer from the previous step, i.e., the Decoder is \textit{autoregressive}. However, the Encoder operates at a single pass, processing the entire input sequence at once. The first Encoder layer receives as its input a list of at most $K$ vectors, i.e., an ordered list of all elements/tokens in the input sequence. Each vector in this sequence is the sum of a suitable representation of the corresponding token (e.g., a sparse one-hot encoding of a word from the supported vocabulary) and a "positional encoding", i.e., a dense vector representation of the index of the respective input token within the overall sequence. Each of the subsequent Encoder layers receives as its input a list of $K$ corresponding, transformed vectors, i.e., the outputs of the previous Encoder layer.

Each layer of the Encoder or the Decoder is composed of two consecutively placed sublayers: a \textit{self-attention} sublayer and a succeeding small MLP. The self-attention mechanism enables the layer to process its input sequence in parallel, while the following fully connected layers transform the output of the self-attention process. However, a self-attention sublayer is itself composed of $M$ parallel, independent self-attention heads. Each head receives a linearly transformed version of each input vector as a query and two linearly transformed versions of the overall input sequence as a key and as a value sequence. Within a head, a common key/value sequence representation is utilized for all queries. The transformations are performed by multiplying each input vector with suitable weight matrices, separately learned per self-attention head, as model parameters. Thus, although all heads of one layer separately receive the same input sequence as a list of queries, of keys and of values, this sequence is differently transformed per head before being fed to them. The output of each self-attention head is one new vector representation per input token, which inherently incorporates appropriate context information from the entire sequence. The outputs of all heads are concatenated and subsequently linearly transformed, using an additional learned weights matrix, before being fed to the succeeding MLP.

\begin{figure}
    \centering
    \includesvg[width=15cm]{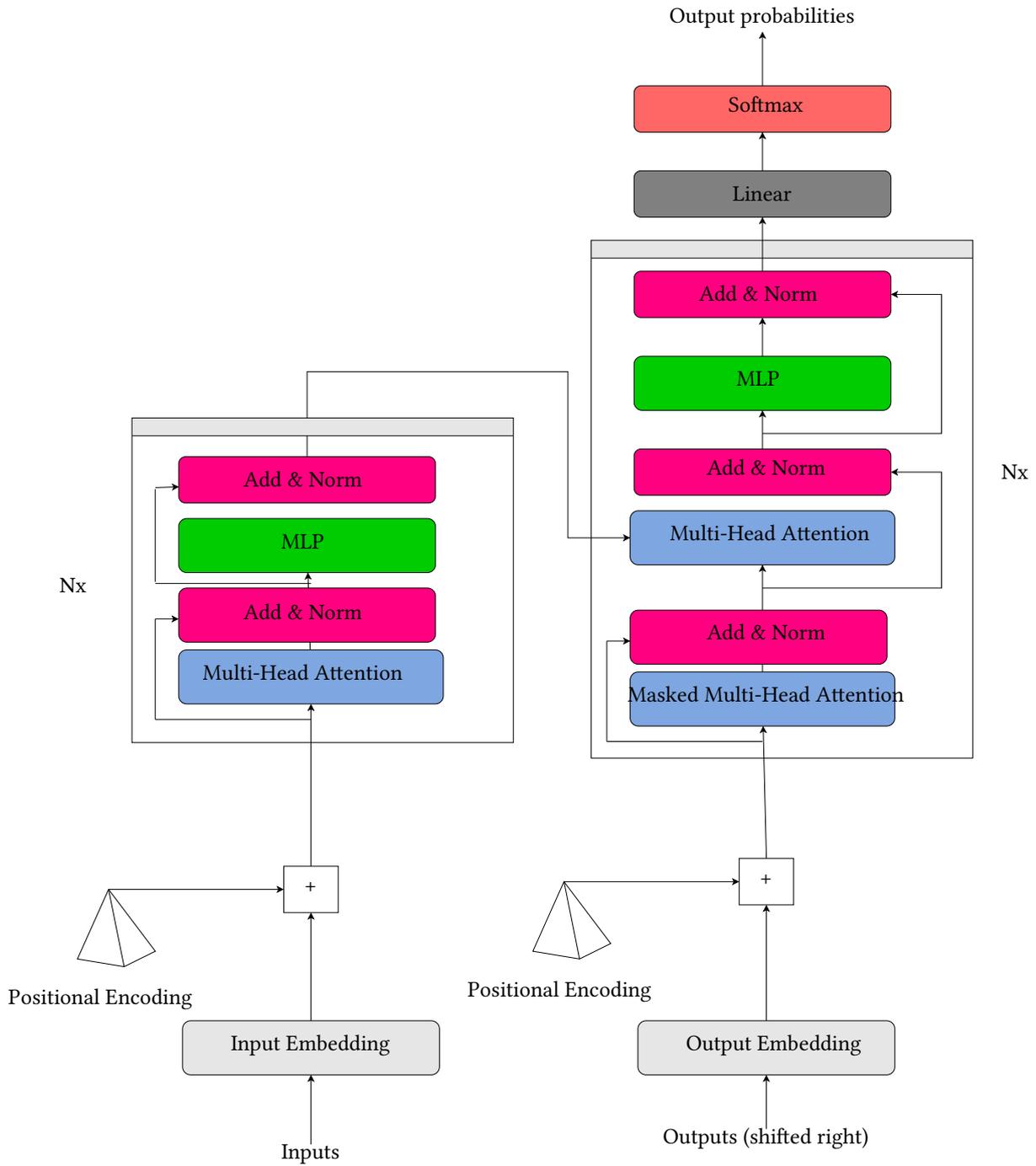}
    \caption{The original Transformer Encoder-Decoder architecture.}
    \label{fig::TransformerVanilla}
\end{figure}

The Decoder has a structure similar to the Encoder, but includes an additional, regular attention mechanism within each of its layers, for also attending to the Encoder's final outputs. However, the use of both an Encoder and a Decoder is not strictly necessary in tasks that do not involve sequence-to-sequence mapping with input/output sequences of different lengths. The original Transformer's Encoder-Decoder architecture is depicted in Figure \ref{fig::TransformerVanilla}.

Overall, the simultaneous processing from multiple, parallel self-attention heads and the resulting lack of recurrence in the Encoder render the Transformer able to achieve faster training and inference times, compared to previous RNNs/LSTMs that process the input sequentially. Additionally, the learnable self-attention mechanism allows the Transformer to easily, selectively and adaptively capture contextual information and long-term dependencies from the entire input sequence, when analyzing each individual token during inference. Thus, Transformer layers effectively have a global receptive field from the get-go. Finally, the use of multiple self-attention heads per layer allows each layer to simultaneously compute multiple different representations from its input sequence, thus capturing different features at once.

Eq. (\ref{eq::Att1}) below succinctly captures the inference-stage processing of a self-attention head:
\begin{equation}
\label{eq::Att1}
    Attention(\textbf{Q},\textbf{K},\textbf{V}) = softmax(\frac{\textbf{QK}^{T}}{\sqrt{d_k}})\textbf{V},
\end{equation}
\noindent where $\textbf{Q} \in \mathbb{R}^{N_q \times d_k}$ are the $N_q$ queries, $\textbf{K} \in \mathbb{R}^{N_k \times d_k}$ are the $N_k$ keys and $\textbf{V} \in \mathbb{R}^{N_k \times d_v}$ are the $N_k$ values. Each query vector and each key vector has a dimension of $d_k$, while each value vector has a dimension of $d_v$.

Multihead self-attention can be formulated as:
\begin{equation}
Multihead(\textbf{Q},\textbf{K},\textbf{V}) = [\textbf{h}_1;...;\textbf{h}_{M}]\textbf{W}^O,
\end{equation}
\noindent where
\begin{equation}
\textbf{h}_i = Attention(\textbf{Q}\textbf{W}^{Q}_i, \textbf{K}\textbf{W}^{K}_i, \textbf{V}\textbf{W}^{V}_i).
\end{equation}
\noindent In this formulation, $\textbf{W}^{Q}_i \in \mathbb{R}^{d_a \times d_k}$, $\textbf{W}^{K}_i \in \mathbb{R}^{d_a \times d_k}$, $\textbf{W}^{V}_i \in \mathbb{R}^{d_a \times d_v}$, $\textbf{W}^{O}_i \in \mathbb{R}^{M d_v \times d_a}$ are learned parameter matrices, $M$ is the number of heads, $d_k=d_v=\frac{d_a}{M}$, and the operator $[;]$ implies concatenation.

Note that the softmax function in Eq. (\ref{eq::Att1}) is separately applied to each row of its argument, resulting in a matrix output. Each row of each $\textbf{h}_i$ matrix is a convex combination of the rows of $\textbf{V}\textbf{W}^{V}_i$, with the respective row of the output of the softmax providing the corresponding weighting coefficients. Scaling by $\sqrt{d_k}$ in Eq. (\ref{eq::Att1}) serves to ensure training stability.

\subsection{Neural Building Blocks for Document Analysis}
\subsubsection{Word Embeddings}
Modern NLP algorithms typically require a very important preprocessing phase: during this phase, each input token is transformed into a semantically meaningful, fixed-size, dense vector, which is typically called \textit{word embedding}. The resulting sequence of word embeddings is the actual input to the main DNN that executes the desired NLP task \cite{mikolov}. Such word embeddings are utilized both at the training and at the inference stage. One trivial way to achieve this is to simply append an initial \textit{embedding layer} at the employed DNN architecture, which is subsequently trained end-to-end. Thus, the embedding layer receives the raw tokens as its input, e.g., as sparse, one-hot encoded vectors, and transforms them to a semantically meaningful, dense vector representation. However, the most widespread approach is to exploit pretrained, separate, dedicated \textit{embedding neural networks} and utilize their outputs as dense word embedding vectors.

One of the most historically significant word embedding neural networks was Google's \textit{Word2Vec} \cite{mikolov}. It is a shallow MLP with one hidden layer, able to compute a useful representation of a word token in the form of a real, dense, fixed-size vector with a dimensionality equal to $H$, i.e., the number of hidden neurons in the model. Word2Vec\cite{mikolov} representations/embeddings have rich semantic content: words with similar or related meaning are mapped to vectors that are approximately parallel in the representation/embedding space (i.e., they have high cosine similarity), while applying typical vector operations to such representations is semantically meaningful (e.g., corresponds to semantic analogies between the represented words) \cite{jurafsky}.

The Word2Vec MLP receives as its input a non-semantic (e.g., one-hot) vector encoding of a word with a dimensionality of $L$ (equal to the supported vocabulary size). It has been pretrained according to the following self-supervised objective. First, for each occurrence of each significant word (e.g., nouns, verbs, adjectives, proper nouns) in the available training dataset/text corpus, the $D$ words immediately before and the $D$ words immediately after the one in question are selected. Each such mapping from a word to its $2D$ neighboring ones in the context of each occurrence is exploited as a training pattern/label pair, using typical error back-propagation and gradient descent. Thus, the co-occurrence of words within a given context is implicitly utilized to learn the relationships between these words.

The input/hidden/output layer has $L$/$H$/$2DL$ neurons, respectively. After the model has been trained, the output layer is discarded and the remaining network can be exploited to produce a semantically rich representation of each input word: this is a dense, $H$-dimensional real vector. Obviously, a very large data set is required to properly train such a model, and in Word2Vec's case that was GoogleNews (100 billion words).

Other word embedding algorithms followed-up and improved upon the Word2Vec concept, such as GloVE \cite{glove}. However, the next significant milestone was the development of context-sensitive word embedding networks, which were able to generate different vector representations for a single word depending on its context, i.e., its position in a sentence. ELMo \cite{elmo} was able to achieve this using an LSTM architecture, but it was Bidirectional Encoder Representations from Transformers (BERT) \cite{toutanova} which originally exploited the power of Transformers in order to significantly advance the relevant SoA. Such DNNs that learn to generate contextualized word embeddings, by being trained to predict the next word of an input sentence, are essentially \textit{language models}.

\subsubsection{BERT}
BERT is a bidirectional, deep Transformer Encoder, without an attached Decoder. It receives an input sequence consisting of numerical representations of a text's tokens/words (e.g., non-semantic indices to a supported vocabulary), to generate a corresponding sequence of refined, semantically encoded output vectors as this text's representation. Initially, the special token [CLS] is externally appended to the beginning of the overall input sequence, so that its output semantic representation will aggregate global contextual information about the entire textual sequence. Also, the special token [SEP] is inserted after each input sentence to separate consecutive sentences. The ordered token representations are then transformed by a preliminary, trainable embedding layer. Subsequently, a learned dense ``segment embedding" vector and a learned position embedding vector\footnote{This differs from the original Transformer's statically defined positional encoding vector.} are added to the vector representation of each input token. Finally, the first actual Transformer layer receives these ordered vector sums as its input sequence. The segment embedding of a token belonging to the $i$-th sentence simply indicates whether $i$ is odd or even\footnote{The baseline BERT supports only 2 sentences, but each of these is a segment of consecutive textual content and not an actual individual sentence in the linguistic sense.}.

The output word embeddings have specialized semantic content in comparison to corresponding Word2Vec representations, i.e., adapted to the context of the specific input sentence. This is because the network processes a complete input text at each iteration of its training stage, taking into account all its words and their order at a single pass, through the self-attention mechanism. Thus, during inference, it processes each input word in the context of its bilaterally ordered phrasal contexts. Therefore, the same word in the context of different sentences, or even placed at a different position in an otherwise identical sentence, can be mapped to different embedding vectors by the pretrained network.

More difficult objectives are employed for pretraining BERT, compared to simply predicting the previous and next words of the current sentence, as in Word2Vec. These objectives are tailored to the Transformer's special features. The most common goal is to predict a few randomly chosen masked words of a complete input sentence based on its remaining words. This is a "Masked Language Model" (MLM) pretraining objective, inspired by the Cloze task \cite{taylor}, where a subset of the input's tokens are randomly masked, i.e., each one is replaced by a special [MASK] token, and the goal is to predict the original words based on the context. The MLM objective encourages the model to fuse the left and the right context and, thus, is particularly suited to deep bidirectional Transformers \cite{toutanova}. A complementary pretraining objective is to classify two jointly fed input sentences as either consecutive or non-consecutive ones (Next Sentence Prediction, NSP). This teaches BERT to understand longer-term dependencies across sentences. In both cases, after training the final prediction head or classification layers are discarded and only the trained Encoder is retained, to generate the word embeddings of input sentences.

\begin{figure}
    \centering
    \includesvg[width=16cm]{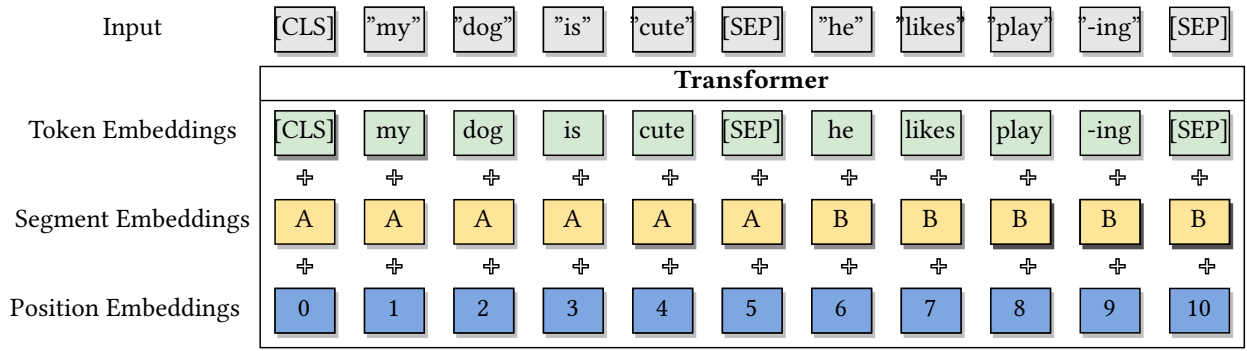}
    \caption{BERT input representation. The input embeddings are the sum of the token embeddings, the segmentation embeddings and the position embeddings.}
    \label{fig::BERTInput}
\end{figure}

A shallow, pretrained Word2Vec model encodes a fixed 1-1 mapping of words to representations/embedding vectors and, thus, can be used in a variety of NLP tasks only for feature extraction at a preprocessing stage. In contrast, a pretrained deep BERT model can optionally be adapted itself to the desired downstream NLP task, by appending appropriate additional final neural layers (usually fully connected or recurrent) and performing minimal fine-tuning on the overall network. During inference, BERT must be given as input a complete sentence, so as to generate the most contextually appropriate embeddings of its words. Due to the fixed maximum sequence size accepted by Transformers, BERT is restricted to receiving at most 512 tokens as its input text. Finally, a pretrained BERT comes with a dictionary of words it has been trained to embed. Thus, if during inference it is found out that the current word is unknown, it is broken into subwords which are considered by the network as different, consecutive elements of the input sequence (\textit{subword tokenization}). In the case of compound words, such a subword is indeed potentially known, otherwise the token decomposition eventually reaches the level of the individual alphanumeric characters of the original word, which always belong to the known dictionary. The final representation of an input word can be derived as the vector average of the embeddings of its subwords. Thus, BERT can easily handle "unsupported" out-of-vocabulary words. Figure \ref{fig::BERTInput} depicts how the input to BERT is constructed.

% Examples of BERT models include BERTBASE (L=12, H=768, A=12, parameters=110M) and BERTLARGE (L=24, H=1024, A=16, parameters=340M) where $L$ = number of layers, $H$ = word embedding dimensionality, $A^{3}$ = number of attention heads. 

\subsubsection{RoBERTa and ELECTRA}
Various improvements over the original BERT have been proposed over the past few years. For instance, "Robustly Optimized BERT pre-training Approach" (RoBERTa) \cite{roberta} aggregated several minor improvements in the training process and slightly modified the MLM pretraining objective, by dynamically changing the masking pattern applied to the training data.

\begin{figure}[H]
    \includesvg[width=12cm]{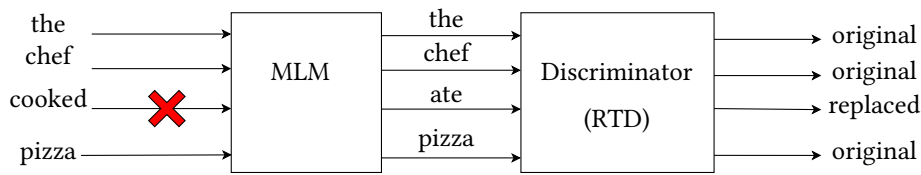}
    \centering
    \caption{Graphical comparison between the MLM and the RTD tasks.}
    \label{fig::MLM-RTDTasks}
\end{figure}

One of the most significant recent improvements over BERT is "Efficiently Learning an Encoder that Classifies Token Replacements Accurately" (ELECTRA) \cite{karta}, which was also motivated by the shortcomings of the MLM self-supervised pretraining objective. The latter "corrupts" the input sentence by replacing a small subset of its tokens with [MASK] and then training the model to reconstruct/predict the original ones. This requires a lot of training iterations/sample to be effective, since the task is essentially defined only over the masked tokens. Thus, ELECTRA proposed "Replaced Token Detection" (RTD) as an alternative self-supervised pretraining task. In RTD, the model learns to distinguish real input tokens from plausible but synthetically generated replacements: a randomly selected subset of the input tokens are replaced with synthetic alternatives (typically the output of a small language model), instead of being masked. The pretraining task is defined over all input tokens, since each of them has to be classified as original or synthetic. Learning from all input tokens leads to a learning process which is both much faster and more efficient; thus, ceteris paribus, ELECTRA achieves increased performance in downstream tasks due to better contextual word embeddings. Figure \ref{fig::MLM-RTDTasks} graphically compares the MLM and the RTD tasks.

\subsubsection{T5}
T5 \cite{t5} is a renowned, general-use Transformer architecture for NLP. The modus operandi of T5 is to transform each desired NLP task, either discrimnative or generative, into a sequence-to-sequence mapping problem (text-to-text). Thus, it follows the standard Encoder-Decoder architectural paradigm, a design choice that allows it to easily handle input and output sequences of different sizes. This is unlike BERT, which is a bidirectional Encoder-only Transformer. Similarly to other language models, T5 is first pretrained on a large-scale corpus with a self-supervised objective, before being finetuned on the desired, typically supervised downstream task. Pretraining enables the model to learn general language patterns and representations, while downstream finetuning specializes it to task-specific nuances. Both training modes fall under the text-to-text format, therefore a common maximum likelihood loss function is employed regardless of the task.

\begin{figure}[H]
    \includesvg[width=12cm]{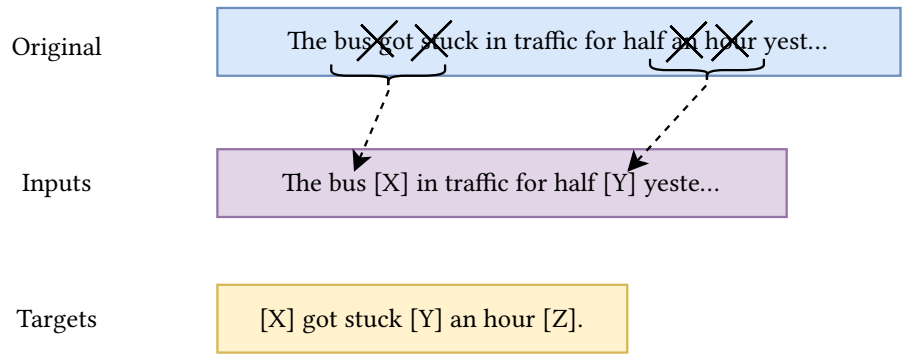}
    \centering
    \caption{The T5 pretraining objective.}
    \label{fig::T5objective}
\end{figure}

The text-to-text nature of the architecture allows T5 to be pretrained with a masked language modelling objective that uses targets composed of multiple consecutive words per each [MASK] token. Unlike BERT, this is done through random corruptions of text segments with varying masking ratios and segment sizes. A simplified example of the T5 self-supervised objective is depicted in Figure \ref{fig::T5objective}. The use of a complete Transformer Encoder-Decoder architecture, coupled with this fitting self-supervised pretraining task and a very large-scale pretraining dataset, jointly permit T5 to achieve remarkable performance across a variety of different language tasks (e.g., document summarization, sentiment analysis, etc.), after proper finetuning.

\subsubsection{BART}
Similarly to T5, BART \cite{BART} is also a Transformer language model that follows a full Encoder-Decoder architecture and is pretrained with a self-supervised sequence-to-sequence mapping objective. As in BERT, the Encoder is bidirectional and pretraining consists in mapping arbitrarily corrupted textual sentences to their original form. However, BART generalizes the BERT objective since a single [MASK] token can cover several consecutive target tokens (as in T5), while alternative corruptions are also possible. After pretraining, downstream finetuning can take place without any corruptions to the input. With this setup, BART achieves very good results in a variety of downstream language analysis tasks.

Notably, both T5 and BART utilize a full Transformer Encoder-Decoder architecture with an autoregressive Decoder and an inherent suitability to text-to-text mapping. An important consequence of this choice is that they can easily be finetuned for generative NLP tasks (text generation, e.g., for document summarization) without \textit{any} architectural additions after pretraining, while this is not possible with the bidirectional Encoder-only BERT. In this sense, BERT and its derivatives can be considered deprecated as of 2023 \cite{tay2022ul2}.

\subsubsection{GPT and Large Language Models}
Recently, NLP algorithms employing pretrained GPT (Generative Pretrained Transformer) \cite{GPT} neural models have been gaining a lot of traction. Similarly to other SoA language models, GPT is a large-scale autoregressive Decoder-only Transformer pretrained in a self-supervised manner and on a gigantic corpus for predicting the next token in a sequence. Due to this traditional causal language modelling objective, pretrained GPT models can be directly employed for text generation without any finetuning: they are given a text \textit{prompt} as input, and they generate a corresponding textual response as an output. Although pretrained GPT models can be finetuned in a typical supervised manner for a specific downstream task, their most impressive capability is that of zero/few-shot downstream language task execution during inference, through careful prompting and without any finetuning. In this case, the desired task is essentially described in natural language on-the-fly via the prompt. However, this setting falls outside the scope of this article.

When it comes to long document analysis, the latest openly available pretrained GPT model is GPT 2.0 \cite{GPT2}, which remains regularly outperformed by smaller language models, such as BART \cite{tanyangxing}, in the finetuning setting. However, the publicity GPT models have attracted due to their text generation capabilities makes them a serious competitor for BERT and its variants. GPT 3.0 achieves more impressive results \cite{tanyangxing}, but is not openly available to the public. GPT 4.0 is even more powerful, but similarly closed-source, like a number of its competitors such as \cite{chowdhery2022palm} and the recently announced Gemini \cite{gemini}. Open-source alternatives are BLOOM \cite{scao2022bloom} and OPT \cite{zhang2022opt}. The scale factor, i.e., highly complex Decoder-only Transformer DNN architectures and a gigantic pretraining corpus/dataset, has proven instrumental for obtaining good performance in all the above cases of Large Language Models (LLMs). However, the observation that scaling up the DNN architecture is only beneficial when accompanied by a corresponding increase in the dataset size has recently led to more reasonably sized models of this nature, i.e., less complex ones, such as Chinchilla \cite{chinchilla} and LLaMA \cite{Touvron2023llama}.

Table \ref{tab::ModelComplexity} showcases the escalation of language model complexity in recent years.

\begin{table}
    \centering
    \begin{threeparttable}
        \begin{tabular}{ |p{3cm}|p{3cm}|p{3cm}|p{3cm}| }
            \hline
            \cellcolor{blue!25}\textbf{Name} & \cellcolor{blue!25}\textbf{Year} & \cellcolor{blue!25}\textbf{\#Parameters} & \cellcolor{blue!25}\textbf{Input Size (\#tokens)} \\
            \hline
             BERT\cite{bert} & 2019 & 110M & 512 \\
             \hline
             BART\cite{BART} & 2019 & 140M & 1024 \\
             \hline
             T5\cite{t5} & 2019 & Up to 11B & 512 \\
             \hline
             GPT\cite{GPT} & 2018 & 120M &  - \\
             \hline
             GPT-2\cite{GPT2} & 2018 & 1.5B & 1024 \\
             \hline
             GPT-3\cite{GPT3} & 2020 & 175B & 4096\\
             \hline
             BLOOM\cite{scao2022bloom} & 2022 & 176B & - \\
             \hline
             OPT\cite{iyer2022opt} & 2022 & 125M to 175B & - \\
             \hline
             PALM\cite{chowdhery2022palm} & 2022 & 540B & 3072 \\
             \hline
             Chinchilla\cite{chinchilla} & 2022 & 70M to 16B & - \\
             \hline
             LLaMa\cite{Touvron2023llama} & 2023 & 7B to 65B & 2048 \\
             \hline
             GPT-4\cite{gpt4} & 2023 & Undisclosed \tnote{\textdagger} & 128,000 as of writing \\
             \hline
             Gemini\cite{gemini} & 2023 & 1.8B to undisclosed & 30720 \\
             \hline
        \end{tabular}
        
        \begin{tablenotes}
          \item[\textdagger]{{\small Rumoured at 1.76 trillion \cite{gpt4-speculation}. The original GPT-4 paper does not disclose any details about the size or architecture of the model.}}
        \end{tablenotes}
        
        \caption{Comparing modern language models by model complexity and input tokens limit (where available).}
        \label{tab::ModelComplexity}
    \end{threeparttable}
\end{table}

\section{Long Document Analysis}
\label{sec::LongDocumentAnalysis}
\subsection{Definition}
This article focuses exclusively on document-level analysis, or long text/long document classification and summarization. There is, however, currently no commonly-accepted definition of what constitutes a ``long text/document". Most distinctions in current literature focus on specific sub-genres of text (for example, algorithms considering only research papers or books as input). This has led to a fuzzy and unclear definition, where texts such as tweets, product and movie reviews are considered ``short documents", while research papers, legal documents and books are considered ``long documents" \cite{sumpubmed, poland, zichao}. Texts such as news reports, individual chapters of books and research paper abstracts are arbitrarily classified as both. An example of these qualitative definitions is presented in \cite{poland}, where the authors argue that short texts are often characterized by their brevity, their structured nature with explicit facts presented in a logical order and the absence of long-range causal relationships and dependencies, making the task much easier for an NLP model. This highlights the difficulty of defining long documents with just quantitative criteria, since even ``long" documents may actually be easily handled if their structure and contents are ``simple" enough.

To reach a practical, strict definition of long texts/documents for the purposes of this article, various types of documents that are typically considered to be long have been taken into account. This includes research papers, which are often utilized as input for long document analysis tasks \cite{sumpubmed, longformer, big_bird, clement} and that typically feature a word count in the range $3,000$ to $10,000$ words \cite{bjork}. This also includes books, which have an average of $80,000$ to $170,000$ words, and long document datasets such as arXiv \cite{clement} and Pubmed (see Section \ref{sec::Datasets}). Thus, the following definition is utilized in the sequel:

\textbf{A ``long document" dataset contains documents with an average word count of at least $2000$ words per document, reaching potentially up to $170,000$ words or more. Additionally, each long document must not be semantically segmented, in the sense that context and entity relationships may persist across different paragraphs and sections}. Long document NLP algorithms must be able to cope with the relatively large average size of these documents, but also handle cases with sizes way above that average. They must also be able to hold and analyze dependencies and contexts across vast sections of text.

This definition covers the typical document types outlined above (tweets, reviews, research papers, books, etc.), while also clearly distinguishing between fuzzier datasets. For example, despite being of identical domain and very similar content, the CNN/Daily Mail, \cite{hugging_face_cnn_dailymail}, 20NewsGroups \cite{20groups} and Hyperpartisan \cite{hyperpartisan} datasets do not fall under the same category. The first one does not meet the quantitative or qualitative criteria (average word threshold/long-span dependencies) of the proposed definition, as opposed to both 20NewsGroups and Hyperpartisan that are indeed frequently used for long document analysis research.

\subsection{Shared Challenges}
Long document analysis algorithms, irrespectively of the specific task, inevitably face various challenges due to the nature of the inputs. A subset of them are common for all input texts, while others are specific to the case of long documents. Below, follows a non-exhaustive list of the most significant relevant challenges.

\begin{itemize}
\item \textbf{The curse of dimensionality}. As the supported vocabulary size grows, the number of possible test-stage word combinations grows exponentially, unlike the fixed training dataset size \cite{bengio2003}. Thus, in semantic document analysis, the longer the test text, the more likely it is for the DNN to encounter sentences unseen during training. This is a challenge for DNNs in the common case of data distribution shifts that lower generalization ability and degrade inference accuracy.

\item \textbf{Polysemy}. A word may express multiple different meanings in different contexts. For example, the word ``bank" may mean a financial institution or a building. DNN models must learn to appropriately differentiate the exact meaning of a word, depending both on the local context (sentence, paragraph) and the global context (document).

\item \textbf{Homonymy}. Homonyms are words that either share spelling (homographs), pronunciation (homophones) or both, but are not semantically related. For instance, the term ``bank" may refer to a financial institution or to the shore of a river. The difference between homonymy and polysemy is that homonyms are not at all related to each other in meaning.

\item \textbf{Figurative language}. Sarcasm, irony and metaphors pose serious challenges to NLP, since the real meaning of a phrase is different from the immediately obvious one \cite{Karamouzas2022}. Figurative language is common to both short and long texts, but in the latter case it can take the form of sustained allegories (e.g., in literary books).

\item \textbf{Unstructured text}. Long texts may be very unstructured for the most part, while communicating information in rather indirect and abstract ways. For example, consider a novel, partitioned in a few dozens of chapters, and contrast it with short texts, such as product reviews or tweets. In the first scenario, crucial information and layers of meaning may be distributed across numerous very long chapters \cite{worsham}. Furthermore, long documents tend to follow constantly evolving textual and contextual conventions, leading to a higher chance of shifts between learned and actual probability distributions. This is especially problematic for literary works, which are fluid and susceptible to changing cultural biases \cite{brazil}.

\item \textbf{Foreign language datasets}. There is limited availability of training content for languages other than the most popular ones (e.g., English, French, Spanish, etc.). Given that both the word embedding networks and the main, task-specific NLP DNNs are typically trained on one language at a time, the lack of extensive datasets in rarer languages leads to less optimal embeddings and, as a result, reduced performance in downstream tasks. The situation is aggravated by limited training sets for the downstream tasks themselves. Thus, less data-intensive machine learning algorithms may be preferred over DNNs for rare-language texts, such as decision tree classifiers \cite{brazil}, or the existing datasets can be artificially augmented \cite{geroge}.
\end{itemize}

Additionally, certain challenges are tied to the nature of the existing DNN architectures. Below, follows a list of such issues that are common across most document analysis tasks (e.g., classification, summarization, etc.), but they are prominent in the case of long document inputs, in contrast to short texts.

\begin{itemize}
\item \textbf{Input size}. The size of the input document is the most significant challenge in long document analysis. For example, the computational overhead of CNNs tends to grow proportionally to their inputs; thus, they are unable to process documents with thousands of sentences, without specialized hardware. On the other hand, RNNs/LSTMs struggle to conserve information and identify long-term dependencies over huge spans of text \cite{worsham_book}. Transformer DNNs, such as BERT and most of its variants, are typically constrained by a rather low maximum token sequence length; usually 512 or 1024 tokens. This limits their training to relatively short texts, forcing brutal token selection during inference. As a result, their performance on long texts is severely limited \cite{dai}. Additionally, vanilla Transformer architectures \cite{ilia} rely on the typical \textit{global self-attention} mechanism, comparing each token to all other input tokens at each self-attention head, leading to a quadratic ($\mathcal{O}(n^2)$) increase in computational and memory requirements with respect to input size.

\item \textbf{Long-range dependencies}. SoA DNN architectures for long document analysis do not rely entirely on a global context, due to computational and qualitative constraints \cite{dai}, thus they may attempt to reduce the input's segments where they consider local context. However, this strategy may not work sufficiently well for documents with long-range context dependencies spanning over multiple, and often highly separated sections of text. This is mostly an issue with unstructured inputs (e.g., novels), but in fact it may arise even with highly structured documents if they are long enough.
\end{itemize}

Ignoring these challenges is equivalent to turning one's back to a plethora of very important use cases and real-world applications. As stated in \cite{poland}, small-to-medium document summarization is not as important as long document summarization, given that smaller texts can be reasonably read by a human without the need for AI automation. In today's era of pretrained LLMs, this is also a task commonly designated to them via prompting \cite{ouyang2022training, iyer2022opt, muennighoff2022crosslingual}, but this case falls outside the scope of this article. Additionally, document classification is impossible in important domains such as books (e.g., genre identification \cite{worsham, worsham_book, brazil, jose}, library management \cite{jose}), academic papers \cite{gales} and legal documents \cite{lulu, dale, merchant} without specialized models.

\subsection{Common Strategies}
In order to cope with the above challenges, DNNs specifically targeting long document analysis frequently use a combination of the following strategies:

\begin{itemize}
    \item \textbf{Sliding Window}, where embeddings are created using a sliding window over the input tokens. This is frequently used to capture local context from long spans of text.
    \item \textbf{Truncation}, which involves minimizing the input text or input embedding size in order to conform to input size limits. This is usually achieved by raw text selection, but can be applied to embeddings as well.
    \item \textbf{Sparse Attention}, where the underlying Transformer architecture is changed in order to allow for larger inputs.
    \item \textbf{Iterative Embedding Creation}, where embeddings are iteratively combined and aggregated in order to finally produce a fixed-sized document embedding. This is most often met in hierarchical architectures.
\end{itemize}

These strategies are frequently combined with Transformer architectures, in order to allow them to cope with high document length. Further details can be found in the next sections.

\subsection{Categorization and Evolution of Long Document Deep Neural Architectures}
This subsection offers an original taxonomy of SoA deep neural methods for long document analysis. This taxonomy is composed of two main categories: neural architectures which abide by the Transformer token input limit, and ones that modify the underlying Transformer architecture to allow for very large token limits.

The first category contains methods employing one of two main strategies: i) either carefully selecting the most representative text and feeding it as-is to the model (\textbf{Feature Selection}) or, ii) transforming the input in order to encode as much information as possible given the fixed token limit (\textbf{Hierarchical Models}). Feature selection may be as simple as using user-defined rules from common heuristics, or as complex as using DNNs and Information Retrieval (IR) theory to select the passages with the largest information gain. Hierarchical Models, on the other hand, use stacked neural layers to iteratively aggregate and combine knowledge from distinct parts of the text, finally generating a ``document embedding" that is fed to a typical Transformer (usually BERT).

Another valid strategy is bypassing the token limitation all-together. This is achieved by two main strategies: i) the attention mechanism may either be modified to allow for linear growth of model parameters with respect to the input tokens (frequently called \textbf{Sparse Attention}), or ii) the Transformer itself may be modified to contain hidden states much like an RNN, allowing the model to access previous knowledge from arbitrary points in the past. Using the terminology presented in \cite{dai-etal-2019-transformer}, the latter approach is called a \textbf{Recurrent Transformers} in this article.

A complete view of the presented taxonomy can be found in Figure \ref{fig::taxonomy}. Note that certain DNNs may belong to more than one categories, borrowing ideas from both strategies. The taxonomy includes the approaches that are being detailed in further sections of this article. The evolution of these methods over time is summarized, with respect to this established taxonomy, in Figure \ref{fig::timeline}.

\begin{figure}
    \includesvg[inkscapelatex=false, width=14cm]{images/taxonomy.svg}
    \fontsize{8}{10}\selectfont
    \centering
    \caption{A proposed taxonomy of DNNs designed for long document analysis, focusing on methods that can be applied for classification or summarization. The methods are presented as the tree leaves, while their color denotes the NLP task they were designed for.}
    \label{fig::taxonomy}
\end{figure}

\begin{figure}
    \includesvg[inkscapelatex=false, width=14cm]{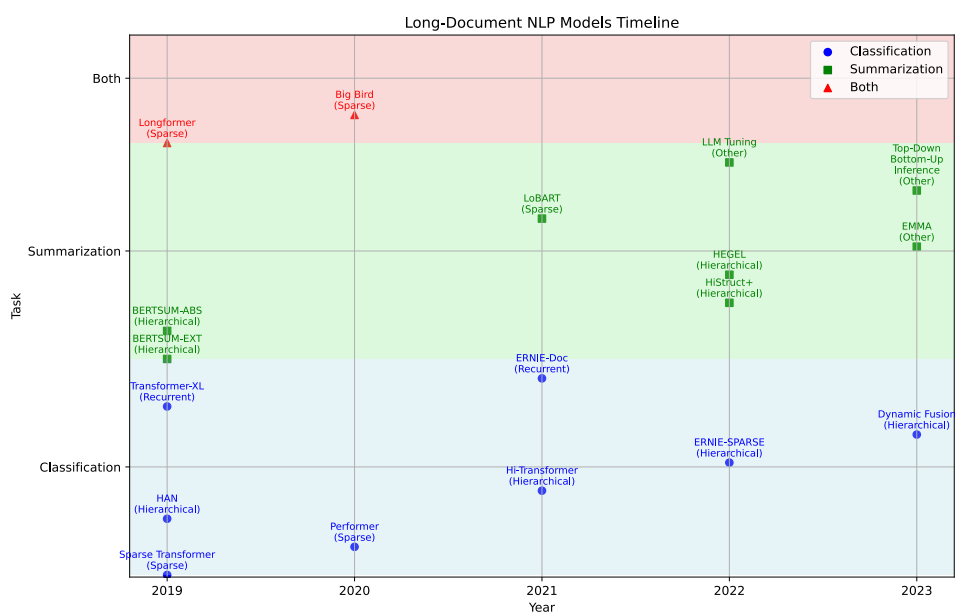}
    \fontsize{8}{10}\selectfont
    \centering
    \caption{A timeline of DNNs for long document analysis, focusing on methods that can be applied for classification or summarization, according to the taxonomy in Figure \ref{fig::taxonomy}. Notice the gradual shift from Sparse to Hierarchical Models. The most recent approaches, such as EMMA, tend to diverge from most common strategies.}
    \label{fig::timeline}
\end{figure}

\section{Document Classification}
\label{sec::Classification}
Document classification is the automatic assignment of one or more class labels to a piece of text, based purely on its contents and assuming a known finite set of discrete labels. In the long document case, the text in question is typically an entire document. As an example, a book of fiction could automatically be assigned a genre label by a text classifier deployed in a library, supporting labels such as "Romance" or "Science Fiction". However, a classification-specific challenge is that of \textbf{multiple labels}: the longer an input document is, the more probable is that it belongs to multiple semantic classes concurrently.

Overall, recent methods for neural long text/document classification mostly rely on pretrained Transformer language models (e.g., BERT) with a classification head appended at the end. The language model may be finetuned for the desired downstream classification task, while training the head. This approach is suitable for most relevant tasks \cite{qian} \cite{ion} \cite{ion2} \cite{mcbert} \cite{ammar} \cite{so}. However, Transformers may be suboptimal for specialized tasks or domains \cite{junhua}, allowing CNN and RNN architectures to occasionally outperform them on long documents. Relevant issues and solutions are discussed below.

\subsection{Solutions}
\label{ssec::ClassificationSolutions}

Traditional machine learning algorithms, such as Naive Bayes \cite{russel}, SVMs (Support Vector Machines) \cite{cortes} and ID3 decision trees \cite{quinlan} have been repeatedly applied for document classification \cite{brazil} \cite{xu}, being competitive even in large literary texts \cite{sicong}. Ensemble algorithms have also proven capable in such tasks especially in the context of hybrid schemes \cite{onan_2017}. However, the field has generally moved decidedly towards DNNs and this subsection presents the relevant SoA. The innovations of the various existing approaches mostly fall under the following categories, which are more or less tied to the challenges faced by long document classification.

\subsubsection{Multi-label Classification}
Multiple labels are essential for many real-world applications \cite{hsu} \cite{patentnet}, thus many methods attempt to identify a document as belonging to multiple classes concurrently. This may be trivial to achieve in itself, using well-known DNN models that have been properly adapted. The varying number of labels per document and class imbalance issues, which may lead to learning extremely biased patterns from the training data, can be handled by applying weighted loss functions. However, multi-label classification poses unique challenges with regard to how accuracy is measured.

One representative SoA approach attempting to tackle this scenario is \cite{sicong}, where multiple different classifiers (including a pretrained BERT) are comparatively evaluated in multi-label multi-class document classification, using the book genre recognition task as a benchmark. The steps taken in \cite{sicong} to build a successful genre recognizer are telling of more general difficulties with long document classification. Regarding the issue of multiple labels, a modified accuracy metric is exploited to measure classification accuracy: the true positive and true negative classifier predictions are first computed separately per class label, with the proposed metric being the weighted average of correctness over all the label classes. Per-class weights are proportional to the class label frequency and normalized to sum to 1. Similarly, the loss function employed during training (Binary Cross Entropy) is modified to include weights for each label, so that each class is scaled inversely to the prominence of the class in the training set.

\subsubsection{Feature Selection and Pruning}

This is the traditional way to handle long textual inputs. It involves systematically discarding as much text as possible at the preprocessing stage, while minimizing information loss. In order to cope with the potentially huge input size in long document analysis, almost all relevant NLP methods are prefaced by a preprocessing stage, where less useful input elements are removed. Appropriately reducing the input size improves both computational efficiency and accuracy.

\paragraph{Text Sampling} The simplest way to achieve this is by sampling the document, with a combination of user-defined rules and random chance. For instance, the book genre recognizer of \cite{sicong} uses an aggressive sampling method which first discards:

\begin{itemize}
    \item "stopwords", i.e., known, extremely frequent words such as "a", "and", "he", etc.,
    \item words that appear less than 20 times,
    \item words that appear in more than 75\% of all books,
    \item words that appear in more than 75\% of all classes,
    \item paragraphs where the frequency of the remaining words is less than a certain threshold,
    \item the contents of each remaining paragraph after the 512th token, because of technical limitations imposed by BERT.
\end{itemize}

Thus, each remaining paragraph is represented by a subset of its words that fall within a restricted, book-level set of ``keywords". Finally, the sampled paragraphs are only those with the highest concentration of keywords. The goal of this ruthless sampling is to keep the input's length manageable for documents containing many thousands of pages. Methods similar or identical to this one are present in most NLP models specialized for long texts; thus, they are utilized extensively in most research cited in this paper.

On the other hand, the book genre recognizer of \cite{worsham} employs only minimal preprocessing, just enough to allow the DNN to run relatively efficiently. It relies on an index, consisting of the 5000 most common words from the supported vocabulary. No other preprocessing/sampling takes place, under the intuition that exposing the DNN to even statistically unimportant words facilitates an understanding of patterns and grammatical modifiers, such as tense and plurality. Words are represented by typical neural embeddings. Different strategies were evaluated for feeding the document to the classification DNN:
\begin{itemize}
    \item Feed only the first 5000 words of the document.
    \item Feed only the last 5000 words of the document.
    \item Feed only 5000 random words of the document.
    \item Feed the entire document, split into chapters.
    \item Utilize a traditional bag-of-words approach.
\end{itemize}

These strategies were then employed on a series of different architectures. Ultimately, the random word selection proved to consistently outperform other selection strategies for a multitude of classifiers, with the best performer being a CNN-Kim \cite{kim} architecture.

\paragraph{Using Information Retrieval (IR)} A more advanced and recent approach to text/feature selection is presented in \citet{li_2023}. It relies on local query-block pre-ranking, allowing the selection of blocks of text that hold the largest semantic significance; these can then be fed to a computationally demanding DNN such as BERT.

Segmentation of the document into blocks is accomplished using the CogLTX Block Decomposition method \cite{ding_2020}. CogLTX is a dynamic programming segmentation approach which emphasizes text containing punctuation marks such as "." and "!". It uses an upper token limit on blocks, ensuring a manageable and uniform structure. The ranking of each individual block is based on their Retrieval Status Value (RSV), which is measured either by BM25 (a ranking function frequently used by search engines \cite{amati_2009}) or by simple cosine similarity.

KeyB(vBERT) is a BERT variant whose inputs are derived by the above feature selection process. KeyB(PARADEk) instead uses PARADE \cite{li_2021_parade}, a SoA set of IR methodologies (involving DNNs) that compute a query-document representation. The latter one is fed alongside the text to a pretrained BERT. Both approaches outperform vanilla ones on IR, although concerns over the memory requirements of KeyB(PARADEk) are raised in \citet{li_2023}.

Similar IR-based methods for feature selection may significantly contribute to the use of full Transformer architectures, in parallel with Hierarchical and Sparse Attention Transformer variants that follow a different rationale for bypassing computational/memory limitations (see below).

As a side-note, it must be pointed out that such IR-based methods for feature selection are not identical to the use of pretrained extractive summarization DNNs. While in principle the preprocessing steps are indeed similar, their computational and memory requirements are significantly higher, while they are likely to be outperformed by modern, dedicated summarization DNNs (see Section \ref{sec::Summarization}).

\subsubsection{Sparse Attention Transformers}
Given the quadratic computational complexity of Transformer DNNs and their prominence in recent years, attempts have been made to reduce their cost. Such approaches try to achieve linear complexity with respect to the input size, by keeping track of a small, fixed-size window of neighboring tokens around each input token, instead of considering all $n$ tokens of the input sequence within each self-attention head. However, the unavoidable trade-off is a reduced ability of the DNN to capture long-term dependencies, which is as important for long texts as the complexity with respect to the input size. It must be noted that such efficient Transformer architectures are task-agnostic and generic ones for long input sequence analysis, not tied to a specific task, although they are presented here in the context of document classification. Below, the terms token and element are used interchangeably, although strictly speaking only the first Transformer layer receives actual tokens as input.

\paragraph{Sparse Transformer} The Sparse Transformer was introduced in \citet{child}, to lower asymptotic complexity from $\mathcal{O}(n^2)$ to $\mathcal{O}(n\sqrt{n})$. This is done by applying sparse factorizations to each \textit{self-attention matrix} computed by the Transformer during inference, which contains the current attention weights of each input element against each of the other elements of the input sequence. Given that such a matrix is separately constructed at each self-attention head of each layer, it is important to effectively reduce the computational and memory cost of this operation. The sparse attention mechanism relies on replacing full attention with several small, optimized attention operations, which jointly approximate the original mechanism sufficiently. Essentially, each token attends to only a subset of the overall input sequence tokens. Additional optimizations, such as efficient sparse attention kernels and improved weight initialization, contribute to a less demanding architecture, which is thus able to handle longer sequential inputs. The Sparse Transformer is a general model suitable for various applications involving very long input sequences, ranging from image compression to document analysis.

\begin{figure}
    \includesvg[width=16cm]{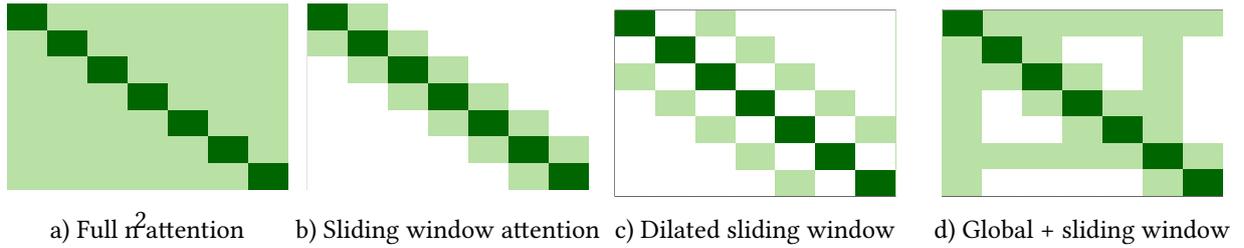}
    \centering
    \caption{Comparing the full self-attention pattern and the configuration of attention patterns in a standard Longformer.}
     \label{fig::longformer}
\end{figure}

\paragraph{Longformer} Surpassing the Sparse Transformer's computational gains, the Longformer \cite{longformer} achieves a linear asymptotic cost and is specialized for long document NLP. Its self-attention mechanism can be directly incorporated within other Transformer models and operates by employing a \textit{dilated sliding attention window}, which is a modification of the classical sliding window approach for \textit{local attention} \cite{saeed, odysseas}. In the typical case, each token's window consists of $w$ tokens surrounding it: $\frac{w}{2}$ tokens to its left and right, respectively. The underlying intuition is that the semantic context of a token can be mostly derived from its neighboring tokens, while the computational complexity of this operation is $\mathcal{O}(n \times w)$. In a sense, the end effect resembles the local receptive fields of neurons in convolutional layers. Similarly to CNNs, stacking $l$ Longformer layers gradually increases this receptive field, so that representations of tokens faraway from the query can be attended within the later layers.

The \textit{dilated sliding window} approach modifies the sliding window with a predetermined dilation factor $d$ added at each step, which determines the spacing between adjacent window positions and allows data analysis at multiple scales simultaneously. Thanks to the introduced "gaps" in the attention pattern, a larger range of tokens within the self-attention matrix can be covered without increasing calculations and computation time. Thus the effective receptive field of the window is of length $l \times d \times w$, while memory and processing costs remain steady.

Since windowed and dilated attentions may not be enough to extract a suitable sequence representation, certain additional, pre-selected locations of the full self-attention matrix are also allowed to be considered as global context. These global tokens attend to/are attended by the entire sequence as normal, but their number is kept constant and relatively small to limit complexity to $\mathcal{O}(n)$. The [CLS] token is a good candidate for being selected as a global one, since its generated representations should convey information about the entire sequence. Overall, as a result of Longformer's innovations, it has been successfully evaluated with inputs of length up to 4096 tokens (compared to BERT's maximum of 512 tokens). A comparison between full, sliding, dilated sliding and global-dilated sliding window self-attention patterns is depicted in Figure \ref{fig::longformer}.

\paragraph{BigBird} BigBird \cite{big_bird} can be seen as a variation of the Longformer, attempting to support much longer input sequences by reducing the complexity of self-attention from quadratic to linear. It also employs a sliding window, since in typical NLP tasks the semantic context of a token can be mostly derived by its neighboring tokens. Instead of the Longformer's dilated sliding windows, it employs random token selection to reach faraway tokens and provide context. This complements the global attention and the regular sliding window attention schemes, adopted and adapted from the Longformer. It has been demonstrated mathematically that this modified sparse attention mechanism satisfies many theoretical properties of the original full attention mechanism, such as universal approximation and Turing-completeness, but seems to require more layers to retain accuracy comparable to full attention architectures.

\paragraph{Performer} Performers \cite{performer} are Transformer architectures which can estimate regular full-rank-attention Transformers with provable accuracy, but using only linear space and time complexity. They achieve this through two key aspects, described below.

To efficiently compute approximate attention, \citet{performer} introduces the FAVOR+ (Fast Attention Via positive Orthogonal Random features) algorithm. It enables linear-time and constant-memory complexity, making it more scalable for long sequences of data such as long documents. As established in Section \ref{ssec::Transformers}, the standard attention mechanism involves calculating an attention matrix $\mathbf{A}$, which has a quadratic time and space complexity with respect to the sequence length $L$. In FAVOR+, the attention matrix $\mathbf{A}$ is approximated through a randomized mapping $\phi$ and a kernel function $K$. The kernelized attention is defined as $\mathbf{A}(i,j) = K(\mathbf{q}_i^T,\mathbf{k}^T_j)$ where $\mathbf{q}_i^T$ and $\mathbf{k}^T_j$ are the $i$-th query and $j$-th key row-vectors, respectively. The key innovation lies in the use of random feature maps $\phi$, with the mapping function defined as $\phi(x) = \frac{h(x)}{\sqrt{m}} (f_1(\omega_1^T x), \cdots, f_l(\omega_1^T x)$ allowing the modeling of various kernels used in practice.

The resulting attention mechanism, denoted as FAVOR+ attention, is characterized by significantly improved time and space complexity compared to regular attention. FAVOR+ has a space complexity of $\mathcal{O}(Lr+Ld+rd)\mathcal{O}(Lr+Ld+rd)$ and a time complexity of $\mathcal{O}(Lrd)$, where $r$ is the dimensionality of the random features. This represents a substantial reduction from the quadratic complexities associated with standard attention mechanisms. Moreover, FAVOR+ addresses stability concerns associated with negative dimension-values in random feature maps. By introducing a robust mechanism, it mitigates potential issues and generally provides a more reliable and effective solution for approximating attention scores.

The last step of FAVOR+ utilizes the standard Gram-Schmidt orthogonalization procedure in order to transform the $\omega_i, i=1,\cdots, m$ vectors to orthogonal vectors, reducing the variance of the attention mechanism's estimator. The authors prove that by utilizing orthogonal features, the variance reduction applies to any dimensionality $d$, instead of previous results necessitating large values of $d$.

A recent improvement on Performer is the Permuteformer architecture \cite{permuteformer} that adds a position embedding vector to the Performer's input, while keeping its linear scaling for long sequences. The permutations are achieved by applying a position-aware transformation on query features and key features to encode positional information. The cost of these permutations is negligible compared to the overall computational cost of the Performer, resulting in these two being almost equally efficient and requiring almost identical amount of training time and resources. Regarding results, Permuteformer improves upon Performer in all classification tasks and accelerates its convergence.

\subsubsection{Hierarchical Transformers}
Hierarchical architectures attempt to handle the large input size of long texts by appropriately building upon original Transformers, instead of modifying them. As before, they are task-agnostic and generic ones, but presented here in the context of document classification. There are two main approaches to building a hierarchical Transformer:

\begin{itemize}
    \item By transforming the Transformer's input via a suitable DNN (e.g., a CNN or RNN/LSTM). The latter one generates a single document representation (document embedding) able to fit into a standard Transformer (e.g., BERT), thus bypassing the input size limitations.
    \item By segmenting the input document. The input segments are independently truncated to the Transformer's input size and the network (e.g., BERT) generates embeddings for each of these segments. A separate DNN then combines the segment embeddings and predicts the document label.
\end{itemize}

\paragraph{Hierarchical Attention Network (HAN)} The first attempt towards a Hierarchical Transformer \cite{ion_han} \cite{zichao} was built on top of standard BERT. During inference, the input document is split into paragraphs that are truncated to fit into BERT. The latter one generates independent paragraph embeddings, which are combined by the succeeding HAN into a single output. The end result is an increased ability to handle long texts, by exploiting the natural partitioning of documents into paragraphs.

A similar approach was later used in \cite{glue_gunner}, where the base models were BERT, RoBERTa, DeBERT \cite{yanguang}, Legal-BERT \cite{ion6} and CaseLaw-Bert \cite{zheng}. In long text analysis tasks, the hierarchical versions proved competitive with Sparse Attention Transformers, such as Longformer and BigBird.

The method in \cite{khandve}, relying on BERT or on Universal Sentence Encoders (USEs) \cite{use} (models learning sentence representations for transfer learning), follows up on this methodology, but replaces the final HAN with a CNN or an LSTM architecture. The BERT+LSTM combination proved to be the best among the evaluated hierarchical approaches, which however were generally outperformed by Sparse Attention Transformers.

\paragraph{Hi-Transformer} The main alternative to \cite{glue_gunner} and its variants is to essentially reverse the process: that is, to employ a hierarchical DNN for obtaining a fixed-size document-level representation/embedding from the entire input, which is then fed to a regular classification DNN. \textit{Hi-Transformer} \cite{qi} achieves this via a Transformer acting as a "Sentence Encoder" (SE), i.e., aggregating and projecting the regular, individual word embeddings of the input document into sentence-level embeddings. Once every sentence has passed through SE, these embeddings are subsequently ordered by positional embedding and fed as input to a subsequent Transformer called "Document Encoder" (DE). The latter's output is a context-aware document embedding, which is fed as input to the next SE along with the original sentences. Thus, the generated, revised sentence embeddings are aware of the global, document-level context. This process is repeated by stacking multiple such layers, until the final output is passed to a pooling layer that extracts the final document embedding. According to \cite{qi}, two such layers seem to be sufficient for outperforming Sparse Attention Transformers in long text analysis. This can be attributed to the readily accessible global/document-level context. Since the final classifier has to analyze a fixed-size document representation and the SE's complexity is linear to the number of sentences, computational demands are not higher than in the sparse attention case. More impressively, Hi-Transformer's accuracy has been observed to actually increase with longer documents, since more relevant context can be extracted.

\paragraph{Hierarchical Sparse Transformer} Hierarchical Transformers have been combined with the sparse attention mechanism, in order to ameliorate the latter's inability to capture all the necessary global/document-level context, in the case of long texts. This inability arises from the low level of global context diffusion in the generated token representations, since the global tokens are limited in number. The \textit{ERNIE-SPARSE} architecture \cite{liu} merges the two schools of thought by using hierarchical attention, which is then fed to a Sparse Transformer in order to increase the information extracted from the global context.

\begin{figure}
    \centering
    \includesvg[width=14cm]{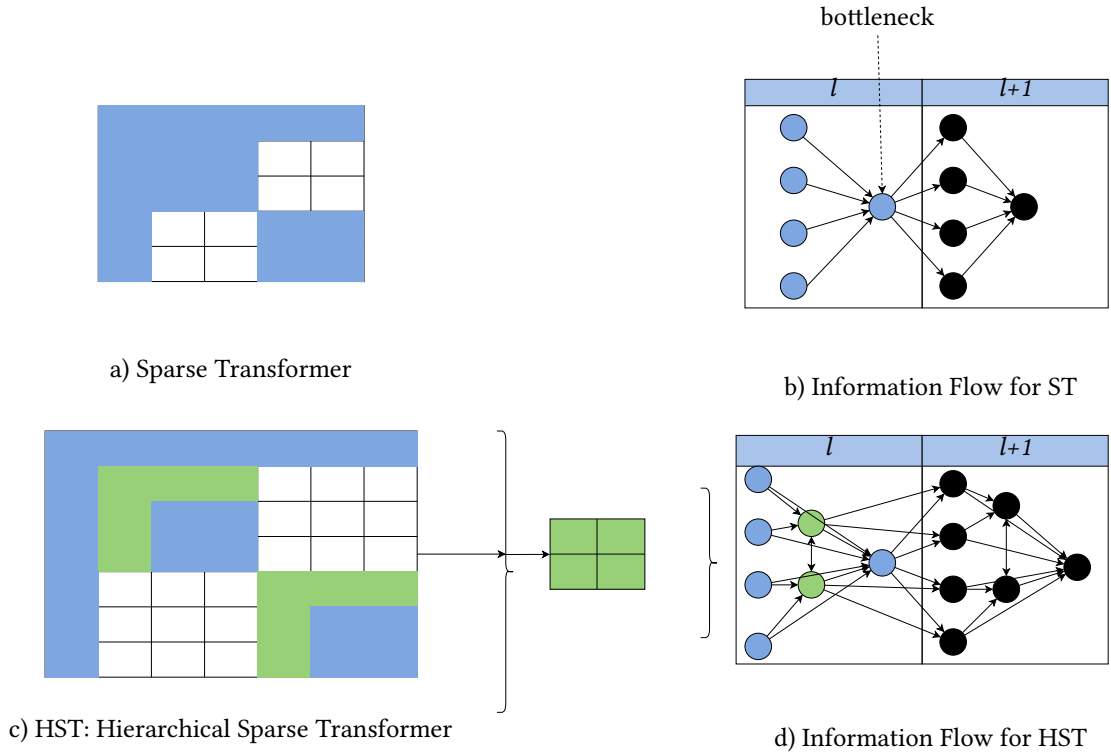}
    \caption{Comparison of Sparse Transformer (ST) and Hierarchical Sparse Transformer (HST). (a) Sparse Transformer comprises global attention and local attention. (b) ST faces a limitation where all sequence details are compressed into a fixed vector size. (c) HST addresses this limitation by incorporating representative tokens into local attention, enabling hierarchical attention. (d) The information flow in HST demonstrates how the interaction among representative nodes can enhance global information interaction pathways.}
    \label{fig::ERNIE}
\end{figure}

ERNIE-SPARSE utilizes a modified attention mechanism, where each query is allowed to attend to a limited number of additional representative tokens within each fixed-size window of the self-attention matrix, along with regular global and local tokens from the input sequence. These inserted representative tokens carry global context from the entire document and are derived in two steps: first, the regular Sparse Transformer is applied to the unaltered input sequence and, then, regular full self-attention is applied to a small collection of selected outputs. ERNIE-SPARSE has been shown to outperform the competition on well-known document classification datasets. It can be considered both a hierarchical and a sparse attention approach, since it modifies the Transformer's internal architecture. Figure \ref{fig::ERNIE} depicts the modified attention mechanism of ERNIE-SPARSE, in comparison to that of the Sparse Transformer.

\paragraph{Hierarchical BERT-Based dynamic fusion}
Another approach taken by \citet{onan_hier} for processing very long documents combines hierarchical information through nodes capturing nested relationships and semantic structures in text data, contextual information through pretrained Transformer models such as BERT or GPT, and a novel dynamic fusion approach. The latter combines outputs from previous stages with an external model, enhancing overall performance. The architecture proceeds via the following stages:

\begin{itemize}
    \item \textbf{Linguistic Feature Extraction} where features such as Part-of-Speech (POS) Tagging, Dependency Parsing, Named Entity Recognition (NER) and Coreference Resolution are extracted.

    \item \textbf{Hierarchical Node Construction with Domain-Specific Knowledge} The Hierarchical Graph is first defined as follows: a graph $G_i$ is constructed for each document $t_i$, where $N_i$ represents nodes (document and sentences) and $E_i$ represents hierarchical relationships. Document nodes are formalized as $N_i = \{n_1, n_2, \cdots , n_{mi}\}$, where $n_1$ is the document node, and $n_2$ to $n_{mi}$ are sentence nodes. Edges $E_i$ represent relationships between the document and its sentences. For each document node, a document graph is created in a similar fashion: domain-specific knowledge graphs, such as WordNet \cite{wordnet}, are integrated to enhance the capture of relationships and semantic meanings between words. Word nodes are formalized as $V_i = \{v_1, v_2, \cdots , v_{mi}\}$, where $v_i$ is the node of the i-th word. The edge $E_i = <v_k, v_l>$ exists only if a domain-specific relationship exists between the $v_k$ and $v_l$ words. The two graphs are then fused into a single hierarchical graph.

    \item \textbf{Dynamic Text Sequential Feature Interaction} which the Dynamic Time Warping (DTW) \cite{dtw} algorithm to align word embeddings, facilitating clearer distinction of temporal relationships between words. The algorithm identifies the optimal path through the sequence of embeddings, minimizing the distance between two sequences. In this context, the involved sequences are the word embeddings for a specific document and a reference sequence, typically represented by average embeddings for a designated category. The integration of the DTW algorithm markedly enhances the model's capacity to capture temporal relationships between words, augmenting the model's understanding of the sequential structure inherent in textual data. This holds particular significance in text classification scenarios, where discerning the temporal order of words is imperative for tasks such as sentiment analysis.

    \item \textbf{Hierarchical Graph and Knowledge Graph Fusion}. The fusion process involves combining the hierarchical graph and knowledge graph to obtain an integrated graph. Several approaches can be employed for fusion, such as graph convolutional networks or attention mechanisms.

    \item \textbf{Model Functionality}. The hierarchical/linguistic representation is fed into a pretrained BERT model, which generates a contextual vector that is processed by an MLP. The latter one outputs the class labels.
    
\end{itemize}

The architecture achieves SoA results in many different benchmark datasets.

\subsubsection{Recurrent Transformers}
The fixed input size of baseline Transformers leads, during both training and inference, to \textit{context fragmentation} in long document analysis: essentially, the DNN analyzes the input as independent, consecutive segments. Thus, a complementary direction of the SoA is to allow the Transformer to learn and extract long-term dependencies beyond the preset fixed input sequence size, while maintaining linear computational complexity and retaining local and global context. As before, they are task-agnostic and generic architectures, but presented here in the context of document classification. 

\paragraph{Transformer-XL and XR-Transformer} The Transformer-XL architecture \cite{dai-etal-2019-transformer} introduces recurrence into the Transformer, thus integrating the RNN concept of retained hidden states. These states are passed within the Transformer from one segment to the succeeding one, thus effectively transmitting the previously established context and allowing the identification of long-term dependencies across segments. Unlike RNN/LSTM recurrence, where each layer's stored state is exploited at the next time step by the same layer, the corresponding Transformer-XL mechanism shifts the transmitted state one layer downwards. For example, layer $l=n$ has access to 2 hidden states from layer $l=n-1$, 4 hidden states from layer $l=n-2$ and $2^n$ from layer $l=1$, which are the original input word embeddings.

This mechanism guarantees $\mathcal{O}(n \times l)$ cost, where $l$ are the number of layers, and significantly faster inference compared to baseline Transformer, since representations computed for previous segments are re-used instead of being computed from scratch for each new segment. It also allows different input sequence sizes between the training and the inference stage. Ultimately, Transformer-XL achieves competitive results on extremely long document analysis, while being significantly more efficient than the baseline Transformer.

Another similar approach aimed at extreme multi-label text classification is the XR-Transformer \cite{XR-Transformer} which employs a three-stage framework, those being partitioning, shortlisting, and ranking. During the partitioning stage, the set of labels is divided into clusters or groups based on certain criteria, in the shortlisting stage a subset of candidate labels are selected from the entire label set, allowing the XR-Transformer to focus on a smaller, more manageable set of potential relevant labels. Finally, the ranking stage involves assigning a priority or order to the selected labels based on their relevance or likelihood of being correct. The XR-Transformer also employs a structure similar to recurrence, but additionally introduces a Hierarchical Label Tree (HLT) that captures the relationships and dependencies between different labels in a structured manner. The HLT is constructed during the training phase through recursive label clustering, forming a tree-like, hierarchical organization of labels that allows training the Transformer on multi-resolution objectives.

\paragraph{ERNIE-Doc} ERNIE-Doc \cite{ernie-doc} builds upon Transformer-XL but makes two passes over the input sequence, similarly to how humans first "skim" a document before paying attention to important sections. To achieve this, the cached hidden state is computed as follows:
\begin{equation}
    \hat{\mathbf{H}} = [\hat{\mathbf{H}}^{1}_{1:T};...;\hat{H}^{N}_{1:T}] \text{ (skim phase)}
\end{equation}
\begin{equation}
    \tilde{\mathbf{h}}_{r+1}^{n-1} = [\tilde{\mathbf{H}};\mathbf{h}_{r+1}^{n-1}] \text{ (retrospective phase)},
\end{equation}
\noindent where $T$ is the number of consecutive document segments, $L$ is the length of the document, $N$ is the number of layers. $\hat{\mathbf{H}}^{i}_{1:T} = [\hat{\mathbf{h}}^{i}_1;...;{\mathbf{h}}^{i}_T]$ is the concatenation of the $T$ cached hidden states initially derived in the skimming phase for the $i$-th layer. Thus, $\tilde{\mathbf{H}} \in \mathbb{R}^{(LTN) \times d}$ is the concatenation of all $\hat{\mathbf{H}}^{i}_{1:T}$ (one per layer), Thus, context from the entire document can be exploited in the retrospective phase in order to form the extended hidden state $\tilde{\mathbf{h}}^{n}_t$.

\begin{figure}
    \centering
    \includesvg[width=15cm]{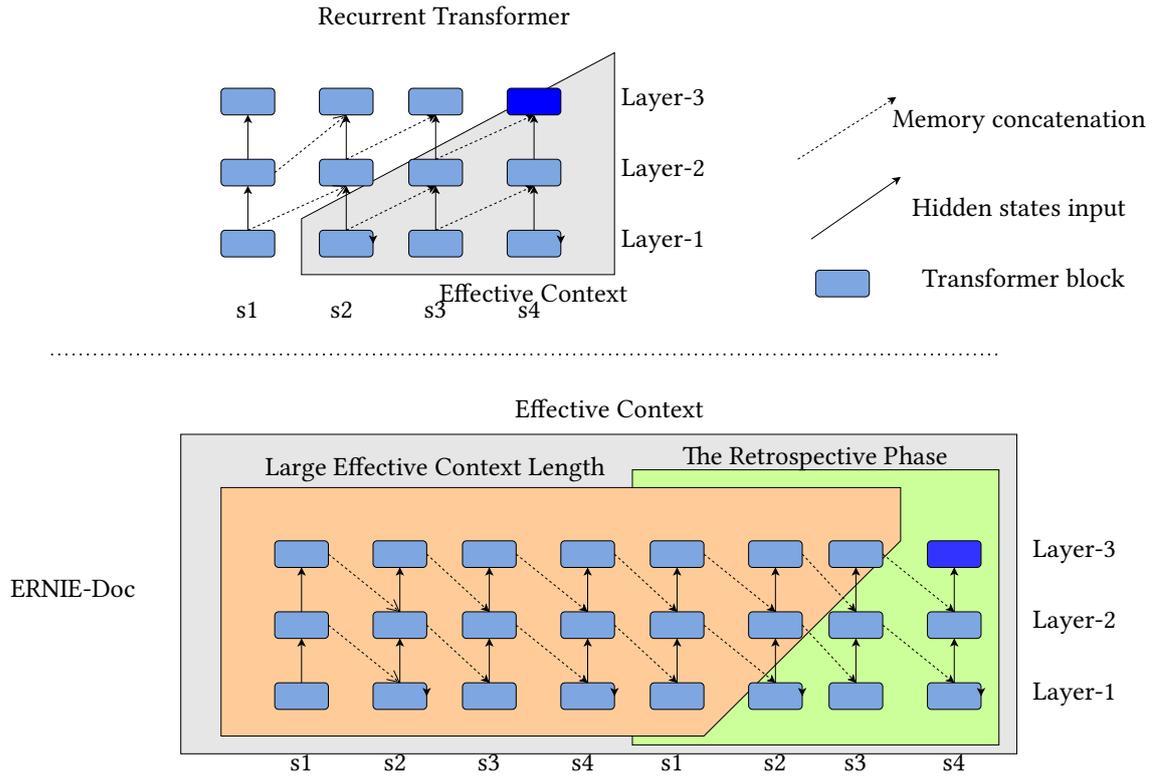}
    \caption{Comparison of 3-layer ERNIE-Doc and Recurrent Transformer, assuming a long document $D$ that is divided into four segments (S1, S2, S3, S4). \textit{Recurrent Transformer} (upper): While training on S4, it can only incorporate contextual information from the preceding two consecutive segments, S2 and S3, due to the linear growth of the effective context length in relation to the number of layers. \textit{ERNIE-Doc} (lower): With the assistance of an enhanced recurrence mechanism, the effective context length is significantly increased. Consequently, S4 can assimilate the information from S1, which was discarded by the Recurrent Transformer. Segments in the retrospective phase contain the contextual information of the entire document.}
    \label{fig::ERNIEDoc}
\end{figure}

This skimming architecture is incompatible with the recurrence mechanism from \cite{dai-etal-2019-transformer} and, due to its linear cost with regard to the number of layers, it may be prohibitive for long documents. To solve this, \cite{ernie-doc} "flattens" the hidden state dependency from one-layer-downwards recurrence to same-layer recurrence, similarly to RNNs. Additionally, a novel, document-level task for self-supervised pretraining is introduced that is called "Segment-Reordering Objective" (SRO). It entails randomly splitting a long document into $m$ segments, shuffling them, then letting the Transformer reorganize them in order to learn their interrelations. After pretraining on both the MLM task and on SRO, ERNIE-Doc proved highly competitive in several long document analysis datasets for a range of tasks such as classification, question answering and key-phrase extraction. Figure \ref{fig::ERNIEDoc} depicts the modified recurrent mechanism of ERNIE-Doc, in comparison to that of basic Recurrent Transformers.

\subsection{Sparse, Hierarchical or Recurrent Transformers?}
Sparse attention, hierarchical and recurrent approaches dominate the current literature for long document analysis using Transformers. However, a number of challenges remain, such as:
\begin{itemize}
\item Sparse Attention Transformers tend to suffer in accuracy, due to their reliance on a small number of global tokens that must encapsulate all document-level context.
\item Hierarchical Transformers are particularly susceptible to context fragmentation, since the individual segment embeddings are derived either without sufficient local context \cite{dai}, or with no access to the global document context \cite{qi}.
\end{itemize}

In general, there is no consensus on which approach performs better, nor any empirical rule on which tasks each architecture performs best in. Given that each approach features its own benefits and drawbacks, as well as potentially domain-specific strengths, the literature as of 2024 recommends manual architecture selection by the user, taking into account the specific problem. However, at a high level of abstraction, hybrid algorithms and recurrence seem to be promising avenues for advancing long text analysis.

\section{Document Summarization}
\label{sec::Summarization}
Automatic Text Summarization (ATS) is the task of generating a short summary from a source input document. In the context of long documents, such as medical and legal records, it is often called "document summarization". The generated summary must be coherent, concise and avoid redundancy, while accurately maintaining the meaning of the most important information in the source material. There are two main approaches to ATS: \textit{extractive} and \textit{abstractive} summarization.

\textbf{Extractive summarization} generates its summary by selecting verbatim sentences found in the source text. This practically reduces summarization into finding a binary selection vector $\mathbf{s} = [s_1, s_2,..., s_n]^T, s_i \in \{0,1\} \forall i \in [1,n]$, where $s_i = 1$/$s_i = 0$ if the $i$-th source input text sentence is/is not included in the summary, respectively. The simplicity and the tolerable computational cost of this approach have made it very common for on-line article summarization. However, its applicability is inherently limited by the fact that not all document types can be summarized well by just a few of their sentences taken verbatim and re-arranged. For example, there is no way to describe the content of most books by using a selection of their own sentences.

\textbf{Abstractive summarization} is the alternative approach of generating new text. It is the method humans use when trying to manually summarize text, by reading the source, extracting its meaning and then writing a condensed document which contains as much of the source's meaning as possible. As expected, modern abstractive algorithms are DNNs \cite{cho}. Compared to extractive alternatives, they generate more meaningful and condensed summaries, while being more adaptable to the input document type. However, there is a much larger risk of the output summary misrepresenting facts and/or not being coherent at all. The most commonly used method for abstractive summarization involves employing a sequence-to-sequence mapping (Seq2Seq) model \cite{seq2seq}, via an Encoder-Decoder DNN architecture. RNNs and Transformers have both been employed in this context, where the Encoder captures suitable representations of the document's tokens and feeds them to the Decoder that generates a summary.

Document summarization comes with its own set of task-specific challenges, essentially applying almost exclusively to the long input document case, since there is typically no reason to summarize a short text.

\begin{itemize}
    \item \textbf{Multiple topics}: The length of the input document gives rise to significant issues which go beyond simple computational and hardware requirements. Extractive DNNs particularly struggle with long documents \cite{xiao}; the greater the length of the input, the more topics it typically covers, thus the harder it is for a DNN to generate a summary covering effectively all of them.

    \item \textbf{Evaluation metrics}: The ROUGE evaluation metric \cite{rouge}, which is the most commonly used metric for summarization tasks, might not be sufficient for abstractive DNNs. It has been claimed to perform considerably worse than in the case of extractive summarization, since it lacks semantic understanding of the words it compares \cite{akter}.

    \item \textbf{Sentence extraction}: Framing extractive summarization as a binary decision task assumes that the meaning of individual sentences is independent of that of others, leading to high redundancy. Even accounting for this fact, such DNNs are still vulnerable to greedily picking general sentences, instead of finding a combination of specific sentences which together summarize the document much more effectively \cite{matchsum}.

    \item \textbf{Repetition}: One of the most common issues of abstractive DNNs dealing with long texts is sentence repetition; this is potentially caused by the over-reliance of a Decoder to its input, which gives rise to an endless loop of phrase repetition \cite{abigail}. Internal issues concerning RNNs with attention mechanisms have been pointed out as the root cause of both this issue and of false information appearing in the output summary \cite{suleiman}.

    \item \textbf{Hallucinations}: Abstractive DNNs additionally suffer from the opposite issue of hallucinations, i.e., generating summaries with content that is not actually present in the original document \cite{Wan2023faithfulness}. According to recent research \cite{van2022mutual}, hallucinations tend to appear during summary generation when the DNN is uncertain about how to continue, thus favouring common phrases it has memorized from its training set.
\end{itemize}
	
\subsection{Solutions}
This section reviews the commonly employed ROUGE evaluation score, as well as extractive and abstractive text summarization methods. The evolution of relevant algorithms, as well as how each family attempted to handle the challenges it faces, are briefly discussed.

\subsubsection{Evaluation Metrics}
\label{sss::SummarizationEvaluation}
ROUGE is a very commonly used metric for evaluating ATS algorithms, featuring as many as 192 variations. The base ROUGE metric \cite{rouge} measures the number of overlapping n-grams/sequences between the generated and ground-truth summary. This is sufficient for many extractive summarization tasks, but faces severe challenges when used for abstractive summarization. ROUGE is unable to distinguish between different grammatical versions of the same word \cite{suleiman} and synonyms \cite{akter}, while being agnostic towards sentence ordering \cite{graham}. Most importantly, the underlying assumption made by ROUGE, that the quality of a summary is correlated to its coverage compared to the human-generated summary, may not be statistically significant \cite{pitler}.

The reasons why ROUGE is still popular despite such severe drawbacks are multiple, including the unavailability of proven competitors and the lack of easy-to-use software implementations or standardized benchmarks for ATS evaluation \cite{fabbri}. Besides, alternative metrics may suffer from similar weaknesses, since they typically do not account for factual consistency between the source document and its evaluated summary \cite{Kry2020}. For a detailed discussion of text summarization metrics, the reader is referred to \cite{fabbri}. For the purposes of this article, a number of common and/or recent alternatives to ROUGE are briefly mentioned below:

\begin{itemize}
    \item METEOR \cite{meteor} (recommended by \cite{suleiman}), functions similarly to ROUGE, while being largely invariant to grammatical variants and synonyms.
    \item Sem-nCG \cite{akter} is a recent, semantically-aware metric focused on extractive summarization.
    \item BERTScore \cite{bert_score} uses the similarity of the contextual embeddings between the generated and the ground-truth summaries.
    \item $S^{3}$ \cite{s3} combines other evaluation metrics through a model, in order to produce an aggregated evaluation of the summary.
    \item BLANC \cite{blanc} is an unsupervised evaluation metric which does not depend on human-generated ground-truth summaries, by comparing BERT's performance on the Cloze task (on the source document) when it has and when it has not additional access to the generated summary.
    \item BARTScore \cite{bart_score} offers multiple directions (Faithfulness, Recall, Precision, F-score) which make it a flexible metric for different evaluation scenarios.
    \item Contextualized Topic Coherence Metrics \cite{CTC} are a group of metrics which evaluate the coherence and semantic similarity among the topics covered by a generated text.
\end{itemize}
Many more evaluation metrics have been proposed, but none has taken a foothold yet. The question of a capable successor to ROUGE is still an open one.

Besides summarization evaluation metrics themselves, factuality metrics have also recently proven instrumental in abstractive summarization due to their application in evaluating or fighting hallucinations. Three important and relevant recent metrics are the following ones:
\begin{itemize}
    \item FactCC (Factual Consistency Checking) \cite{Kry2020} is essentially a BERT model that has been finetuned on predicting, as a binary output, the factual consistency of each summary sentence with the original document.
    \item SummaC (Summary Consistency) \cite{Laban2022summac} is somewhat similar to FactCC in conception, but is implemented as a Natural Language Inference (NLI) DNN \cite{Jeretic2020} that checks whether each summary-document sentence pair is consistent. Per-sentence scores can be combined into a single score per summary.
    \item CTCScore (Compression, Transduction and Creation Scores) \cite{Deng2021compression} is a suite of various metrics for evaluating different aspects of natural language generation, including a consistency aspect, based on pretrained DNN language models that are utilized for evaluating information alignment between different texts. The particular property of CTCScore is that, by design, it can be exploited for evaluating performance in many different generative NLP tasks.
\end{itemize}

\subsubsection{Extractive summarization}
\label{sssec:extractive}
This Subsection presents the gradual evolution of extractive summarization DNNs towards the current SoA.

\paragraph{Building Summaries Using Hierarchies} The first attempts to use DNNs for extractive summarization, while certainly obsolete today, developed core concepts that were later re-used and improved.

Thus, the method in \cite{lapata} developed a neural Encoder-Decoder architecture, which is trained in a supervised manner by preprocessing human-written summaries to construct extractive ground-truth summaries (binary selection vectors). Architecturally, the Encoder is composed of a shallow CNN, called "Convolutional Sentence Encoder", and a subsequent LSTM, called "Recurrent Document Encoder". The first one analyzes a matrix of Word2Vec word embeddings, separately for each sentence of the source text, aggregating them to dense sentence embeddings. The second one sequentially receives these embeddings as input and generates a document-level embedding in the form of a fixed-size vector.

The Decoder is one of two models following different, mutually exclusive strategies: a "Sentence Extractor", which approaches the extraction task as a sentence selection problem, and a "Word Extractor" which approaches it as a word generation task. The latter one is practically a limited-efficiency abstractive summarizer, with its supported vocabulary constrained to that of the source document, so it will not be detailed here. The Sentence Extractor is an LSTM-MLP combination, sequentially labeling individual source sentences for inclusion in the summary based on two criteria: i) whether they are relevant, and ii) whether they are mutually redundant with other sentences. The selection decision for the $t$-th sentence is computed as follows:
\begin{equation}
    \mathbf{\hat{h}_t} = LSTM(p_{t-1}\mathbf{s_{t-1}}, \mathbf{\hat{h}_{t-1}}),
\end{equation}
\begin{equation}
    p_t = \sigma(MLP(\mathbf{\hat{h}_t}; \mathbf{h_t})),
\end{equation}
\noindent where $\hat{\mathbf{h}}_t$/$\mathbf{h}_t$ is the Extractor's/Encoder LSTM's hidden state for the $t$-th sentence/time step, respectively. $p_{t-1}$ represents the assigned probability that the previous sentence $\mathbf{s}_{t-1}$ belongs to the summary, and $;$ is the concatenation operation.

% The datasets used for either of the Decoders are adapted from the CNN/Daily Mail dataset. The extractive dataset for the Sentence Encoder is generated by selecting the sentences which feature the most semantic similarity with the ground-truth summary. The dataset for the Word Extractor on the other hand, was generated by selecting the words with the highest lexical overlap between the highlights (ground-truth summary sentences) of the article with Out Of Vocabulary (OOV) words being replaced by the most semantically equivalent word in the article.

\begin{figure}
    \centering
    \includesvg[width=12cm]{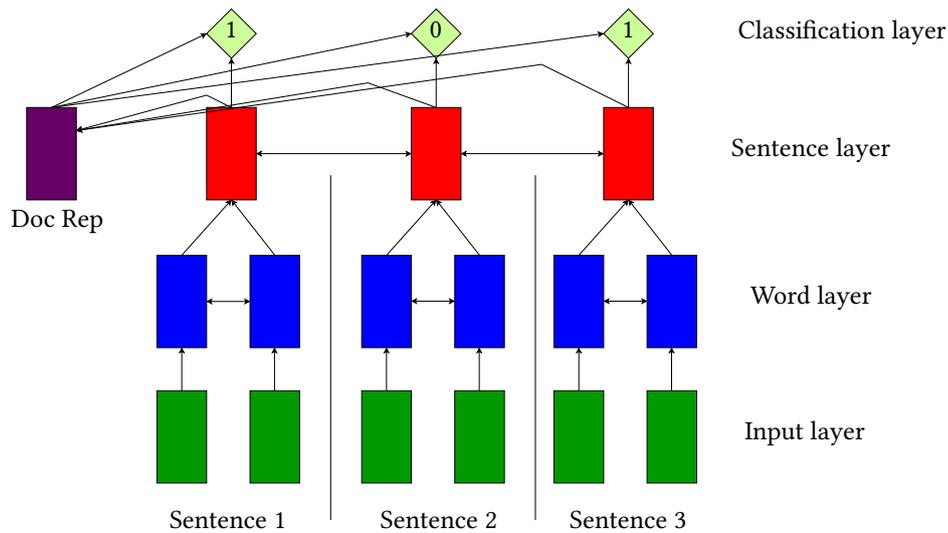}
    \caption{The SummaRuNNer architecture. Bidirectional RNNs are represented by double-pointed arrows.}
    \label{fig::SummaRunner}
\end{figure}

Following-up on \cite{lapata}, "SummaRuNNer" \cite{nallapati} was proposed as an extractive summarization RNN with the useful ability to be directly trainable on abstractive ground-truth summaries. It contains multiple layers of bidirectional GRUs and a hierarchical Encoder. Similarly to \cite{lapata}, the first subnetwork generates sentence embeddings from individual word representations, which are sequentially fed as input to the second GRU. The latter one analyzes them and outputs a document-level embedding from the sentence representations; this is done by average pooling the second subnetwork's hidden states. The Decoder is simply a logistic binary classifier, assigning to each source sentence a "summary"/"non-summary" label. To be directly trainable on datasets with abstractive ground-truth summaries, a separate RNN Decoder can be optionally attached to the Encoder \textit{only} during training and tasked to emit the most probable word at each time step. Figure \ref{fig::SummaRunner} depicts the SummaRuNNer architecture.

% \begin{equation}
%     d = tanh(\mathbf{W_{d}} \frac{1}{N} \sum^{N^{d}}_{j=1}[\mathbf{h^{f}_{j}};\mathbf{h^{b}_{j}}] + \mathbf{b})
% \end{equation}
% where $\mathbf{h^{f}_{j}}$ and $\mathbf{h^{b}_{j}}$ are the hidden states corresponding to the $j$th sentence of the forward and backward sentence-level RNNs respectively, $N^{d}$ is the number of sentences in the document and ‘;’ represents vector concatenation. The Decoder then classifies documents taking into account the salience, novelty as well as the absolute and relative positional embeddings of the content. % The model can be trained via extractive training, on datasets where the summary is a binary vector, where $1$ means a sentence needs to be included and $0$ that it should be skipped, and on abstractive summaries. This allows the model to train on many datasets which feature human-generated summaries without further pre-processing, which was the case in \cite{lapata}.% In the case of an extractive dataset the loss function is a standard CrossEntropyLoss function between the predicted and ground-truth binary summary vectors.

The method "LSTM-Ext" \cite{xiao} improves on \cite{lapata} and \cite{nallapati} by considering both global and local context information along with each sentence, in order to decide on its inclusion into the generated summary. Inspired by how humans read lengthy text, it was reasonably hypothesized that hierarchical information could benefit ATS. The "LSTM-minus" algorithm \cite{wang} is applied to generate local context-aware embeddings at the section level. Similarly to \cite{lapata} and \cite{nallapati}, an Encoder-Decoder architecture is employed. The Encoder is again composed of a Sentence Encoder (SE) and a Document Encoder (DE), but now also includes a "Topic Encoder" (TE) that provides access to local (section) context surrounding each sentence.

SE is a GRU that maps input word embeddings to a fixed-size sentence-level vector, while DE is a bidirectional GRU that encodes the sentence embeddings into a document-level vector representation. The final document embedding is computed as the concatenation of the final hidden states of the backward and the forward GRU. TE \cite{wang} separately embeds local context about each sentence's section by employing the LSTM-minus method; the result is an additional vector per section. Subsequently, for each sentence, the embeddings of itself, of its section and of the entire document are concatenated, optionally with weights assigned by an attention mechanism, and fed to the Decoder. The latter one is simply a binary MLP classifier that predicts whether each sentence should be included in the summary or not.

The architecture's efficiency seems to increase with increasing document length, making it a powerful architecture for long documents, while the most benefit comes from the use of local rather than global context. Thus, LSTM-Ext was among the first modern document summarization methods that explicitly handle the fact that longer texts tend to cover multiple topics. On the other hand, it is not designed to avoid unnecessary repetitions in the generated summary, resulting in outputs with potentially high redundancy.

\paragraph{Hierarchical Summaries using BERT} In recent years, Transformer and BERT variants gradually replaced RNN and Word2Vec/GloVe variants in extractive summarization. However, older ideas have been retained and improved, with their efficiency increased, in an attempt to resolve long-standing challenges of document summarization.\label{sssec: HierBERT}

The method in \cite{lapata_bert} uses BERT for both extractive and abstractive summarization (BERTSUM-EXT and BERTSUM-ABS, respectively). Instead of using sentence-level vector representations/embeddings, BERT inherently generates individual token representations. Thus, BERTSUM-EXT inserts special [CLS] tokens in-between sentences as separators, to notify the DNN of distinct sentences and capture sentence-level context within the respective generated token representations. This modified pretrained BERT is utilized as the Sentence Encoder component within an Encoder-Decoder architecture. The Document Encoder is a Transformer that receives the sentence-level embeddings of the [CLS] tokens that are derived by BERT and transforms them so as to contain richer document-level context. The final resulting sentence embeddings are fed to a shallow, fully connected binary classifier that predicts whether each sentence is to be included in the summary or not. All input documents are truncated to a fixed number of supported tokens. The overall architecture is finetuned in a supervised manner, using ground-truth binary selection vectors.

Despite the relatively strong performance of BERTSUM-EXT, it faces the typical challenges encountered by Transformers in long document analysis, such as varying and diverse topics. To counter this, the approach in \cite{histruct} extends BERTSUM-EXT by taking advantage of the usually strong hierarchical structure of long texts (individual sentences compose titled sections, which jointly form the document). Thus, it computes hierarchical positional embeddings that capture the relative position of a token within each section and the position of that section in the document. These representations, along with additional representations of the section titles, are incorporated into the sentence-level embeddings by the Document Encoder, which feeds the final shallow, fully connected binary classifier. Of course, the overall "HiStruct+" architecture is finetuned for extractive summarization. The method seems to perform well in long scientific articles with a rigid section structure, while still achieving SoA performance in less formal documents of short-to-medium length. Finally, by evaluating the method using pretrained Transformer language models other than BERT, the Longformer seems to come at the top.

Another approach aimed at long document extractive summarization is GBT-EXTSUM (Global BERT-based Transformer for Extractive Summarization) \cite{global_BERT}. It is an architecture composed of stacked propagation layers, each consisting of a Transformer layer, a BiGRU for inter-block information propagation and an MLP. GBT-EXTSUM achieves scalability to long document sequences without relying on attention mechanisms, by introducing a hierarchical structure with recurrent hierarchical modules and Bidirectional Gated Recurrent Units (BiGRU). This design allows for the propagation of information between text blocks at different levels, eliminating the need for attention mechanisms and providing an effective solution for processing lengthy documents. Furthermore, it eliminates the need for positional encoding on block representations, contributing to the efficiency of summarization tasks.

\paragraph{HyperGraph Transformers} HyperGraph Transformers are a recently proposed approach to long document extractive summarization, exploiting ideas from previous hierarchical methods and implementing them using Graph Neural Networks (GNNs).

The method in \cite{hegel} attempts to deal with the Transformer's quadratic complexity and context retention problems by developing a hybrid architecture called "HEGEL" (HypErGraph transformer for Extractive Long document summarization), relying on the so-called \textit{Hypergraph Transformer}. HEGEL models the document as a graph $G = (V, E)$, where $V$ is a set of nodes and $E$ a set of \textit{hyperedges}: an edge that can connect two \textit{or more} nodes. We use the notations $u \in e$ and $u \notin e$ to represent whether a node is connected to a hyperedge in the graph $G$ or not, respectively.

In HEGEL, the graph nodes are the document's sentences and extractive summarization is modelled as a node classification task. Each sentence is initially semantically embedded to a dense vector using a pretrained BERT. There are three kinds of hyperedges in this document graph:
\begin{itemize}
    \item A \textit{section hyperedge} connects all sentences that belong to the same section. For instance, in the context of a scientific article, a section could be the problem statement, the methodology or the conclusions. An incidence matrix $A^{sec}$ is defined as:

    \begin{equation}
     \mathbf{A}^{sec}_{i,j} =
      \begin{cases}
          1, \text{ if } s_{i} \in e^{sec}_{j} \\
          0, \text{ if } s_{i} \notin e^{sec}_{j},
      \end{cases}
    \end{equation}
    where $s_{i} \in e^{kw}_{j}$ means the $i$-th sentence belongs in the $j$-th section.

    \item A \textit{topic hyperedge} connects all sentences sharing the same topic. Topic clustering is implemented by using Latent Dirichlet Allocation \cite{dirichet}. The incidence matrix
    $\mathbf{A}^{topic}$ is defined as:
    
     \begin{equation}
     \mathbf{A}^{topic}_{i,j} =
      \begin{cases}
          1, \text{ if } s_{i} \in e^{topic}_{j} \\
          0, \text{ if } s_{i} \notin e^{topic}_{j},
      \end{cases}
    \end{equation}
    where $s_{i} \in e^{kw}_{j}$ means that the $i$-th sentence belongs to the $j$-th topic.

    \item A \textit{keyword hyperedge} connects all sentences containing the same keywords. These keywords are extracted using Key-BERT \cite{keybert} from the training sample. The incidence matrix $\mathbf{A}^{kw}$ is defined as:

    \begin{equation}
     \mathbf{A}^{topic}_{i,j} =
      \begin{cases}
          1, \text{ if } s_{i} \in e^{kw}_{j} \\
          0, \text{ if } s_{i} \notin e^{kw}_{j},
      \end{cases}
    \end{equation}
    where $s_{i} \in e^{kw}_{j}$ means that the $i$-th sentence contains the $j$-th keyword.
\end{itemize}

The overall incidence matrix $A \in \mathbb{R}^{n\times m}$ is generated by concatenating the three incidence matrices as:
\begin{equation}
    \mathbf{A} = \mathbf{A}_{sec};\mathbf{A}_{topic};\mathbf{A}_{kw}.
\end{equation}

HEGEL receives as its input both the sequence of initial sentence embeddings and the overall incidence matrix. It adopts the hierarchical positional embeddings from \cite{histruct} and defines an alternative "hypergraph attention" mechanism, along with the corresponding multihead hypergraph attention module, instead of regular self-attention heads. As in the vanilla Transformer, these components are contained within a deep neural architecture with standard fully connected layers, skip connections and layer normalization mechanisms. This Hypergraph Transformer extracts semantically rich sentence-level representations that are well aware of local context, global context and semantic interrelations between different parts of the document. The overall architecture is trained end-to-end for binary sentence classification using a ground-truth binary selection vector per document.

\begin{figure}
    \centering
    \includesvg[width=14cm]{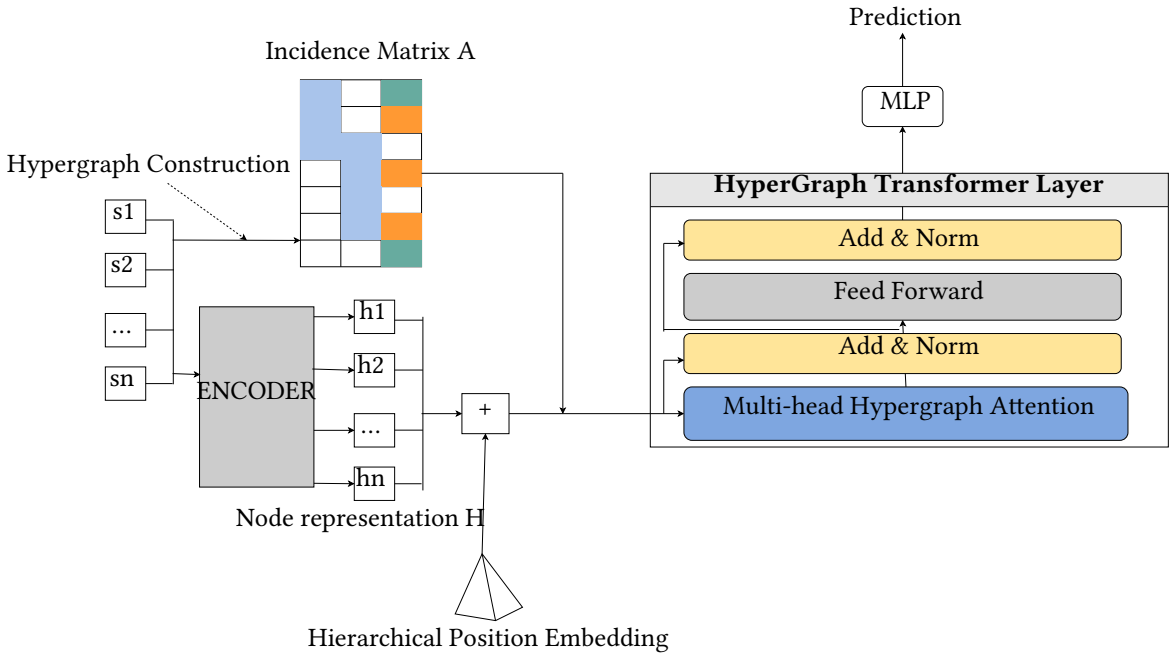}
    \caption{The overall HEGEL architecture.}
    \label{fig::HEGEL}
\end{figure}

HEGEL manages to outperform all extractive and abstractive baselines on several datasets of scientific articles, demonstrating the potential of GNNs for long document summarization. It remains to see if such architectures can retain this level of performance on less rigid and hierarchical document types. Figure \ref{fig::HEGEL} depicts schematically HEGEL's architecture.

\subsubsection{Abstractive Summarization}
\label{sssec::AbstractiveSum}
Abstractive summarization is typically approached as a sequence-to-sequence language modeling task, where a neural Decoder predicts the next word of the summary at each time step. The vast majority of modern relevant methods are trained in a supervised manner, using ground-truth summaries. This is arguably a more advanced task than its extractive alternative.

\paragraph{Pointer-Generator Networks} One of the first truly successful abstractive summarizers is essentially a hybrid one: a DNN that dynamically chooses whether to select a source sentence verbatim, as an extractive method, or synthesize the next word in the summary \cite{abigail}. Inspired by the human ability to refer to unknown objects by pointing at them, such Pointer-Generator Networks (PGNs) are an improvement over the "Word Extractor" proposed by \cite{lapata}.

PGN can copy out-of-vocabulary (OOV) words from the source document, making it possible to generate text with rare words and keep a smaller vocabulary, thus reducing computational and memory costs. Essentially, it is a modified, attention-augmented sequence-to-sequence Encoder-Decoder architecture implemented similarly to \cite{nallapati}, where both the Encoder and the Decoder are one-layer, bidirectional LSTMs. The Decoder ends at a softmax layer and at each time step generates a new word, originating either in the source document or in a global supported vocabulary. Thus, at each time step of the Encoder's/Decoder's unfolding, the next token is read/outputted from the input/to the generated summary, respectively. In order to decide on the course of action for the next word, the PGN defines its output distribution over the union of the global and the document's vocabulary (called the "extended vocabulary"), thus deciding at each time step whether it should copy or generate the next word. If a word is OOV, it will be copied from the source document by sampling from the attention distribution, while if it does not exist in the document a new word will be generated.

The attention mechanism is defined as in \cite{bahdanau}. During inference, a \textit{coverage vector} is updated at each Decoder time step, containing the sum of attention distributions over all previous time steps. It is provided as an additional input to the attention mechanism, so that the latter avoids giving attention to already extracted words and, thus, repeating them in the summary. To enhance this behaviour, PGN is also guided during training by an extra \textit{coverage loss}: a regularizer that penalizes repeated attention to the same input tokens. Thus, the overall loss function computed at the $t$-th Decoder's time step is:
\begin{equation}
L_{o}^t = L_{ML}^t + \lambda \sum_i min(a_i^t, c_i^t),
\end{equation}
\noindent where $\lambda$ is a scalar weighting hyperparameter, $\mathbf{a}^t$ is the current attention distribution, index $i$ traverses the Encoder's time steps, $\mathbf{c}^t$ is the current coverage vector, while $L_{ML}^t$ is the primary loss term at the $t$-th Decoder's time step. Typically, $L_{ML}^t$ is the negative log likelihood of the $t$-th ground-truth summary word according to the final Decoder output, i.e., a generated probability distribution defined over all supported words of the extended vocabulary. The overall loss is summed over all the Decoder's time steps.%Figure \ref{fig::PGN} depicts schematically the inference-stage operation of PGN.
% \begin{figure}
%     \centering
%     \includesvg[width=16cm]{images/pointer.svg}
%     \caption{Baseline sequence-to-sequence model with attention. The model may attend to relevant words in the source text to generate novel words, e.g., to produce the novel word beat in the abstractive summary Germany beat Argentina 2-0 the model may attend to the words victorious and win in the source text.}
%     \label{fig::PGN}
% \end{figure}

PGN does not use any pretrained word embeddings, learning instead token representations from scratch. At the time of its appearance it showed remarkable success in handling OOV words, while also featuring a smaller vocabulary size and faster training times. It thus successfully handled most of summarization challenges. However, although historically significant, PGN cannot handle long documents well. As a result, more efficient and specialized architectures were later developed.

\paragraph{BERT \& Sparse Attention Transformers}  As described in Subsection \ref{sssec:extractive}, BERT has been adapted for extractive summarization in the form of BERTSUM-EXT \cite{lapata_bert}. That neural architecture has also been employed for abstractive summarization in the form of BERTSUM-ABS.

The Encoder of BERTSUM-ABS is a pretrained BERTSUM-EXT model, a choice which provides efficiency gains, due to transfer learning, and task-specific gains, due to emerging synergies between extractive and abstractive summarization \cite{wei2, gehrmann}. The Decoder is a regular, randomly initialized 6-layer Transformer. Different finetuning and training schedules are utilized for the Encoder and the Decoder, respectively, to avoid stability issues during optimization. Overall, the approach attained only marginal performance gains over competitors.

Section \ref{ssec::ClassificationSolutions} reviewed the issues vanilla Transformers face with very long documents, with the biggest challenge being the quadratic cost of the self-attention mechanism. The various sparse attention solutions that attempt to bypass this restriction can also be employed for document summarization, by replacing a regular Encoder with a Sparse Transformer variant. One such architecture is the "Longformer-Encoder-Decoder" (LED), i.e., where Longformers are utilized both for the Encoder and the Decoder. Critically, the Decoder pays full attention to the entire Encoder and to already decoded positions of the input. As expected, LED scales linearly with the input size, and has been shown to achieve highly competitive performance.

Similarly, BigBird has also been employed for abstractive document summarization \cite{big_bird}. The sparse attention mechanism is used only in the Encoder, which handles the long input, while the Decoder exploits full attention, as the length of the summary (in tokens) is small by definition. BigBird achieved significant performance gains on long document summarization and competitive performance on smaller ones.

BigBird makes use of Gap-Sentences Generation (GSG) for abstractive summarization \cite{pegasus}. It is a self-supervised pretraining objective especially constructed to aid downstream finetuning for text summarization. Essentially being a variant of T5's pretraining objective, it consists in masking an entire sentence with a single [MASK] token. Then, during pretraining, each input is a set of consecutive sentences, with a subset of them selected and hidden according to GSG. The DNN is tasked to generate the masked sentences jointly, in the form of a pseudo-summary, based on the seen ones. The selection of which input sentences to mask is not random, but based on the ROUGE score of each sentence with the remaining input text. Thus, the ones  finally hidden with [MASK] tokens are those that have been automatically judged as the most important ones. It must be noted that GSG was originally introduced as the main novelty of the PEGASUS neural language model, following a Transformer Encoder-Decoder architecture, like BART or T5.

GSG and a variant of the sparse attention mechanism is also employed for abstractive document summarization in LongT5, which is a modification of the basic T5 neural architecture \cite{Guo22LongT5}. Thus, the LongT5 Encoder contains self-attention sublayers where each input token attends to its local neighbourhood plus to all global tokens. The latter ones are being dynamically constructed before each attention operation, by summing and normalizing different, non overlapping subsets of the input token embeddings. Despite the approximate nature of this sparse attention mechanism, the resulting ability to process longer inputs leads to overall improved accuracy in long document summarization.

Another variant of the Sparse Attention Transformer mechanism is the EMMA Memory-Augmented Transformer Summarizer \cite{luca}, which was developed for low resource systems and uses text segmentation in order to better process the text in chunks. It then feeds the data into a modified version of BART which utilizes a long-term memory matrix for each layer, combining the previous and current memory matrices in order to avoid losing long-term details. This modified BART model also features cross-memory attention in order to consider information from previous chunks when processing the current chunk. This results in the EMMA model generating more contextually coherent summaries, while retaining its low space complexity of $\mathcal{O}(L^2_c) $ (where $ L_c $ is the predetermined max chunk size).

\paragraph{Local Attention and Content Selection} The majority of modern abstractive summarizers learn to implicitly select pertinent content from the input in order to generate the summary. Recently, attempts have been made to combine the advantages of Sparse Attention Transformers, which allow the handling of longer documents, with explicit content selection before the abstractive summary generation \cite{gales}.

The main abstractive DNN in \cite{gales} is a modified version of BART \cite{BART}, called LoBART. Vanilla BART supports input sequences with a length up to 1024 tokens. LoBART modifies its layers so as to employ the sparse attention scheme of Longformer, with the difference that only local and no global attention is utilized. This allows LoBART to drop its memory requirements significantly, at the cost of missing context over large spans of text. The reduced costs result in a sliding window of $W=700$ tokens, which is calculated as $2 \times D$ where $D$ is the empirical average attention distance of $D \in [250, 350]$ tokens for an attention mechanism with uniform weights, attending to a distance of $1024$. Even with this compromise, important context may still be missed, especially in long documents.

As a result, the method in \cite{gales} feeds LoBART with suitably preprocessed input sequences, so as to effectively increase its context span. This preprocessing takes place separately during the training and the test stage. At training, the input data are first passed through a content selection algorithm. This can either be an ``oracle" (ORC), which selects sentences based on the ground-truth (the ROUGE score of the selected input vs its abstract), or a hierarchical, extractive DNN pretrained for multitask content selection (MCS). The straightforward "ORC" preselection keeps an average of 21.3\% of sentences, drastically reducing the input's size and leads to 56\% of the input documents being shorter than the typical BART input length of 1024. Missing sentences are padded with random sentences sampled from the input document. On the other hand, MCS is a GRU-based, attention-augmented Encoder-Decoder architecture followed by a classification layer. It employs a hierarchical design similar to that of \cite{liu} (see Section \ref{ssec::ClassificationSolutions}), but in this case it is only utilized to preprocess the input document. In order to exploit synergies between extractive and abstractive summarization, similarly to \cite{lapata_bert}, MCS is jointly pretrained (in a multitask fashion) both for predicting the ground-truth summary token-by-token and for binary sentence classification as "summary" or "non-summary".

During inference, each sentence is first scored for retention by suitably combining the two different predictions (one per task) made by MCS. The top-scoring ones are fed to LoBART as the input sequence. LoBART has been trained for abstractive summarization in the usual manner, but the two-stage content preselection stage (explicit during preprocessing, implicit during the main summarization process) allows the method to efficiently handle longer documents.

\paragraph{Combining Top-Down and Bottom-Up Inference} An innovative abstractive methodology relying on a composite Transformer architecture is proposed in \cite{pan_bottom}, aiming to facilitate the sharing of global and local context between different token representations. This is done by allowing information to flow both from the individual token representations to the document embedding (bottom-up) and vice versa (top-down).

The Encoder in \cite{pan_bottom} features a hierarchical approach similar to certain ones utilized for extractive summarization (e.g., \cite{xiao}): individual token representations computed by an initial (bottom-level) deep Transformer are pooled in segments to form top-level embeddings of coarse granularity at a paragraph/section-level. This is the so-called "bottom-up" route. The preliminary top-level embeddings (one per section) are then transformed via a separate, shallow Transformer. However, the main novelty of \cite{pan_bottom} is that these final top-level representations are then injected to the individual token embeddings via a separate, subsequent deep"top-down" Transformer. The latter's layers are equipped with regular self-attention, as well as with cross-attention between the top-level representations and the individual token embeddings. The final output token representations are attended by the Decoder to generate the desired abstractive summary in the usual manner.

By properly adapting Sparse Attention Transformers, the bottom-level and the top-down layers are assigned efficient local attention mechanisms only. Top-level layers can afford to be assigned full attention due to the much smaller number of sections, compared to the number of tokens. Thus, the top-down information flow enriches the lower-level embeddings with global context and the ability to capture long-range dependencies: individual token representations effectively attend to aggregated global information from the entire document, despite the local attention-only mechanisms of bottom-level layers.

The methodology is rather generic and model-agnostic, allowing it to be used with efficient Sparse Attention Transformers. This fact and the hierarchical design facilitate exceptional summarization performance in long documents, even at book-length texts. Exploring its applicability to other tasks, such as document classification, is a promising research direction.

\paragraph{Instruction-tuned summarization} An orthogonal direction of recent research is that of supervised finetuning of pretrained language models on ``instruction" and corresponding ground-truth summary pairs. The instruction is a command to summarize in natural language, followed by the long text to be summarized. The goal is to then use, during inference, the finetuned DNN like one would use a pretrained LLM under a zero/few-shot prompting mode. This can be done both with Encoder-Decoder Transformer architectures pretrained on masked token objectives and with LLMs, which typically are Decoder-only Transformer architectures pretrained on traditional next token prediction objectives. Both T5-based architectures \cite{victor2022multitask} \cite{chung2022scaling} and LLM-based architectures \cite{ouyang2022training} \cite{iyer2022opt} \cite{muennighoff2022crosslingual} have recently followed this approach, typically under a multitask instruction tuning setting where text summarization is one among multiple tasks. According to common automated evaluation metrics, this approach typically leads to subpar summarization results in common public benchmarks, compared to SoA dedicated summarization DNNs. However, such summaries generated from instruction-tuned LLMs can actually be better than ones obtained from dedicated summarizers and even surpass the ground-truth ones in quality, according to manual, subjective evaluation \cite{goyal2022news}. Although instruction-tuned summarizers tend to perform a lot worse with long text inputs, there have been recent breakthroughs such as the Longform dataset \cite{Longform}, constructed via diverse instruction generation methods; finetuning LLMs on Longform has recently led to good results on various downstream tasks. Overall, this is a still an open area of cutting-edge research, where progress in automated evaluation metrics is a crucial prerequisite for further advances (as discussed in Subsection \ref{sss::SummarizationEvaluation}).

\paragraph{Dealing with hallucinations} A few dedicated methods have appeared in the abstractive summarization literature, concerned with how to reduce the appearance of hallucinations in the generated summary. Thus, these methods are not summarizers themselves, but can be employed in conjunction with an abstractive summarization DNN in order to improve its output.

There are two main families of hallucination containment methods: \textit{post-processing} approaches and \textit{faithfulness-aware training}. In the first case, a separate DNN is appended after the summarizer and evaluates the latter one's outputs. An important such method is \cite{Liu2021SimCLS}, where a sequence-to-sequence summarizer is followed by a scoring RoBERTa model. The summarizer outputs multiple candidate summaries and the scoring model is trained to rank the candidates according to their ROUGE evaluation score with regard to the ground-truth summary, using contrastive learning. Thus, the two models are being trained separately in a supervised manner, while during the test phase the scoring model predicts quality scores without access to ground-truth. The candidate summary with the highest score is finally selected. An alternative post-processing approach \cite{Ladhak2022} models the issue as a trade-off between faithfulness to the original document and abstractiveness (in contrast to extraction), and then attempts to generate a summary that transcends said trade-off by being both faithful and abstractive. This is achieved by training a follow-up selector DNN to explicitly choose among a set of candidate summaries, at different levels of extractiveness, the one that is both the most abstractive and faithful.

In contrast to such approaches, faithfulness-aware training trains the summarizer itself with additional objectives for hallucination containment. Several methods have appeared in this direction, for example focusing on filtering the training data \cite{nan2021entity} \cite{goyal2021annotating}. A more recent method \cite{Wan2022Fact} modifies the GSG pretraining objective of the PEGASUS architecture \cite{pegasus}, where the employed ROUGE score is combined with the factuality metric FactCC \cite{Kry2020}. Downstream finetuning for summarization is also modified in three complementary ways: i) the [MASK] token from the GSG pretraining objective is inserted into the downstream training document as well, so that factuality knowledge acquired during pretraining is better retrained, ii) hallucinations are explicitly removed from the ground-truth summaries of the downstream dataset, and iii) an additional finetuning objective, borrowed from the contrastive self-supervised learning literature \cite{Chen2020simple}, encourages the summarizer to internally generate summary representations that are close to the embedding of the original document and away from those of artificially constructed non-factual summaries.

An alternative method for faithfulness-aware training \cite{zhang2022improving} explicitly quantifies faithfulness as the ratio of the summary's named entities derived from the original document. This is measured for all training documents, with respect to the corresponding ground-truth summaries, and converted to a suitable code representation which is given to the summarizer as a conditioning mechanism during training. During inference, the code denoting high faithfulness is given as input along with the test document. The method in \cite{Xiao2022entity} also focuses on the consistency of named entities, attempting to eliminate hallucinated entities that do not appear in the original document, while retaining ones that do appear in the ground-truth summary. This is achieved by incorporating the proposed ``SpanCopy" mechanism into the summarizer's Decoder, assuming an Encoder-Decoder architecture. SpanCopy directly copies in the output entities from the full-length source document and its activation is controlled by a final classifier that decides whether the next span of text will be copied or generated in the usual manner; thus, the method bears similarities to the PGN architecture, but operates the named entity level. The entire DNN is trained end-to-end in a supervised manner.

A different aspect has been explored in \cite{Wan2023faithfulness}, where alternative strategies by which the Decoder decides on the exact next output token during summary generation are compared with regard to their effect on hallucinations. There are various such decoding strategies, beyond the default greedy selection of the most probable output token, which are commonly utilized in text generation tasks. The most widespread alternatives to greedy selection are \textit{nucleus sampling} \cite{holtzman2019curious} and \textit{beam search}, with the latter one identified in \cite{Wan2023faithfulness} as the one that generates the most faithful summaries. The authors modify it to more explicitly consider hallucinations containment in two ways: i) by internal re-ranking of the candidates, based on summarization evaluation metrics that reward faithfulness, and ii) by looking ahead into the future contingent summary that can be generated based on the current partial summary and scoring it for faithfulness; this added heuristic is incorporated into the search process.

Overall, dealing with hallucinations in abstractive summaries is a currently very active research topic that overlaps with factuality/constistency checking and NLI.

\subsection{Extractive or Abstractive Summarization?}
SoA summarization models for long texts are divided into 2 categories, abstractive and extractive summarizers, each with their own strengths and weaknesses.
\begin{itemize}
    \item Extractive summarization DNNs tend to be more simple, factually consistent and have a much smaller risk of introducing errors, since the extracted sentences come directly from the source text. However, they lack creativity and may suffer from redundant, repeated information.
    \item Abstractive summarization DNNs solve the problems that extractive summarizers present, generating more creative and diverse summaries without significant repetitions, and are also able to condense information more efficiently\cite{koh}. However, they suffer from hallucinations and lack of factual consistency, introducing a much bigger risk of containing inaccurate information. As a result, it is more complex and difficult to properly utilise them in practical situations.
\end{itemize}
Generally, both approaches have their strengths and SoA architectures can employ both to achieve good results. For instance, certain recent approaches for fighting hallucinations in the abstractive case essentially attempt to force the DNN to become more extractive. Extractive summarization DNNs seem to outperform abstractive ones in relevancy and informativeness, however longer text input size seems to narrow that gap down. Hybrid models seem to be the most promising avenue for long text summarization in most tasks where the human-like element that abstractive summarizers provide is important. A thorough review and comparison of the two types of architectures specifically for long texts can be found in \cite{koh}.

\section{Sentiment Analysis}
\label{sec::Sentiment}
Opinion mining (OM) or sentiment analysis (SA) is the extraction of \textit{polarity sentiment} (positive/negative) from a piece of text using NLP algorithms \cite{pakistan, sergio}. It is typically approached as a particular case of text classification, although sentiment score regression is also common \cite{Karamouzas2022}. Thus, possible opinions are modelled either as a set of discrete classes (e.g., "positive", "neutral", "negative"), or as a continuous range of sentiment scores (e.g., in the $[-1, 1]$ range). Additionally, emotional dimensions besides polarity are occasionally considered (e.g., classes such as "fear", "joy", "sadness", etc.), potentially by relying on psychological theory (e.g., \cite{Plutchik2001}). Common use-cases mainly concern SA of short texts, such as social media posts, product reviews and online discussions, with obvious applications in marketing and brand management \cite{sa_example_1, sa_example_2, sa_example_3} or political analysis \cite{sa_example_4, sa_example_5, karamouzas2022SNAM}. For an in-depth review of SA applications, the reader is referred to \cite{sa_applications}. 

SA for long texts is not the most common task, since context and tone can change throughout a lengthy passage, making it difficult to maintain a constant sentiment. Moreover, the presence of complex language structures and nuanced expressions in long texts can pose challenges for SA models. Additionally, in certain domains SA is meaningless due to the typical lack of emotionally charged words (e.g., in academic articles, medical or legal documents, etc.). However, it is important for automatic analysis of long journalistic articles, to extract their political viewpoint, financial texts, to predict how markets will react to the presented news, or literary books, in order to process narrative hints. For example, it can be used to determine characters' emotional states, or for identifying key narrative points, sections and anticipated reader feelings \cite{omori}. These predictions are potentially useful in themselves, but they can also be exploited as auxiliary input for particular book classifications tasks (e.g., automatic genre recognition in library software).

\subsection{Solutions} 
SA has peculiarities that distinguish it from vanilla document classification. Traditional SA algorithms rely on ``sentiment lexicons", i.e., predefined vocabularies of words that are heavily weighed towards positive or negative emotions. For example, words such as ``good", ``bad", ``great", ``terrible", etc. are likely candidates for inclusion in such as lexicon \cite{shelly}. Algorithms relying on lexicons and/or traditional classifiers (e.g., Naive Bayes, SVMs, ensembles \cite{onan_2017_sa}, etc.) were common before the advent of DNNs \cite{pang2} \cite{pang}. However, texts with complicated vocabulary often lack ``simple" keywords that are included in most lexicons. In such cases, most lexicon-based algorithms struggle to classify accurately.

As a result, long document SA has moved decisively towards DNNs in the past few years. The main body of relevant research approaches the field as simply an instance of document classification, so the content of Section \ref{sec::Classification} will not be repeated here. The relevant challenges are common or analogous to those found in most long document analysis tasks, such as prohibitive computational costs, difficulty to process long input sequences, long-range dependencies that are hard to identify and sentiment variation across different sections or paragraphs.

For example, \cite{onan2022_sa} utilizes a BiLSTM with an embedding layer and a group-wise enhancement mechanism, which is an alternative to a ``traditional" attention mechanism \cite{liu_2019_attention}. The group-wise enhancement mechanism adjusts the significance of sub-features within specific spatial and semantic groups in the text. It employs attention factors for each spatial position in semantic groups, allowing independent enhancement of learned expressions and suppression of noise. Inspired by \cite{li2017}, it utilizes feature weighting schemes on LSTM and GRU to extract informative features. Its goal is to leverage general information from critical regions, approximating the semantic vector for each group. It does so by dividing feature groups, obtaining global statistical features through spatial averaging, and scaling context vectors based on generated importance coefficients. This approach aims to refine representation and discriminative capabilities in DNNs, particularly concerning uninformative features.

SA can greatly benefit from sarcasm/figurative language (S/FL) detection \cite{onan_2019_sarcasm}, to a degree much greater than vanilla document classification. Neural S/FL detection has mostly been conducted using Bi-LSTMs, GRU or CNNs for short-to-medium length texts \cite{onan_sarcasm_survey, onan_2021}, therefore integrating such approaches into document/book-level analysis is an open challenge. An interesting avenue would be to exploit related short text semantic detectors of high specificity (e.g., S/FL detection) at the sentence-level, given that such tools are mature \cite{dmitry} \cite{shelly} and are indeed being used to aid short text SA \cite{Karamouzas2022}, aggregating their results in a hierarchical fashion to facilitate document-level SA. However, this research direction has not been significantly exploited.

Existing deep neural methods for long text SA that offer a particular angle, compared to simple document classification, are reviewed below. The majority of these and other similar methods have not been evaluated on actually long documents, due to the lack of relevant public annotated datasets, but they can in principle support them. Overall, however, long document SA is more of a near-future vision, rather than a currently blooming field.

\subsubsection{Hybrid architectures} 
Initial efforts tackled documents that were a few paragraphs long, instead of book-length. For example, a composite, hierarchical deep neural architecture is proposed in \cite{tang} for classifying long user reviews (e.g., of movies, restaurants, etc.). In order to handle polysemy, the method exploits the "principle of compostitionality" \cite{francis}, which states that the meaning of a body of text (a sentence, a paragraph or a document) depends on the meanings of its constituents, by using an LSTM to generate sentence-level embeddings for the entire document. These are sequentially fed to a subsequent GRU that encodes their relations and semantics, outputting a document representation that is classified by a final softmax layer. The GRU was explicitly selected due to its virtual lack of an output gate, since it was deemed necessary for all available sentence-level information to be used in the construction of the document representation. The overall architecture is trained end-to-end for classification and receives dense Word2Vec embeddings at its input layer.

A similar method \cite{morocco} combines a CNN and a bidirectional LSTM to extract the sentiment of newspaper articles. It uses Doc2Vec \cite{doc2vec} to pre-generate paragraph-level embeddings that form the input matrix for the CNN. An alternative hierarchical method is presented in \cite{mao2022document}, where an initial bidirectional LSTM sequentially receives word embeddings per sentence and transforms them to sentence embeddings. A second bidirectional LSTM equipped with an attention mechanism receives the latter representations to generate a document-level embedding in the form of a matrix, which is then fed to a final 2D CNN followed by the softmax classifier layer. As a general statement, it can be said that such straightforward hierarchical approaches (e.g., \cite{bhuvaneshwari2022sentiment}) have become a de facto baseline in long text SA, with the recurrence of LSTMs typically exploited to handle the variable and large number of tokens, sentences and paragraphs in long documents.

\subsubsection{Sentiment Analysis with Transformers}
SoA models make use of Transformers, with BERT and its variations being very commonly used for SA \cite{toutanova}. The employed pretrained language model is typically finetuned on the desired SA dataset. For instance, this is how RoBERTa is utilized in \cite{xinzhao} for SA of financial texts. However, the document-level sentiment classification task is implemented in an ensemble learning fashion, using weighted joint prediction from multiple independent classifiers, and then combined with a separate pipeline for \textit{key entity detection}, so that specific entities (e.g., companies, countries, etc.) are identified in the text. Each of these entities is individually fed to an instance of an appropriately finetuned RoBERTA model (different than the SA ensemble participants) along with the document, in order for the DNN to predict whether each recognized entity is critical for its meaning.

\section{Public Long Document Datasets}
\label{sec::Datasets}
Long text analysis is not a well-explored field, particularly for document-level analysis of book-length texts (e.g., literary fiction). As a result, compared to small text analysis, the amount of relevant public, annotated datasets remains fairly limited. Furthermore, there is a clear imbalance with regards to the different NLP tasks; document summarization datasets dominate the field, while long document datasets for SA are nearly non-existent.

Traditionally, concerning document summarization, the most commonly used dataset is \textbf{CNN/DailyMail} \cite{nallapati2}, featuring over 330k articles and their respective highlights, as written by journalists at CNN and the Daily Mail. On average, each article has $781$ input tokens and $56$ summary (target) tokens. While the dataset is certainly large and ships with convenient APIs \cite{tf_cnn_dailymail, kaggle_cnn_dailymail, hugging_face_cnn_dailymail}, its individual articles are not be long enough for modern long document summarization. This has not prevented its continuing use as a benchmark, for measuring the efficiency of contemporary summarizers in short-to-medium length documents \cite{big_bird}.

Two of the most prominent long document datasets at the time of writing are the \textbf{arXiv} \cite{clement} \cite{gong} and \textbf{PubMed} datasets. Both are composed of scientific research articles, and are used mainly for summarization. They feature a huge number of documents, remarkable document length and user-curated summaries in the form of article abstracts. Their success is evident from their inclusion in many popular machine learning libraries \cite{tf_datasets} \cite{kaggle_arxiv} \cite{hugging_face_arxiv} \cite{hugging_face_pubmed} and from their popularity as benchmarks in current document summarization literature. Besides these ``main" datasets, many variations have been made available. For example SumPubMed \cite{sumpubmed} is a pre-processed, curated variant of the Pubmed dataset and which aims at making summarization more challenging.

%Yet, a few generic NLP datasets do include long documents. One of the most historically important ones is \textbf{OntoNotes} \cite{ontonotes}, which features a variety of texts and contains a number of rather long ones. OntoNotes is mainly annotated for syntax analysis tasks. Many other general-use NLP datasets 

%An important one is the \textbf{Columbia Quoted Speech Attribution} corpus \cite{elson}, which contains English literature texts from the 19th and 20th centuries that have been annotated for quotation attribution. Although it targets quotation extraction and speaker identification tasks, its data can still be utilised for document-level analysis.

Certain datasets are notable for their focus on literary texts, rather than scientific articles which are the subject of most commonly used long-document datasets. An example would be the Project Dialogism Novel Corpus (PDNC dataset) \cite{vishnubhotla2022project}, which comprises 22 English novels. A modern large-scale dataset for literary fiction texts is also \textbf{LitBank} \cite{bamman}, a dataset of 100 English literature works that fully includes word annotations and co-reference relations between words that belong in different groups.

Recently, two more challenging datasets have been released: \textbf{BookSum} \cite{poland} and \textbf{SummScreen} \cite{summset}. BookSum covers fiction books from multiple domains in literature, including short stories, novels and plays. SummScreen includes pairs of movie transcripts and respective human-written summaries. They both require the summarizer to combine narrative plot events from implicit subplots, potentially progressing in parallel, forcing it to rely on extremely long context spans and dependencies.

%The large-scale \textbf{CSL} dataset \cite{li}, developed in 2022, includes a plethora of Chinese scientific literature books. Modern neural language models trained on this dataset (e.g., BERT) achieved really high accuracy in tasks such as document classification for a language as complicated as Mandarin.

Other notable datasets include a subset of long texts, such as the \textbf{OPUS} dataset for machine translation \cite{tiedemann}, while domain-specific datasets are crucial for domain-specific NLP algorithms, either for training or for domain finetuning \cite{dai}. Below is a brief list of such datasets, featuring primarily long documents for classification or summarization tasks:
\begin{itemize}
    \item \textbf{LexGLUE} \cite{glue_gunner}, a classification dataset containing a collection of smaller datasets of legal texts.
    
    %\item \textbf{MIMIC-III} \cite{mimic} (Medical Information Mart for Intensive Care), which contains Intensive Care Unit (ICU) discharge summaries. This dataset has the benefit of being annotated with labels following the ICD-9 (The International Classification of Diseases, Ninth Revision) hierarchy.

    %\item \textbf{ECtHR-ARG} \cite{habernal}, which contains legal texts (court decisions) from the European Court of Human Rights, along with annotation for \textit{argument mining}. However this dataset only contains about 300 cases.

    \item \textbf{ContractNLI} \cite{koreeda}, a classification dataset for document-level Natural Language Inference (NLI). Each data point is a contract text and a set of hypotheses, with the three classes indicating whether each hypothesis agrees with, contradicts or is not mentioned by the contract.

    \item \textbf{20 Newsgroup} \cite{20groups}, a dataset containing approximately 20,000 Usenet newsgroup documents partitioned evenly across 20 different newsgroups. This dataset provides a great amount and variety of data, but its articles tend to be significantly shorter than dedicated long document datasets.

    \item \textbf{Hyperpartisan News Detection} \cite{hyperpartisan}, which contains documents for political standpoint classification. This dataset contains both long and short articles, therefore a split must be performed based on the text's length for long document analysis.
\end{itemize}

The above-mentioned datasets are summarized in Table \ref{tab::Dataset}. The posted dataset statistics have been either made available on their original citation, computed for the purposes of this article, or mentioned in other, properly cited papers. Given that these datasets are of gigantic size, utilize different formats and present drifts due to occasional expansions, certain sources may conflict on the exact size of these datasets.

\begin{table}
    \begin{tabular}
        { |p{2cm}|p{2cm}|p{4cm}|p{2cm}|p{2.5cm}|p{1.5cm}| }
        \hline
        \cellcolor{blue!25}\textbf{Name} & \cellcolor{blue!25}\textbf{Domain} & \cellcolor{blue!25}\textbf{Tasks} & \cellcolor{blue!25}\textbf{Documents} & \cellcolor{blue!25}\textbf{Average Document Size (words)} & \cellcolor{blue!25}\textbf{Language} \\
        \hline
        CNN/Daily Mail (benchmark) \cite{nallapati2} & News Articles & Document Summarization & 300K &  781 & English \\
        \hline
        %OntoNotes \cite{ontonotes} & General & Named Entity Recognition, Coreference Resolution, Semantic Role Labeling & 10k-100k & - (English v4 + v12 splits) & English, Arabic, Chinese \\
        %\hline
        Litbank \cite{bamman} & Literature  & Named Entity Recognition & 100 & 2,000 & English \\
        \hline
        PDNC \cite{vishnubhotla2022project} & Literature (Old) & Quote Attribution, Speaker identification & 35,978 & 79,745 & English \\
        \hline
        %CSL \cite{li} & Academic & Text Classification, Document Summarization, Keyword Generation (KG) & 396k & - & Chinese \\
        %\hline
        arXiv \cite{clement} & Academic & Document Summarization, Text Retrieval & 1.7M & 8,029.19 \cite{arxiv_size} & English \\
        \hline
        Pubmed & Academic & Language Modeling, Document Classification, Document Summarization & 19,717 & 4,000 \cite{sumpubmed} & English \\
        \hline
        SubPubPubmed \cite{sumpubmed} & Academic & Document Summarization & 33,772 & 4,227/1,578/1,891 depending on version & English \\
        \hline
        BookSum \cite{poland} & Literature & Text (Narrative) Summarization & 405 & 112,885.15 (BookSum Full) & English \\
        \hline
        SummScreen \cite{summset} & Screenplays & Document Summarization & 22,503 & 7,605.4 (FD subset) / 6,420.7 (TMS subset)   & English \\
        \hline
        LexGLUE \cite{glue_gunner} (only classification subsets) & Legal & Multiclass Document Classification & 184,214 & 4,761.9 & English \\
        \hline
        %MIMIC-III \cite{mimic} & Medical & Multiclass Document Classification & 112K & 709.3 & English \\
        %\hline
        %ECTHR-ARG & Legal & Multiclass Document Classification & 11K & - & English \\
        %\hline
        ContractNLI \cite{koreeda} & Legal & Multiclass Document Classification, Natural Language Inference & 607 & 2,254.3 & English \\
        \hline
        20 Newsgroup \cite{20groups} & News & Multiclass Document Classification & 50,139 & 2,036.3 & English \\
        \hline
        Hyperpartisan \cite{hyperpartisan} & News & Document Classification & 750,645 & 4,848.17 & English \\
        \hline
    \end{tabular}
    \caption{Reviewed long document datasets and their properties, compared to the well-known and popular CNN-DailyMail dataset of short texts.}
    \label{tab::Dataset}
\end{table}

\section{Conclusions}
\label{sec::Conclusions}
\begin{table}
    \scriptsize
    \begin{tabular}
         { |p{3cm}|p{3cm}|p{9cm}|  }
         \hline
         \cellcolor{blue!25}\textbf{Method} & \cellcolor{blue!25}\textbf{Addressed Issues} & \cellcolor{blue!25}\textbf{Details}\\
        \hline \rowcolor{lightgray}
         \multicolumn{3}{|c|}{Metrics \& Preprocessing} \\
         \hline
         Weighted Multi-label Classification \cite{sicong}  & Varying Count of Labels Per Document, Class Imbalance & By using multi-label metrics and loss functions using weighting, effective multi-label multi-class classifiers can be trained and accurately evaluated.\\
         \hline
         Feature Pruning \cite{sicong, worsham} & Dataset Size \break \break Computational Cost & Ruthless pruning and careful selection / sampling can significantly cut down on the computational/memory costs of long documents. On the other hand, even words traditionally considered semantically useless can be significant for a model to understand subtle grammatical and syntactical rules.\\
         \hline
        \rowcolor{lightgray}
         \multicolumn{3}{|c|}{Sparse Attention Transformers} \\
         \hline
          Sparse Transformers \cite{child} & $\mathcal{O}(n^2)$ Transformer Cost & By utilizing a sparse attention mechanism, and with various mathematical optimizations, a standard Transformer can become viable at parsing entire long document inputs.\\
        \hline
         Dilated Sliding Window (Longformer) \cite{longformer} & $\mathcal{O}(n^2)$ Transformer Cost \break\break Context Fragmentation & Using local self-attention with a dilated sliding window, while also allowing a limited number of global tokens, the Longformer becomes capable of successfully parsing ever-increasing-length documents, while keeping track of local context.\\
         \hline
         Sliding Window + Random Selection + Global Attention (BigBird) \cite{big_bird} & $\mathcal{O}(n^2)$ Transformer Cost \break\break Context Fragmentation & Combining Longformer's sliding window, block attention scheme and random token selection can encapsulate local and global context more efficiently while keeping computational costs low.\\
         \hline
        \rowcolor{lightgray}
         \multicolumn{3}{|c|}{Hierarchical Transformers} \\
         \hline
         Hierarchical Transformers combined with traditional networks \cite{ion_han, zichao, khandve, use} & Transformer Input Limit & Instead of using Transformers for the entire document, they can be used to produce segment embeddings, which are then combined by traditional networks (HAN, LSTM, CNN).\\
         \hline
         Hierarchical Transformers with BERT (Hi-Transformer) \cite{qi} &
         Global Context \break\break Context Fragmentation & BERT can be used to produce sentence embeddings, then document embeddings which can be fed to new, context-aware sentence embeddings iteratively.\\
         \hline
         Hierarchical Sparse Transformers (ERNIE-Sparse) \cite{liu} & $\mathcal{O}(n^2)$ Transformer Cost \break\break Local/Global Context issues & By using a hierarchical architecture embedded in a Sparse Transformer, ERNIE-SPARSE imparts token representations with additional global context.\\
         \hline
        \rowcolor{lightgray}
         \multicolumn{3}{|c|}{Recurrent Transformers} \\
         \hline
         Recurrent Transformers \cite{dai} & $\mathcal{O}(n^2)$ Transformer Cost \break\break Context Fragmentation \break\break Global Context & By keeping the context of previously inputted segments in hidden states and by utilizing relative positional embeddings, the Recurrent Transformer effectively supports much longer input sequences during inference, irrespective of training-stage sequence length.\\
         \hline
         Skimming Recurrent Transformers \cite{dai-etal-2019-transformer} & $\mathcal{O}(n^2)$ Transformer Cost \break\break Context Fragmentation \break\break Global Context \break\break Forward context & Forward and global context of greater quality can be obtained by passing through the input text twice, once by peeking small text segments (skimming phase), and then by performing a full pass (retrospective phase), fully utilizing the recurrence mechanism.\\
         \hline
    \end{tabular}
    \caption{Issues and solutions for long document classification.}
    \label{tab::text_class_table}
\end{table}

\begin{table}
    \scriptsize
    \begin{tabular}
         { |p{3cm}|p{3cm}|p{9cm}|  }
         \hline
         \cellcolor{blue!25}\textbf{Method} & \cellcolor{blue!25}\textbf{Addressed Issues} & \cellcolor{blue!25}\textbf{Details}\\
        \hline
       \rowcolor{lightgray} 
         \multicolumn{3}{|c|}{Extractive Document Summarization} \\
         \hline
         Hierarchical Summarization \cite{lapata, nallapati, xiao} & DNN Use for Summarization  \break\break Document Structure Under-utilization & Traditional Encoder-Decoder DNN architectures attempt to explicitly incorporate context and hierarchical elements to the features used by CNN and RNN/LSTM extractive models.\\
         \hline
         BERTSUM-Ext \cite{nallapati2} & Poor Traditional DNN Performance\break\break Weaknesses in Sentence Embeddings & By using BERT as Encoder and attaching stacked Transformer layers, BERT's expressive power leads to accurate extractive summaries.\\  
         \hline
         HiStruct+ \cite{histruct} & Segment/Document Context Loss\break\break Under-utilization of Document Structure & By encoding the hierarchical and topical structure of a document, the Transformer can leverage more information about rigidly structured documents (e.g., research articles).\\
         \hline
         HEGEL \cite{hegel} & Under-utilization of Document Structure\break\break $\mathcal{O}(n^2)$ Transformer Cost\break\break Long-span Dependencies & Encoding each document as a hypergraph under multiple different views leads to the model being exceptionally adept at summarizing very long and rigidly hierarchical documents.\\
         \hline
\rowcolor{lightgray}
         \multicolumn{3}{|c|}{Abstractive Document Summarization} \\
         \hline
         Pointer-Generator Networks \cite{abigail} & Word Repetition\break\break Vocabulary Limitations\break\break  Long Document Computational Cost & By allowing the network to choose whether to copy the next summary word from the source text or predict its own, the model can benefit from a significantly expanded vocabulary, reduce computation costs by keeping a smaller vocabulary and bypass the word repetition problem found in many extractive summarization models. \\
         \hline
         Sparse Attention Transformers \cite{longformer, big_bird, Guo22LongT5} & $\mathcal{O}(n^2)$ Transformer Cost \break\break  Vocabulary Limitations & By using a modified Sparse Transformer as the Encoder, the model can leverage the expressive power of Transformers on long documents. \\
         \hline
         Local Attention and Content Selection \cite{gales} & $\mathcal{O}(n^2)$ Transformer Cost  \break\break Long Document Computational Cost & Pretrained models can be used for content selection in order for a Sparse Attention model to be fed only useful data, both in training and inference.\\ 
         \hline
    \end{tabular}
    \caption{Issues and solutions for long document summarization.}
    \label{tab::text_sum_table}
\end{table}

% \begin{table}
%     \begin{tabular}
%          { |p{5cm}|p{3cm}|p{3cm}|p{3cm}|  }
%          \hline
%          \cellcolor{blue!25}\textbf{Model} & \cellcolor{blue!25}\textbf{CNN / Daily Mail} & \cellcolor{blue!25}\textbf{ArXiv} &
%          \cellcolor{blue!25}\textbf{PubMed}\\
%         \hline
%        \rowcolor{lightgray} 
%          \multicolumn{4}{|c|}{Extractive Document Summarization} \\
%          \hline
%          SummaRuNNer \cite{nallapati} & 42.0 & 42.81 & 43.89\\
%          \hline
%          BERTSUM-Ext \cite{nallapati2} & 43.85 & - & - \\  
%          \hline
%          HiStruct+ \cite{histruct} & 43.65 & 45.22 & - \\
%          \hline
%          HEGEL \cite{hegel} & - & 46.61 & 43.89\\
%          \hline
% \rowcolor{lightgray}
%          \multicolumn{4}{|c|}{Abstractive Document Summarization} \\
%          \hline
%          Pointer-Generator Networks \cite{abigail} & 39.53 & 32.06 & 35.86 \\
%          \hline
%          BigBird \cite{big_bird} & 43.84 & 46.63 & 46.32 \\
%          \hline
%          Local Attention and Content Selection (Abstractive) \cite{gales} & - & 46.59 & 47.47\\ 
%          \hline
%     \end{tabular}
%     \caption{The ROUGE-1 scores (higher is better) for SoA extractive and abstractive summarizers on the three most popular summarization datasets (where available). These results are only indicative, since the ROUGE score's correlation with summarization quality is under debate (see Section \ref{sss::SummarizationEvaluation}).}
%     \label{tab::ResultsSummarization}
% \end{table}

Analyzing long documents currently presents a daunting challenge, even with the recent NLP breakthroughs. While traditional machine learning and statistical methods provide limited functionality on these tasks, they ultimately prove less adaptable and less accurate than modern DNNs. However, the latter ones need to face long-standing issues, such as the curse of dimensionality, limited datasets, prohibitively large inputs and difficult writing formats (e.g., in literary fiction).

Most major issues are common to all long document analysis tasks and currently tend to require similar solutions, such as limiting the model's vocabulary, modifying existing Transformer architectures to handle longer input sequences, efficiently exploiting the hierarchical structure of documents and artificially augmenting the training datasets. Task-specific solutions have also been proposed, such as using content preselection and specialized GNNs to capture long-spanning context in summarization, or passing over the input twice and using document-level representations to enrich sentence-level representations in classification. The identified issues and existing solutions are summarized in Tables \ref{tab::text_class_table} and \ref{tab::text_sum_table}, concerning document classification and summarization, respectively. Overall, further investigation of hybrid and recurrent Transformer architectures, as well as of the top-down Transformer, seem to be promising directions for near-future improvements. Incorporation of such advances into generic LLMs is also an avenue worth exploring, along with the development of novel, improved evaluation metrics for automatic text summarization.

While this survey focuses only on document-level analysis for long texts, it becomes apparent how attempts at solving specific tasks can lead to solutions or insights with impact to all NLP fields. However, it is important to note that existing SoA solutions rely on useful trade-offs and, therefore, cannot be used universally. Finally, there is significant research potential for research into literary sentiment analysis, in collaboration with the digital/computational Humanities community. The necessary theoretical background already exists, while the mentioned breakthroughs in long document NLP have not been yet applied.

Automated analysis of long documents is essentially a field still in its infancy, with breakthroughs leading to large jumps in computational efficiency and accuracy still frequent. Since 2015 neural network models have evolved from being constrained to (occasionally artificially) small-sized documents, to being able to effectively parse large articles and books. It is not unreasonable to expect that continuing research will help bypass most of the current limitations within the next few years.

\section*{Acknowledgement}
The research leading to these results has received funding from the European Union’s Internal Security Fund under grant agreement No 101103298 (KLEPTOTRACE). This publication reflects only the authors’ views. The European Commission is not responsible for any use that may be made of the information it contains.

\bibliographystyle{elsarticle-num-names} 
\bibliography{refs}

\begin{thebibliography}{212}
\expandafter\ifx\csname natexlab\endcsname\relax\def\natexlab#1{#1}\fi
\providecommand{\url}[1]{\texttt{#1}}
\providecommand{\href}[2]{#2}
\providecommand{\path}[1]{#1}
\providecommand{\DOIprefix}{doi:}
\providecommand{\ArXivprefix}{arXiv:}
\providecommand{\URLprefix}{URL: }
\providecommand{\Pubmedprefix}{pmid:}
\providecommand{\doi}[1]{\href{http://dx.doi.org/#1}{\path{#1}}}
\providecommand{\Pubmed}[1]{\href{pmid:#1}{\path{#1}}}
\providecommand{\bibinfo}[2]{#2}
\ifx\xfnm\relax \def\xfnm[#1]{\unskip,\space#1}\fi
%Type = Inproceedings
\bibitem[{Worsham and Kalita(2018)}]{worsham}
\bibinfo{author}{J.~Worsham}, \bibinfo{author}{J.~Kalita},
\newblock \bibinfo{title}{Genre identification and the compositional effect of
  genre in literature},
\newblock in: \bibinfo{booktitle}{Proceedings of the International Conference
  on Computational Linguistics}, \bibinfo{year}{2018}.
%Type = Book
\bibitem[{Worsham(2014)}]{worsham_book}
\bibinfo{author}{J.~M. Worsham}, \bibinfo{title}{Towards Literary Genre
  Identification: Applied Neural Networks for Large Text Classification},
  \bibinfo{publisher}{University of Colorado}, \bibinfo{year}{2014}.
%Type = Inproceedings
\bibitem[{Monte-Serrat et~al.(2021)Monte-Serrat, Machado, and Ruiz}]{brazil}
\bibinfo{author}{D.~M. Monte-Serrat}, \bibinfo{author}{M.~T. Machado},
  \bibinfo{author}{E.~E.~S. Ruiz},
\newblock \bibinfo{title}{A machine learning approach to literary genre
  classification on portuguese texts: circumventing {NLP}’s standard
  varieties},
\newblock in: \bibinfo{booktitle}{Anais do XIII Simp{\'o}sio Brasileiro de
  Tecnologia da Informa{\c{c}}{\~a}o e da Linguagem Humana},
  \bibinfo{organization}{SBC}, \bibinfo{year}{2021}.
%Type = Inproceedings
\bibitem[{L{\"u}schow and Tello(2021)}]{jose}
\bibinfo{author}{A.~L{\"u}schow}, \bibinfo{author}{J.~C. Tello},
\newblock \bibinfo{title}{Towards genre classification in the library catalog},
\newblock in: \bibinfo{booktitle}{Proceedings of the Conference on Digital
  Curation Technologies (Qurator)}, \bibinfo{year}{2021}.
%Type = Article
\bibitem[{Manakul and Gales(2021)}]{gales}
\bibinfo{author}{P.~Manakul}, \bibinfo{author}{M.~J.~F. Gales},
\newblock \bibinfo{title}{Long-span dependencies in {Transformer}-based
  summarization systems},
\newblock \bibinfo{journal}{arXiv} \bibinfo{volume}{arXiv:2105.03801}
  (\bibinfo{year}{2021}).
%Type = Inproceedings
\bibitem[{Nabizadeh et~al.(2020)Nabizadeh, Kolossa, and Heckmann}]{nabizadeh}
\bibinfo{author}{N.~Nabizadeh}, \bibinfo{author}{D.~Kolossa},
  \bibinfo{author}{M.~Heckmann},
\newblock \bibinfo{title}{{M}y{F}ixit: An annotated dataset, annotation tool,
  and baseline methods for information extraction from repair manuals},
\newblock in: \bibinfo{booktitle}{Proceedings of the Language Resources and
  Evaluation Conference}, \bibinfo{publisher}{European Language Resources
  Association}, \bibinfo{year}{2020}.
%Type = Article
\bibitem[{Chen et~al.(2022)Chen, Cong, and Lv}]{shuo}
\bibinfo{author}{X.~Chen}, \bibinfo{author}{P.~Cong}, \bibinfo{author}{S.~Lv},
\newblock \bibinfo{title}{A long-text classification method of chinese news
  based on {BERT and CNN}},
\newblock \bibinfo{journal}{IEEE Access} \bibinfo{volume}{10}
  (\bibinfo{year}{2022}) \bibinfo{pages}{34046--34057}.
%Type = Article
\bibitem[{Wan et~al.(2019)Wan, Papageorgiou, Seddon, and Bernardoni}]{lulu}
\bibinfo{author}{L.~Wan}, \bibinfo{author}{G.~Papageorgiou},
  \bibinfo{author}{M.~Seddon}, \bibinfo{author}{M.~Bernardoni},
\newblock \bibinfo{title}{Long-length legal document classification},
\newblock \bibinfo{journal}{arXiv} \bibinfo{volume}{arXiv:1912.06905}
  (\bibinfo{year}{2019}).
%Type = Article
\bibitem[{Dale(2019)}]{dale}
\bibinfo{author}{R.~Dale},
\newblock \bibinfo{title}{Law and word order: {NLP} in legal tech},
\newblock \bibinfo{journal}{Natural Language Engineering} \bibinfo{volume}{25}
  (\bibinfo{year}{2019}) \bibinfo{pages}{211--217}.
%Type = Inproceedings
\bibitem[{Merchant and Pande(2018)}]{merchant}
\bibinfo{author}{K.~Merchant}, \bibinfo{author}{Y.~Pande},
\newblock \bibinfo{title}{{NLP-}based latent semantic analysis for legal text
  summarization},
\newblock in: \bibinfo{booktitle}{Proceedings of the IEEE International
  Conference on Advances in Computing, Communications and Informatics
  (ICACCI)}, \bibinfo{year}{2018}.
%Type = Article
\bibitem[{Mademlis et~al.(2023)Mademlis, Mancuso, Paternoster, Evangelatos,
  Finlay, Hughes, Radoglou-Grammatikis, Sarigiannidis, Stavropoulos, Votis, and
  Papadopoulos}]{mademlis2023invisible}
\bibinfo{author}{I.~Mademlis}, \bibinfo{author}{M.~Mancuso},
  \bibinfo{author}{C.~Paternoster}, \bibinfo{author}{S.~Evangelatos},
  \bibinfo{author}{E.~Finlay}, \bibinfo{author}{J.~Hughes},
  \bibinfo{author}{P.~Radoglou-Grammatikis},
  \bibinfo{author}{P.~Sarigiannidis}, \bibinfo{author}{G.~Stavropoulos},
  \bibinfo{author}{K.~Votis}, \bibinfo{author}{G.~T. Papadopoulos},
\newblock \bibinfo{title}{The invisible arms race: digital trends in illicit
  goods trafficking and {AI}-enabled responses},
\newblock \bibinfo{journal}{Authorea Preprints}  (\bibinfo{year}{2023}).
%Type = Article
\bibitem[{Bayer et~al.(2023)Bayer, Kaufhold, Buchhold, Keller, Dallmeyer, and
  Reuter}]{hold}
\bibinfo{author}{M.~Bayer}, \bibinfo{author}{M.-A. Kaufhold},
  \bibinfo{author}{B.~Buchhold}, \bibinfo{author}{M.~Keller},
  \bibinfo{author}{J.~Dallmeyer}, \bibinfo{author}{C.~Reuter},
\newblock \bibinfo{title}{Data augmentation in {Natural Language Processing}: a
  novel text generation approach for long and short text classifiers},
\newblock \bibinfo{journal}{International Journal of Machine Learning and
  Cybernetics} \bibinfo{volume}{14} (\bibinfo{year}{2023})
  \bibinfo{pages}{135--150}.
%Type = Inproceedings
\bibitem[{Yogarajan et~al.(2021)Yogarajan, Montiel, Smith, and
  Pfahringer}]{vithya}
\bibinfo{author}{V.~Yogarajan}, \bibinfo{author}{J.~Montiel},
  \bibinfo{author}{T.~Smith}, \bibinfo{author}{B.~Pfahringer},
\newblock \bibinfo{title}{Transformers for multi-label classification of
  medical text: An empirical comparison},
\newblock in: \bibinfo{booktitle}{Proceedings of the International Conference
  on Artificial Intelligence in Medicine}, \bibinfo{year}{2021}.
%Type = Inproceedings
\bibitem[{Bambroo and Awasthi(2021)}]{bambroo}
\bibinfo{author}{P.~Bambroo}, \bibinfo{author}{A.~Awasthi},
\newblock \bibinfo{title}{{LegalDB: Long DistilBERT} for legal document
  classification},
\newblock in: \bibinfo{booktitle}{Proceedings of the International Conference
  on Advances in Electrical, Computing, Communication and Sustainable
  Technologies (ICAECT)}, \bibinfo{year}{2021}.
%Type = Article
\bibitem[{Kowsari et~al.(2019)Kowsari, Jafari~Meimandi, Heidarysafa, Mendu,
  Barnes, and Brown}]{kowasari}
\bibinfo{author}{K.~Kowsari}, \bibinfo{author}{K.~Jafari~Meimandi},
  \bibinfo{author}{M.~Heidarysafa}, \bibinfo{author}{S.~Mendu},
  \bibinfo{author}{L.~Barnes}, \bibinfo{author}{D.~Brown},
\newblock \bibinfo{title}{Text classification algorithms: A survey},
\newblock \bibinfo{journal}{Information} \bibinfo{volume}{10}
  (\bibinfo{year}{2019}).
%Type = Article
\bibitem[{Torfi et~al.(2020)Torfi, Shirvani, Keneshloo, Tavaf, and Fox}]{torfi}
\bibinfo{author}{A.~Torfi}, \bibinfo{author}{R.~A. Shirvani},
  \bibinfo{author}{Y.~Keneshloo}, \bibinfo{author}{N.~Tavaf},
  \bibinfo{author}{E.~A. Fox},
\newblock \bibinfo{title}{{Natural Language Processing} advancements by deep
  learning: A survey},
\newblock \bibinfo{journal}{arXiv} \bibinfo{volume}{arXiv:2003.01200}
  (\bibinfo{year}{2020}).
%Type = Inproceedings
\bibitem[{Chai and Li(2019)}]{chai}
\bibinfo{author}{J.~Chai}, \bibinfo{author}{A.~Li},
\newblock \bibinfo{title}{Deep learning in {Natural Language Processing}: A
  state-of-the-art survey},
\newblock in: \bibinfo{booktitle}{Proceedings of the International Conference
  on Machine Learning and Cybernetics (ICMLC)}, \bibinfo{year}{2019}.
%Type = Article
\bibitem[{Schr{\"{o}}der and Niekler(2020)}]{nielker}
\bibinfo{author}{C.~Schr{\"{o}}der}, \bibinfo{author}{A.~Niekler},
\newblock \bibinfo{title}{A survey of active learning for text classification
  using {Deep Neural Networks}},
\newblock \bibinfo{journal}{arXiv} \bibinfo{volume}{arXiv:2008.07267}
  (\bibinfo{year}{2020}).
%Type = Article
\bibitem[{Li et~al.(2022)Li, Peng, Li, Xia, Yang, Sun, Yu, and He}]{qian}
\bibinfo{author}{Q.~Li}, \bibinfo{author}{H.~Peng}, \bibinfo{author}{J.~Li},
  \bibinfo{author}{C.~Xia}, \bibinfo{author}{R.~Yang},
  \bibinfo{author}{L.~Sun}, \bibinfo{author}{P.~S. Yu},
  \bibinfo{author}{L.~He},
\newblock \bibinfo{title}{A survey on text classification: From traditional to
  deep learning},
\newblock \bibinfo{journal}{ACM Transactions on Intelligent Systems and
  Technology} \bibinfo{volume}{13} (\bibinfo{year}{2022}).
%Type = Article
\bibitem[{Gupta and Gupta(2019)}]{gupta}
\bibinfo{author}{S.~Gupta}, \bibinfo{author}{S.~K. Gupta},
\newblock \bibinfo{title}{Abstractive summarization: An overview of the state
  of the art},
\newblock \bibinfo{journal}{Expert Systems with Applications}
  \bibinfo{volume}{121} (\bibinfo{year}{2019}) \bibinfo{pages}{49--65}.
%Type = Article
\bibitem[{Hussein(2018)}]{hussein}
\bibinfo{author}{D.~M. E.-D.~M. Hussein},
\newblock \bibinfo{title}{A survey on sentiment analysis challenges},
\newblock \bibinfo{journal}{Journal of King Saud University-Engineering
  Sciences} \bibinfo{volume}{30} (\bibinfo{year}{2018})
  \bibinfo{pages}{330--338}.
%Type = Incollection
\bibitem[{Rumelhart(1986)}]{rumelhart}
\bibinfo{author}{D.~E. Rumelhart},
\newblock \bibinfo{title}{Learning internal representations by error
  propagation, in parallel distributed processing},
\newblock \bibinfo{publisher}{MIT press}, \bibinfo{year}{1986}, pp.
  \bibinfo{pages}{318--362}.
%Type = Article
\bibitem[{Jordan(1997)}]{jordan}
\bibinfo{author}{M.~I. Jordan},
\newblock \bibinfo{title}{Serial order: A parallel distributed processing
  approach} \bibinfo{volume}{121} (\bibinfo{year}{1997})
  \bibinfo{pages}{471--495}.
%Type = Article
\bibitem[{Lecun et~al.(1998)Lecun, Bottou, Bengio, and Haffner}]{lenet}
\bibinfo{author}{Y.~Lecun}, \bibinfo{author}{L.~Bottou},
  \bibinfo{author}{Y.~Bengio}, \bibinfo{author}{P.~Haffner},
\newblock \bibinfo{title}{Gradient-based learning applied to document
  recognition},
\newblock \bibinfo{journal}{Proceedings of the IEEE} \bibinfo{volume}{86}
  (\bibinfo{year}{1998}) \bibinfo{pages}{2278--2324}.
%Type = Article
\bibitem[{Krizhevsky et~al.(2017)Krizhevsky, Sutskever, and Hinton}]{alexnet}
\bibinfo{author}{A.~Krizhevsky}, \bibinfo{author}{I.~Sutskever},
  \bibinfo{author}{G.~E. Hinton},
\newblock \bibinfo{title}{{ImageNet} classification with deep {Convolutional
  Neural Networks}},
\newblock \bibinfo{journal}{Communications of the ACM} \bibinfo{volume}{60}
  (\bibinfo{year}{2017}) \bibinfo{pages}{84–--90}.
%Type = Article
\bibitem[{Siegelmann and Sontag(1991)}]{siegelman}
\bibinfo{author}{H.~T. Siegelmann}, \bibinfo{author}{E.~D. Sontag},
\newblock \bibinfo{title}{Turing computability with neural nets},
\newblock \bibinfo{journal}{Applied Mathematics Letters} \bibinfo{volume}{4}
  (\bibinfo{year}{1991}) \bibinfo{pages}{77--80}.
%Type = Article
\bibitem[{Hochreiter and Schmidhuber(1997)}]{sepp}
\bibinfo{author}{S.~Hochreiter}, \bibinfo{author}{J.~Schmidhuber},
\newblock \bibinfo{title}{{Long Short-Term Memory}},
\newblock \bibinfo{journal}{Neural computation} \bibinfo{volume}{9}
  (\bibinfo{year}{1997}) \bibinfo{pages}{1735--80}.
%Type = Article
\bibitem[{Cho et~al.(2014)Cho, Van~Merri{\"e}nboer, Bahdanau, and
  Bengio}]{Cho2014}
\bibinfo{author}{K.~Cho}, \bibinfo{author}{B.~Van~Merri{\"e}nboer},
  \bibinfo{author}{D.~Bahdanau}, \bibinfo{author}{Y.~Bengio},
\newblock \bibinfo{title}{On the properties of neural machine translation:
  {Encoder-Decoder} approaches},
\newblock \bibinfo{journal}{arXiv} \bibinfo{volume}{arXiv:1409.1259}
  (\bibinfo{year}{2014}).
%Type = Inproceedings
\bibitem[{Bahdanau et~al.(2015)Bahdanau, Cho, and Bengio}]{bahdanau}
\bibinfo{author}{D.~Bahdanau}, \bibinfo{author}{K.~Cho},
  \bibinfo{author}{Y.~Bengio},
\newblock \bibinfo{title}{{Neural Machine Translation} by jointly learning to
  align and translate},
\newblock in: \bibinfo{booktitle}{Proceedings of the International Conference
  on Learning Representations (ICLR)}, \bibinfo{year}{2015}.
%Type = Inproceedings
\bibitem[{Vaswani et~al.(2017)Vaswani, Shazeer, Parmar, Uszkoreit, Jones,
  Gomez, Kaiser, and Polosukhin}]{ilia}
\bibinfo{author}{A.~Vaswani}, \bibinfo{author}{N.~Shazeer},
  \bibinfo{author}{N.~Parmar}, \bibinfo{author}{J.~Uszkoreit},
  \bibinfo{author}{L.~Jones}, \bibinfo{author}{A.~N. Gomez},
  \bibinfo{author}{L.~Kaiser}, \bibinfo{author}{I.~Polosukhin},
\newblock \bibinfo{title}{Attention is all you need},
\newblock in: \bibinfo{booktitle}{Proceedings of the Advances in Neural
  Information Processing Systems (NIPS)}, \bibinfo{year}{2017}.
%Type = Article
\bibitem[{Mikolov et~al.(2013)Mikolov, Chen, Corrado, and Dean}]{mikolov}
\bibinfo{author}{T.~Mikolov}, \bibinfo{author}{K.~Chen},
  \bibinfo{author}{G.~Corrado}, \bibinfo{author}{J.~Dean},
\newblock \bibinfo{title}{Efficient estimation of word representations in
  vector space},
\newblock \bibinfo{journal}{arXiv preprint arXiv:1301.3781}
  (\bibinfo{year}{2013}).
%Type = Book
\bibitem[{Jurafsky and Martin(2000)}]{jurafsky}
\bibinfo{author}{D.~Jurafsky}, \bibinfo{author}{J.~H. Martin},
  \bibinfo{title}{Speech and Language Processing},
  \bibinfo{publisher}{Pearson}, \bibinfo{year}{2000}.
%Type = Inproceedings
\bibitem[{Pennington et~al.(2014)Pennington, Socher, and Manning}]{glove}
\bibinfo{author}{J.~Pennington}, \bibinfo{author}{R.~Socher},
  \bibinfo{author}{C.~Manning},
\newblock \bibinfo{title}{{G}lo{V}e: Global vectors for word representation},
\newblock in: \bibinfo{booktitle}{Proceedings of the Conference on Empirical
  Methods in Natural Language Processing ({EMNLP})},
  \bibinfo{publisher}{Association for Computational Linguistics},
  \bibinfo{year}{2014}.
%Type = Inproceedings
\bibitem[{Peters et~al.(2018)Peters, Neumann, Iyyer, Gardner, Clark, Lee, and
  Zettlemoyer}]{elmo}
\bibinfo{author}{M.~E. Peters}, \bibinfo{author}{M.~Neumann},
  \bibinfo{author}{M.~Iyyer}, \bibinfo{author}{M.~Gardner},
  \bibinfo{author}{C.~Clark}, \bibinfo{author}{K.~Lee},
  \bibinfo{author}{L.~Zettlemoyer},
\newblock \bibinfo{title}{Deep contextualized word representations},
\newblock in: \bibinfo{booktitle}{Proceedings of the Conference of the North
  {A}merican Chapter of the Association for Computational Linguistics},
  \bibinfo{year}{2018}.
%Type = Inproceedings
\bibitem[{Devlin et~al.(2019)Devlin, Chang, Lee, and Toutanova}]{toutanova}
\bibinfo{author}{J.~Devlin}, \bibinfo{author}{M.-W. Chang},
  \bibinfo{author}{K.~Lee}, \bibinfo{author}{K.~Toutanova},
\newblock \bibinfo{title}{{BERT}: Pre-training of deep bidirectional
  {Transformers} for language understanding},
\newblock in: \bibinfo{booktitle}{Proceedings of the Conference of the North
  {A}merican Chapter of the Association for Computational Linguistics: Human
  Language Technologies, Volume 1 (Long and Short Papers)},
  \bibinfo{publisher}{Association for Computational Linguistics},
  \bibinfo{year}{2019}.
%Type = Article
\bibitem[{Taylor(1953)}]{taylor}
\bibinfo{author}{W.~L. Taylor},
\newblock \bibinfo{title}{“cloze procedure”: A new tool for measuring
  readability},
\newblock \bibinfo{journal}{Journalism Quarterly} \bibinfo{volume}{30}
  (\bibinfo{year}{1953}) \bibinfo{pages}{415--433}.
%Type = Article
\bibitem[{Liu et~al.(2019)Liu, Ott, Goyal, Du, Joshi, Chen, Levy, Lewis,
  Zettlemoyer, and Stoyanov}]{roberta}
\bibinfo{author}{Y.~Liu}, \bibinfo{author}{M.~Ott}, \bibinfo{author}{N.~Goyal},
  \bibinfo{author}{J.~Du}, \bibinfo{author}{M.~Joshi},
  \bibinfo{author}{D.~Chen}, \bibinfo{author}{O.~Levy},
  \bibinfo{author}{M.~Lewis}, \bibinfo{author}{L.~Zettlemoyer},
  \bibinfo{author}{V.~Stoyanov},
\newblock \bibinfo{title}{{RoBERTa}: A robustly optimized {BERT} pretraining
  approach},
\newblock \bibinfo{journal}{arXiv} \bibinfo{volume}{arXiv:1907.11692}
  (\bibinfo{year}{2019}).
%Type = Inproceedings
\bibitem[{Clark et~al.(2020)Clark, Luong, Le, and Manning}]{karta}
\bibinfo{author}{K.~Clark}, \bibinfo{author}{M.~Luong}, \bibinfo{author}{Q.~V.
  Le}, \bibinfo{author}{C.~D. Manning},
\newblock \bibinfo{title}{{ELECTRA:} pre-training text encoders as
  discriminators rather than generators},
\newblock in: \bibinfo{booktitle}{Proceedings of the International Conference
  on Learning Representations (ICLR)}, \bibinfo{year}{2020}.
%Type = Article
\bibitem[{Raffel et~al.(2020)Raffel, Shazeer, Roberts, Lee, Narang, Matena,
  Zhou, Li, and Liu}]{t5}
\bibinfo{author}{C.~Raffel}, \bibinfo{author}{N.~Shazeer},
  \bibinfo{author}{A.~Roberts}, \bibinfo{author}{K.~Lee},
  \bibinfo{author}{S.~Narang}, \bibinfo{author}{M.~Matena},
  \bibinfo{author}{Y.~Zhou}, \bibinfo{author}{W.~Li}, \bibinfo{author}{P.~J.
  Liu},
\newblock \bibinfo{title}{Exploring the limits of transfer learning with a
  unified text-to-text transformer},
\newblock \bibinfo{journal}{Journal of Machine Learning Research}
  \bibinfo{volume}{21} (\bibinfo{year}{2020}) \bibinfo{pages}{5485--5551}.
%Type = Article
\bibitem[{Lewis et~al.(2019)Lewis, Liu, Goyal, Ghazvininejad, Mohamed, Levy,
  Stoyanov, and Zettlemoyer}]{BART}
\bibinfo{author}{M.~Lewis}, \bibinfo{author}{Y.~Liu},
  \bibinfo{author}{N.~Goyal}, \bibinfo{author}{M.~Ghazvininejad},
  \bibinfo{author}{A.~Mohamed}, \bibinfo{author}{O.~Levy},
  \bibinfo{author}{V.~Stoyanov}, \bibinfo{author}{L.~Zettlemoyer},
\newblock \bibinfo{title}{{BART:} denoising sequence-to-sequence pre-training
  for natural language generation, translation and comprehension},
\newblock \bibinfo{journal}{arXiv} \bibinfo{volume}{arXiv:1910.13461}
  (\bibinfo{year}{2019}).
%Type = Inproceedings
\bibitem[{Tay et~al.(2022)Tay, Dehghani, Tran, Garcia, Wei, Wang, Chung, Bahri,
  Schuster, and Zheng}]{tay2022ul2}
\bibinfo{author}{Y.~Tay}, \bibinfo{author}{M.~Dehghani}, \bibinfo{author}{V.~Q.
  Tran}, \bibinfo{author}{X.~Garcia}, \bibinfo{author}{J.~Wei},
  \bibinfo{author}{X.~Wang}, \bibinfo{author}{H.~W. Chung},
  \bibinfo{author}{D.~Bahri}, \bibinfo{author}{T.~Schuster},
  \bibinfo{author}{S.~e.~a. Zheng},
\newblock \bibinfo{title}{{UL2}: Unifying language learning paradigms},
\newblock in: \bibinfo{booktitle}{Proceedings of the International Conference
  on Learning Representations (ICLR)}, \bibinfo{year}{2022}.
%Type = Misc
\bibitem[{Radford et~al.(2018{\natexlab{a}})Radford, Narasimhan, Salimans, and
  Sutskever}]{GPT}
\bibinfo{author}{A.~Radford}, \bibinfo{author}{K.~Narasimhan},
  \bibinfo{author}{T.~Salimans}, \bibinfo{author}{I.~Sutskever},
  \bibinfo{title}{Improving language understanding by generative pre-training},
  \bibinfo{year}{2018}{\natexlab{a}}.
%Type = Misc
\bibitem[{Radford et~al.(2018{\natexlab{b}})Radford, Wu, Child, Luan, Amodei,
  and Sutskever}]{GPT2}
\bibinfo{author}{A.~Radford}, \bibinfo{author}{J.~Wu},
  \bibinfo{author}{R.~Child}, \bibinfo{author}{D.~Luan},
  \bibinfo{author}{D.~Amodei}, \bibinfo{author}{I.~Sutskever},
  \bibinfo{title}{Language models are unsupervised multitask learners},
  \bibinfo{year}{2018}{\natexlab{b}}.
%Type = Inproceedings
\bibitem[{Tan et~al.(2021)Tan, Yang, Al-Shedivat, Xing, and Hu}]{tanyangxing}
\bibinfo{author}{B.~Tan}, \bibinfo{author}{Z.~Yang},
  \bibinfo{author}{M.~Al-Shedivat}, \bibinfo{author}{E.~P. Xing},
  \bibinfo{author}{Z.~Hu},
\newblock \bibinfo{title}{Progressive generation of long text with pretrained
  language models},
\newblock in: \bibinfo{booktitle}{Proceedings of the Conference of the North
  American Chapter of the Association for Computational Linguistics: Human
  Language Technologies}, \bibinfo{year}{2021}.
%Type = Article
\bibitem[{Chowdhery et~al.(2022)Chowdhery, Narang, Devlin, Bosma, Mishra,
  Roberts, Barham, Chung, Sutton, and Gehrmann}]{chowdhery2022palm}
\bibinfo{author}{A.~Chowdhery}, \bibinfo{author}{S.~Narang},
  \bibinfo{author}{J.~Devlin}, \bibinfo{author}{M.~Bosma},
  \bibinfo{author}{G.~Mishra}, \bibinfo{author}{A.~Roberts},
  \bibinfo{author}{P.~Barham}, \bibinfo{author}{H.~W. Chung},
  \bibinfo{author}{C.~Sutton}, \bibinfo{author}{S.~e.~a. Gehrmann},
\newblock \bibinfo{title}{{PaLM}: Scaling language modeling with pathways},
\newblock \bibinfo{journal}{arXiv preprint arXiv:2204.02311}
  (\bibinfo{year}{2022}).
%Type = Article
\bibitem[{Team(2023)}]{gemini}
\bibinfo{author}{G.~Team},
\newblock \bibinfo{title}{Gemini: A family of highly capable multimodal
  models},
\newblock \bibinfo{journal}{arXiv preprint 2312.11805}  (\bibinfo{year}{2023}).
%Type = Article
\bibitem[{Scao et~al.(2022)Scao, Fan, Akiki, Pavlick, Ili{\'c}, Hesslow,
  Castagn{\'e}, Luccioni, Yvon, and Gall{\'e}}]{scao2022bloom}
\bibinfo{author}{T.~L. Scao}, \bibinfo{author}{A.~Fan},
  \bibinfo{author}{C.~Akiki}, \bibinfo{author}{E.~Pavlick},
  \bibinfo{author}{S.~Ili{\'c}}, \bibinfo{author}{D.~Hesslow},
  \bibinfo{author}{R.~Castagn{\'e}}, \bibinfo{author}{A.~S. Luccioni},
  \bibinfo{author}{F.~Yvon}, \bibinfo{author}{M.~e.~a. Gall{\'e}},
\newblock \bibinfo{title}{{BLOOM: A 176B}-parameter open-access multilingual
  language model},
\newblock \bibinfo{journal}{arXiv preprint arXiv:2211.05100}
  (\bibinfo{year}{2022}).
%Type = Article
\bibitem[{Zhang et~al.(2022)Zhang, Roller, Goyal, Artetxe, Chen, Chen, Dewan,
  Diab, Li, and Lin}]{zhang2022opt}
\bibinfo{author}{S.~Zhang}, \bibinfo{author}{S.~Roller},
  \bibinfo{author}{N.~Goyal}, \bibinfo{author}{M.~Artetxe},
  \bibinfo{author}{M.~Chen}, \bibinfo{author}{S.~Chen},
  \bibinfo{author}{C.~Dewan}, \bibinfo{author}{M.~Diab},
  \bibinfo{author}{X.~Li}, \bibinfo{author}{X.~V. e.~a. Lin},
\newblock \bibinfo{title}{{OPT}: Open pre-trained transformer language models},
\newblock \bibinfo{journal}{arXiv preprint arXiv:2205.01068}
  (\bibinfo{year}{2022}).
%Type = Misc
\bibitem[{Hoffmann et~al.(2022)Hoffmann, Borgeaud, Mensch, Buchatskaya, Cai,
  Rutherford, de~Las~Casas, Hendricks, Welbl, Clark, Hennigan, Noland,
  Millican, van~den Driessche, Damoc, Guy, Osindero, Simonyan, Elsen, Rae,
  Vinyals, and Sifre}]{chinchilla}
\bibinfo{author}{J.~Hoffmann}, \bibinfo{author}{S.~Borgeaud},
  \bibinfo{author}{A.~Mensch}, \bibinfo{author}{E.~Buchatskaya},
  \bibinfo{author}{T.~Cai}, \bibinfo{author}{E.~Rutherford},
  \bibinfo{author}{D.~de~Las~Casas}, \bibinfo{author}{L.~A. Hendricks},
  \bibinfo{author}{J.~Welbl}, \bibinfo{author}{A.~Clark},
  \bibinfo{author}{T.~Hennigan}, \bibinfo{author}{E.~Noland},
  \bibinfo{author}{K.~Millican}, \bibinfo{author}{G.~van~den Driessche},
  \bibinfo{author}{B.~Damoc}, \bibinfo{author}{A.~Guy},
  \bibinfo{author}{S.~Osindero}, \bibinfo{author}{K.~Simonyan},
  \bibinfo{author}{E.~Elsen}, \bibinfo{author}{J.~W. Rae},
  \bibinfo{author}{O.~Vinyals}, \bibinfo{author}{L.~Sifre},
  \bibinfo{title}{Training compute-optimal large language models},
  \bibinfo{year}{2022}. \href{http://arxiv.org/abs/2203.15556}{{\tt
  arXiv:2203.15556}}.
%Type = Article
\bibitem[{Touvron et~al.(2023)Touvron, Lavril, Izacard, Martinet, Lachaux,
  Lacroix, Rozi{\`e}re, Goyal, Hambro, and Azhar}]{Touvron2023llama}
\bibinfo{author}{H.~Touvron}, \bibinfo{author}{T.~Lavril},
  \bibinfo{author}{G.~Izacard}, \bibinfo{author}{X.~Martinet},
  \bibinfo{author}{M.-A. Lachaux}, \bibinfo{author}{T.~Lacroix},
  \bibinfo{author}{B.~Rozi{\`e}re}, \bibinfo{author}{N.~Goyal},
  \bibinfo{author}{E.~Hambro}, \bibinfo{author}{F.~e.~a. Azhar},
\newblock \bibinfo{title}{{LLaMA}: Open and efficient foundation language
  models},
\newblock \bibinfo{journal}{arXiv preprint arXiv:2302.13971}
  (\bibinfo{year}{2023}).
%Type = Misc
\bibitem[{Devlin et~al.(2019)Devlin, Chang, Lee, and Toutanova}]{bert}
\bibinfo{author}{J.~Devlin}, \bibinfo{author}{M.-W. Chang},
  \bibinfo{author}{K.~Lee}, \bibinfo{author}{K.~Toutanova},
  \bibinfo{title}{Bert: Pre-training of deep bidirectional transformers for
  language understanding}, \bibinfo{year}{2019}.
  \href{http://arxiv.org/abs/1810.04805}{{\tt arXiv:1810.04805}}.
%Type = Article
\bibitem[{Brown et~al.(2020)Brown, Mann, Ryder, Subbiah, Kaplan, Dhariwal,
  Neelakantan, Shyam, Sastry, and Askell}]{GPT3}
\bibinfo{author}{T.~B. Brown}, \bibinfo{author}{B.~Mann},
  \bibinfo{author}{N.~Ryder}, \bibinfo{author}{M.~Subbiah},
  \bibinfo{author}{J.~Kaplan}, \bibinfo{author}{P.~Dhariwal},
  \bibinfo{author}{A.~Neelakantan}, \bibinfo{author}{P.~Shyam},
  \bibinfo{author}{G.~Sastry}, \bibinfo{author}{A.~e.~a. Askell},
\newblock \bibinfo{title}{Language models are few-shot learners},
\newblock \bibinfo{journal}{arXiv} \bibinfo{volume}{arXiv:2005.14165}
  (\bibinfo{year}{2020}).
%Type = Article
\bibitem[{Iyer et~al.(2022)Iyer, Lin, Pasunuru, Mihaylov, Simig, Yu, Shuster,
  Wang, Liu, and Koura}]{iyer2022opt}
\bibinfo{author}{S.~Iyer}, \bibinfo{author}{X.~V. Lin},
  \bibinfo{author}{R.~Pasunuru}, \bibinfo{author}{T.~Mihaylov},
  \bibinfo{author}{D.~Simig}, \bibinfo{author}{P.~Yu},
  \bibinfo{author}{K.~Shuster}, \bibinfo{author}{T.~Wang},
  \bibinfo{author}{Q.~Liu}, \bibinfo{author}{P.~S. e.~a. Koura},
\newblock \bibinfo{title}{{OPT-IML}: Scaling language model instruction meta
  learning through the lens of generalization},
\newblock \bibinfo{journal}{arXiv preprint arXiv:2212.12017}
  (\bibinfo{year}{2022}).
%Type = Misc
\bibitem[{OpenAI(2023)}]{gpt4}
\bibinfo{author}{OpenAI}, \bibinfo{title}{Gpt-4 technical report},
  \bibinfo{year}{2023}.
%Type = Misc
\bibitem[{Schreiner(2023)}]{gpt4-speculation}
\bibinfo{author}{M.~Schreiner}, \bibinfo{title}{Gpt-4 architecture, datasets,
  costs and more leaked}, \bibinfo{year}{2023}. \URLprefix
  \url{https://the-decoder.com/gpt-4-architecture-datasets-costs-and-more-leaked/}.
%Type = Inproceedings
\bibitem[{Gupta et~al.(2021)Gupta, Bharti, Nokhiz, and Karnick}]{sumpubmed}
\bibinfo{author}{V.~Gupta}, \bibinfo{author}{P.~Bharti},
  \bibinfo{author}{P.~Nokhiz}, \bibinfo{author}{H.~Karnick},
\newblock \bibinfo{title}{{SumPubMed}: Summarization dataset of {P}ub{M}ed
  scientific articles},
\newblock in: \bibinfo{booktitle}{Proceedings of the Annual Meeting of the
  Association for Computational Linguistics and the International Joint
  Conference on Natural Language Processing: Student Research Workshop},
  \bibinfo{publisher}{Association for Computational Linguistics},
  \bibinfo{year}{2021}, pp. \bibinfo{pages}{292--303}.
%Type = Inproceedings
\bibitem[{Kry{\'s}ci{\'n}ski et~al.(2022)Kry{\'s}ci{\'n}ski, Rajani, Agarwal,
  Xiong, and Radev}]{poland}
\bibinfo{author}{W.~Kry{\'s}ci{\'n}ski}, \bibinfo{author}{N.~Rajani},
  \bibinfo{author}{D.~Agarwal}, \bibinfo{author}{C.~Xiong},
  \bibinfo{author}{D.~Radev},
\newblock \bibinfo{title}{{BookSum}: A collection of datasets for long-form
  narrative summarization},
\newblock in: \bibinfo{booktitle}{Proceedings of the Findings of the
  Association for Computational Linguistics (ACL)}, \bibinfo{year}{2022}.
%Type = Inproceedings
\bibitem[{Yang et~al.(2016)Yang, Yang, Dyer, He, Smola, and Hovy}]{zichao}
\bibinfo{author}{Z.~Yang}, \bibinfo{author}{D.~Yang},
  \bibinfo{author}{C.~Dyer}, \bibinfo{author}{X.~He},
  \bibinfo{author}{A.~Smola}, \bibinfo{author}{E.~Hovy},
\newblock \bibinfo{title}{{Hierarchical Attention Networks} for document
  classification},
\newblock in: \bibinfo{booktitle}{Proceedings of the Conference of the North
  {A}merican Chapter of the Association for Computational Linguistics: Human
  Language Technologies}, \bibinfo{year}{2016}.
%Type = Article
\bibitem[{Beltagy et~al.(2020)Beltagy, Peters, and Cohan}]{longformer}
\bibinfo{author}{I.~Beltagy}, \bibinfo{author}{M.~Peters},
  \bibinfo{author}{A.~Cohan},
\newblock \bibinfo{title}{Longformer: The long-document {Transformer}},
\newblock \bibinfo{journal}{arXiv} \bibinfo{volume}{arXiv:2004.05150}
  (\bibinfo{year}{2020}).
%Type = Inproceedings
\bibitem[{Zaheer et~al.(2020)Zaheer, Guruganesh, Dubey, Ainslie, Alberti,
  Ontanon, Pham, Ravula, Wang, and Yang}]{big_bird}
\bibinfo{author}{M.~Zaheer}, \bibinfo{author}{G.~Guruganesh},
  \bibinfo{author}{K.~A. Dubey}, \bibinfo{author}{J.~Ainslie},
  \bibinfo{author}{C.~Alberti}, \bibinfo{author}{S.~Ontanon},
  \bibinfo{author}{P.~Pham}, \bibinfo{author}{A.~Ravula},
  \bibinfo{author}{Q.~Wang}, \bibinfo{author}{L.~e.~a. Yang},
\newblock \bibinfo{title}{{Big Bird: Transformers} for longer sequences},
\newblock in: \bibinfo{booktitle}{Proceedings of the Advances in Neural
  Information Processing Systems (NIPS)}, \bibinfo{year}{2020}.
%Type = Article
\bibitem[{Clement et~al.(2019)Clement, Bierbaum, O'Keeffe, and Alemi}]{clement}
\bibinfo{author}{C.~B. Clement}, \bibinfo{author}{M.~Bierbaum},
  \bibinfo{author}{K.~P. O'Keeffe}, \bibinfo{author}{A.~A. Alemi},
\newblock \bibinfo{title}{On the use of {arXiv} as a dataset},
\newblock \bibinfo{journal}{arXiv} \bibinfo{volume}{arXiv:1905.00075}
  (\bibinfo{year}{2019}).
%Type = Article
\bibitem[{Bjork et~al.(2009)Bjork, Roos, and Lauri}]{bjork}
\bibinfo{author}{B.-C. Bjork}, \bibinfo{author}{A.~Roos},
  \bibinfo{author}{M.~Lauri},
\newblock \bibinfo{title}{Scientific journal publishing: yearly volume and open
  access availability},
\newblock \bibinfo{journal}{Information Research: An International Electronic
  Journal} \bibinfo{volume}{14} (\bibinfo{year}{2009}).
%Type = Misc
\bibitem[{hug(date)}]{hugging_face_cnn_dailymail}
\bibinfo{title}{{Hugging Face CNN-DailyMail Dataset}},
  \bibinfo{howpublished}{\url{https://pan.webis.de/data.html\#pan-semeval-hyperpartisan-news-detection-19}},
  \bibinfo{year}{nodate}. \bibinfo{note}{Accessed: 2023-05-30}.
%Type = Misc
\bibitem[{20 Groups(date)}]{20groups}
20 Groups, \bibinfo{title}{20 newsgroups},
  \bibinfo{howpublished}{\url{http://qwone.com/~jason/20Newsgroups/}},
  \bibinfo{year}{nodate}. \bibinfo{note}{Accessed: 2023-04-30}.
%Type = Misc
\bibitem[{Hyperpartisan(date)}]{hyperpartisan}
Hyperpartisan, \bibinfo{title}{Hyperpartisan dataset},
  \bibinfo{howpublished}{\url{https://pan.webis.de/data.html\#pan-semeval-hyperpartisan-news-detection-19}},
  \bibinfo{year}{nodate}. \bibinfo{note}{Accessed: 2023-04-30}.
%Type = Article
\bibitem[{Bengio et~al.(2003)Bengio, Ducharme, Vincent, and
  Jauvin}]{bengio2003}
\bibinfo{author}{Y.~Bengio}, \bibinfo{author}{R.~Ducharme},
  \bibinfo{author}{P.~Vincent}, \bibinfo{author}{C.~Jauvin},
\newblock \bibinfo{title}{A neural probabilistic language model},
\newblock \bibinfo{journal}{Journal of Machine Learning Research}
  \bibinfo{volume}{3} (\bibinfo{year}{2003}) \bibinfo{pages}{1137--1155}.
%Type = Inproceedings
\bibitem[{Karamouzas et~al.(2022)Karamouzas, Mademlis, and
  Pitas}]{Karamouzas2022}
\bibinfo{author}{D.~Karamouzas}, \bibinfo{author}{I.~Mademlis},
  \bibinfo{author}{I.~Pitas},
\newblock \bibinfo{title}{Neural knowledge transfer for sentiment analysis in
  texts with figurative language},
\newblock in: \bibinfo{booktitle}{Proceedings of the IEEE International
  Workshop on Machine Learning for Signal Processing (MLSP)},
  \bibinfo{year}{2022}.
%Type = Misc
\bibitem[{I.(2020)}]{geroge}
\bibinfo{author}{T.~I.}, \bibinfo{title}{Development and application of
  language models in {Greek} literature}, \bibinfo{year}{2020}.
  \DOIprefix\doi{10.26262/heal.auth.ir.323350}.
%Type = Inproceedings
\bibitem[{Dai et~al.(2022)Dai, Chalkidis, Darkner, and Elliott}]{dai}
\bibinfo{author}{X.~Dai}, \bibinfo{author}{I.~Chalkidis},
  \bibinfo{author}{S.~Darkner}, \bibinfo{author}{D.~Elliott},
\newblock \bibinfo{title}{Revisiting {Transformer}-based models for long
  document classification},
\newblock in: \bibinfo{booktitle}{Proceedings of the Findings of the
  Association for Computational Linguistics (EMNLP)}, \bibinfo{year}{2022}.
%Type = Article
\bibitem[{Ouyang et~al.(2022)Ouyang, Wu, Jiang, Almeida, Wainwright, Mishkin,
  Zhang, Agarwal, Slama, and Ray}]{ouyang2022training}
\bibinfo{author}{L.~Ouyang}, \bibinfo{author}{J.~Wu},
  \bibinfo{author}{X.~Jiang}, \bibinfo{author}{D.~Almeida},
  \bibinfo{author}{C.~Wainwright}, \bibinfo{author}{P.~Mishkin},
  \bibinfo{author}{C.~Zhang}, \bibinfo{author}{S.~Agarwal},
  \bibinfo{author}{K.~Slama}, \bibinfo{author}{A.~e.~a. Ray},
\newblock \bibinfo{title}{Training language models to follow instructions with
  human feedback},
\newblock \bibinfo{journal}{Proceedings of the Advances in Neural Information
  Processing Systems (NIPS)}  (\bibinfo{year}{2022}).
%Type = Article
\bibitem[{Muennighoff et~al.(2022)Muennighoff, Wang, Sutawika, Roberts,
  Biderman, Scao, Bari, Shen, Yong, and
  Schoelkopf}]{muennighoff2022crosslingual}
\bibinfo{author}{N.~Muennighoff}, \bibinfo{author}{T.~Wang},
  \bibinfo{author}{L.~Sutawika}, \bibinfo{author}{A.~Roberts},
  \bibinfo{author}{S.~Biderman}, \bibinfo{author}{T.~L. Scao},
  \bibinfo{author}{M.~S. Bari}, \bibinfo{author}{S.~Shen},
  \bibinfo{author}{Z.-X. Yong}, \bibinfo{author}{H.~e.~a. Schoelkopf},
\newblock \bibinfo{title}{Cross-lingual generalization through multitask
  finetuning},
\newblock \bibinfo{journal}{arXiv preprint arXiv:2211.01786}
  (\bibinfo{year}{2022}).
%Type = Inproceedings
\bibitem[{Dai et~al.(2019)Dai, Yang, Yang, Carbonell, Le, and
  Salakhutdinov}]{dai-etal-2019-transformer}
\bibinfo{author}{Z.~Dai}, \bibinfo{author}{Z.~Yang}, \bibinfo{author}{Y.~Yang},
  \bibinfo{author}{J.~Carbonell}, \bibinfo{author}{Q.~Le},
  \bibinfo{author}{R.~Salakhutdinov},
\newblock \bibinfo{title}{Transformer-{XL}: Attentive language models beyond a
  fixed-length context},
\newblock in: \bibinfo{booktitle}{Proceedings of the Annual Meeting of the
  Association for Computational Linguistics}, \bibinfo{year}{2019}.
%Type = Inproceedings
\bibitem[{Chalkidis et~al.(2019)Chalkidis, Fergadiotis, Malakasiotis, and
  Androutsopoulos}]{ion}
\bibinfo{author}{C.~Chalkidis}, \bibinfo{author}{M.~Fergadiotis},
  \bibinfo{author}{P.~Malakasiotis}, \bibinfo{author}{I.~Androutsopoulos},
\newblock \bibinfo{title}{Large-scale multi-label text classification on {EU}
  legislation},
\newblock in: \bibinfo{booktitle}{Proceedings of the Annual Meeting of the
  Association for Computational Linguistics}, \bibinfo{year}{2019}.
%Type = Article
\bibitem[{Chalkidis et~al.(2020)Chalkidis, Fergadiotis, Kotitsas, Malakasiotis,
  Aletras, and Androutsopoulos}]{ion2}
\bibinfo{author}{I.~Chalkidis}, \bibinfo{author}{M.~Fergadiotis},
  \bibinfo{author}{S.~Kotitsas}, \bibinfo{author}{P.~Malakasiotis},
  \bibinfo{author}{N.~Aletras}, \bibinfo{author}{I.~Androutsopoulos},
\newblock \bibinfo{title}{An empirical study on large-scale multi-label text
  classification including few and zero-shot labels},
\newblock \bibinfo{journal}{Proceedings of the Conference on Empirical Methods
  in Natural Language Processing}  (\bibinfo{year}{2020}).
%Type = Article
\bibitem[{Zhang et~al.(2020)Zhang, Jia, Yin, Dong, Gao, and Hua}]{mcbert}
\bibinfo{author}{N.~Zhang}, \bibinfo{author}{Q.~Jia}, \bibinfo{author}{K.~Yin},
  \bibinfo{author}{L.~Dong}, \bibinfo{author}{F.~Gao},
  \bibinfo{author}{N.~Hua},
\newblock \bibinfo{title}{Conceptualized representation learning for chinese
  biomedical text mining},
\newblock \bibinfo{journal}{arXiv} \bibinfo{volume}{arXiv:2008.10813}
  (\bibinfo{year}{2020}).
%Type = Article
\bibitem[{Khadhraoui et~al.(2022)Khadhraoui, Bellaaj, Ammar, Hamam, and
  Jmaiel}]{ammar}
\bibinfo{author}{M.~Khadhraoui}, \bibinfo{author}{H.~Bellaaj},
  \bibinfo{author}{M.~B. Ammar}, \bibinfo{author}{H.~Hamam},
  \bibinfo{author}{M.~Jmaiel},
\newblock \bibinfo{title}{Survey of {BERT}-based models for scientific text
  classification: {COVID}-19 case study},
\newblock \bibinfo{journal}{Applied Sciences} \bibinfo{volume}{12}
  (\bibinfo{year}{2022}).
%Type = Article
\bibitem[{Lee et~al.(2019)Lee, Yoon, Kim, Kim, Kim, So, and Kang}]{so}
\bibinfo{author}{J.~Lee}, \bibinfo{author}{W.~Yoon}, \bibinfo{author}{S.~Kim},
  \bibinfo{author}{D.~Kim}, \bibinfo{author}{S.~Kim}, \bibinfo{author}{C.~H.
  So}, \bibinfo{author}{J.~Kang},
\newblock \bibinfo{title}{{BioBERT}: a pre-trained biomedical language
  representation model for biomedical text mining},
\newblock \bibinfo{journal}{Bioinformatics} \bibinfo{volume}{36}
  (\bibinfo{year}{2019}) \bibinfo{pages}{1234--1240}.
%Type = Article
\bibitem[{Chen et~al.(2022)Chen, Wu, Chen, Lu, and Ding}]{junhua}
\bibinfo{author}{H.~Chen}, \bibinfo{author}{L.~Wu}, \bibinfo{author}{J.~Chen},
  \bibinfo{author}{W.~Lu}, \bibinfo{author}{J.~Ding},
\newblock \bibinfo{title}{A comparative study of automated legal text
  classification using random forests and deep learning},
\newblock \bibinfo{journal}{Information Processing \& Management}
  \bibinfo{volume}{59} (\bibinfo{year}{2022}) \bibinfo{pages}{102798}.
%Type = Book
\bibitem[{Russell and Norvig(2016)}]{russel}
\bibinfo{author}{S.~Russell}, \bibinfo{author}{P.~Norvig},
  \bibinfo{title}{{Artificial Intelligence}: A Modern Approach},
  \bibinfo{publisher}{Pearson Education}, \bibinfo{address}{Upper Saddle River,
  New Jersey 07458}, \bibinfo{year}{2016}.
%Type = Article
\bibitem[{Cortes and Vapnik(1995)}]{cortes}
\bibinfo{author}{C.~Cortes}, \bibinfo{author}{V.~Vapnik},
\newblock \bibinfo{title}{Support-vector networks},
\newblock \bibinfo{journal}{Machine Learning} \bibinfo{volume}{20}
  (\bibinfo{year}{1995}) \bibinfo{pages}{273--297}.
%Type = Book
\bibitem[{Quinlan(1994)}]{quinlan}
\bibinfo{author}{J.~R. Quinlan}, \bibinfo{title}{S.L. C4.5: Programs for
  Machine Learning}, \bibinfo{publisher}{Morgan Kaufmann Publishers},
  \bibinfo{address}{Upper Saddle River, New Jersey 07458},
  \bibinfo{year}{1994}.
%Type = Article
\bibitem[{Xu et~al.(2012)Xu, Guo, Ye, and Cheng}]{xu}
\bibinfo{author}{B.~Xu}, \bibinfo{author}{X.~Guo}, \bibinfo{author}{Y.~Ye},
  \bibinfo{author}{J.~Cheng},
\newblock \bibinfo{title}{An improved random forest classifier for text
  categorization},
\newblock \bibinfo{journal}{Journal of King Saud University - Computer and
  Information Sciences} \bibinfo{volume}{7} (\bibinfo{year}{2012})
  \bibinfo{pages}{2913--2920}.
%Type = Misc
\bibitem[{Liu et~al.(2020)Liu, Huang, Li, Sun, Wu, and Zhang}]{sicong}
\bibinfo{author}{S.~Liu}, \bibinfo{author}{Z.~Huang}, \bibinfo{author}{Y.~Li},
  \bibinfo{author}{Z.~Sun}, \bibinfo{author}{J.~Wu},
  \bibinfo{author}{H.~Zhang}, \bibinfo{title}{{DeepGenre}: Deep neural networks
  for genre classification in literary works}, \bibinfo{year}{2020}.
%Type = Article
\bibitem[{Onan et~al.(2017)Onan, Korukoğlu, and Bulut}]{onan_2017}
\bibinfo{author}{A.~Onan}, \bibinfo{author}{S.~Korukoğlu},
  \bibinfo{author}{H.~Bulut},
\newblock \bibinfo{title}{A hybrid ensemble pruning approach based on consensus
  clustering and multi-objective evolutionary algorithm for sentiment
  classification},
\newblock \bibinfo{journal}{Information Processing \& Management}
  \bibinfo{volume}{53} (\bibinfo{year}{2017}) \bibinfo{pages}{814--833}.
%Type = Inproceedings
\bibitem[{Hsu et~al.(2020)Hsu, Chang, and Chang}]{hsu}
\bibinfo{author}{C.-C. Hsu}, \bibinfo{author}{P.-C. Chang},
  \bibinfo{author}{A.~Chang},
\newblock \bibinfo{title}{Multi-label classification of {ICD} coding using deep
  learning},
\newblock in: \bibinfo{booktitle}{Proceedings of the IEEE International
  Symposium on Community-centric Systems (CcS)}, \bibinfo{year}{2020}.
%Type = Article
\bibitem[{Haghighian~Roudsari et~al.(2022)Haghighian~Roudsari, Afshar, Lee, and
  Lee}]{patentnet}
\bibinfo{author}{A.~Haghighian~Roudsari}, \bibinfo{author}{J.~Afshar},
  \bibinfo{author}{W.~Lee}, \bibinfo{author}{S.~Lee},
\newblock \bibinfo{title}{{PatentNet}: multi-label classification of patent
  documents using deep learning-based language understanding},
\newblock \bibinfo{journal}{Scientometrics}  (\bibinfo{year}{2022})
  \bibinfo{pages}{1--25}.
%Type = Inproceedings
\bibitem[{Kim(2014)}]{kim}
\bibinfo{author}{Y.~Kim},
\newblock \bibinfo{title}{{Convolutional Neural Networks} for sentence
  classification},
\newblock in: \bibinfo{booktitle}{Proceedings of the Conference on Empirical
  Methods in Natural Language Processing ({EMNLP})}, \bibinfo{year}{2014}.
%Type = Article
\bibitem[{Li et~al.(2023)Li, Popa, Chagnon, Cinar, and Gaussier}]{li_2023}
\bibinfo{author}{M.~Li}, \bibinfo{author}{D.~N. Popa},
  \bibinfo{author}{J.~Chagnon}, \bibinfo{author}{Y.~G. Cinar},
  \bibinfo{author}{E.~Gaussier},
\newblock \bibinfo{title}{The power of selecting key blocks with local
  pre-ranking for long document information retrieval},
\newblock \bibinfo{journal}{ACM Transactions on Information Systems}
  \bibinfo{volume}{41} (\bibinfo{year}{2023}) \bibinfo{pages}{1–35}.
%Type = Inproceedings
\bibitem[{Ding et~al.(2020)Ding, Zhou, Yang, and Tang}]{ding_2020}
\bibinfo{author}{M.~Ding}, \bibinfo{author}{C.~Zhou},
  \bibinfo{author}{H.~Yang}, \bibinfo{author}{J.~Tang},
\newblock \bibinfo{title}{{CogLTX}: Applying {BERT} to long texts},
\newblock in: \bibinfo{booktitle}{Proceedings of the Advances in Neural
  Information Processing Systems (NIPS)}, \bibinfo{publisher}{Curran
  Associates, Inc.}, \bibinfo{year}{2020}, pp. \bibinfo{pages}{12792--12804}.
%Type = Inbook
\bibitem[{Amati(2009)}]{amati_2009}
\bibinfo{author}{G.~Amati}, \bibinfo{title}{BM25}, \bibinfo{publisher}{Springer
  US}, \bibinfo{address}{Boston, MA}, \bibinfo{year}{2009}, pp.
  \bibinfo{pages}{257--260}.
%Type = Article
\bibitem[{Li et~al.(2023)Li, Yates, MacAvaney, He, and Sun}]{li_2021_parade}
\bibinfo{author}{C.~Li}, \bibinfo{author}{A.~Yates},
  \bibinfo{author}{S.~MacAvaney}, \bibinfo{author}{B.~He},
  \bibinfo{author}{Y.~Sun},
\newblock \bibinfo{title}{{PARADE}: Passage representation aggregation for
  document reranking},
\newblock \bibinfo{journal}{ACM Transactions on Information Systems}
  \bibinfo{volume}{42} (\bibinfo{year}{2023}) \bibinfo{pages}{1--26}.
%Type = Article
\bibitem[{Child et~al.(2019)Child, Gray, Radford, and Sutskever}]{child}
\bibinfo{author}{R.~Child}, \bibinfo{author}{S.~Gray},
  \bibinfo{author}{A.~Radford}, \bibinfo{author}{I.~Sutskever},
\newblock \bibinfo{title}{Generating long sequences with {Sparse
  Transformers}},
\newblock \bibinfo{journal}{arXiv} \bibinfo{volume}{arXiv:1904.10509}
  (\bibinfo{year}{2019}).
%Type = Article
\bibitem[{Saeed et~al.(2020)Saeed, Wang, Peng, Hussain, and Nawaz}]{saeed}
\bibinfo{author}{H.~A. Saeed}, \bibinfo{author}{H.~Wang},
  \bibinfo{author}{M.~Peng}, \bibinfo{author}{A.~Hussain},
  \bibinfo{author}{A.~Nawaz},
\newblock \bibinfo{title}{Online fault monitoring based on {Deep Neural
  Network} \& sliding window technique},
\newblock \bibinfo{journal}{Progress in Nuclear Energy} \bibinfo{volume}{121}
  (\bibinfo{year}{2020}) \bibinfo{pages}{103236}.
%Type = Inproceedings
\bibitem[{Krystalakos et~al.(2018)Krystalakos, Nalmpantis, and
  Vrakas}]{odysseas}
\bibinfo{author}{O.~Krystalakos}, \bibinfo{author}{C.~Nalmpantis},
  \bibinfo{author}{D.~Vrakas},
\newblock \bibinfo{title}{Sliding window approach for online energy
  disaggregation using {Artificial Neural Networks}},
\newblock in: \bibinfo{booktitle}{Proceedings of the Hellenic Conference on
  Artificial Intelligence (SETN)}, \bibinfo{year}{2018}.
%Type = Article
\bibitem[{Choromanski et~al.(2020)Choromanski, Likhosherstov, Dohan, Song,
  Gane, Sarl{\'{o}}s, Hawkins, Davis, Mohiuddin, Kaiser, Belanger, Colwell, and
  Weller}]{performer}
\bibinfo{author}{K.~Choromanski}, \bibinfo{author}{V.~Likhosherstov},
  \bibinfo{author}{D.~Dohan}, \bibinfo{author}{X.~Song},
  \bibinfo{author}{A.~Gane}, \bibinfo{author}{T.~Sarl{\'{o}}s},
  \bibinfo{author}{P.~Hawkins}, \bibinfo{author}{J.~Davis},
  \bibinfo{author}{A.~Mohiuddin}, \bibinfo{author}{L.~Kaiser},
  \bibinfo{author}{D.~Belanger}, \bibinfo{author}{L.~J. Colwell},
  \bibinfo{author}{A.~Weller},
\newblock \bibinfo{title}{Rethinking attention with performers},
\newblock \bibinfo{journal}{CoRR} \bibinfo{volume}{abs/2009.14794}
  (\bibinfo{year}{2020}). \URLprefix \url{https://arxiv.org/abs/2009.14794}.
  \href{http://arxiv.org/abs/2009.14794}{{\tt arXiv:2009.14794}}.
%Type = Misc
\bibitem[{Chen(2021)}]{permuteformer}
\bibinfo{author}{P.~Chen}, \bibinfo{title}{Permuteformer: Efficient relative
  position encoding for long sequences}, \bibinfo{year}{2021}.
  \href{http://arxiv.org/abs/2109.02377}{{\tt arXiv:2109.02377}}.
%Type = Inproceedings
\bibitem[{Chalkidis et~al.(2019)Chalkidis, Androutsopoulos, and
  Aletras}]{ion_han}
\bibinfo{author}{I.~Chalkidis}, \bibinfo{author}{I.~Androutsopoulos},
  \bibinfo{author}{N.~Aletras},
\newblock \bibinfo{title}{Neural legal judgment prediction in {E}nglish},
\newblock in: \bibinfo{booktitle}{Proceedings of the Annual Meeting of the
  Association for Computational Linguistics (ACL)}, \bibinfo{year}{2019}.
%Type = Article
\bibitem[{Chalkidis et~al.(2021)Chalkidis, Jana, Hartung, Bommarito,
  Androutsopoulos, Katz, and Aletras}]{glue_gunner}
\bibinfo{author}{I.~Chalkidis}, \bibinfo{author}{A.~Jana},
  \bibinfo{author}{D.~Hartung}, \bibinfo{author}{M.~Bommarito},
  \bibinfo{author}{I.~Androutsopoulos}, \bibinfo{author}{D.~Katz},
  \bibinfo{author}{N.~Aletras},
\newblock \bibinfo{title}{{LexGLUE}: {A} benchmark dataset for legal language
  understanding in {English}}  (\bibinfo{year}{2021}).
%Type = Inproceedings
\bibitem[{Chen et~al.(2020)Chen, Sun, Yang, and Lin}]{yanguang}
\bibinfo{author}{Y.~Chen}, \bibinfo{author}{Y.~Sun}, \bibinfo{author}{Z.~Yang},
  \bibinfo{author}{H.~Lin},
\newblock \bibinfo{title}{Joint entity and relation extraction for legal
  documents with legal feature enhancement},
\newblock in: \bibinfo{booktitle}{Proceedings of the International Conference
  on Computational Linguistics}, \bibinfo{year}{2020}.
%Type = Inproceedings
\bibitem[{Chalkidis et~al.(2020)Chalkidis, Fergadiotis, Malakasiotis, Aletras,
  and Androutsopoulos}]{ion6}
\bibinfo{author}{I.~Chalkidis}, \bibinfo{author}{M.~Fergadiotis},
  \bibinfo{author}{P.~Malakasiotis}, \bibinfo{author}{N.~Aletras},
  \bibinfo{author}{I.~Androutsopoulos},
\newblock \bibinfo{title}{{LEGAL}-{BERT}: The {Muppets} straight out of law
  school},
\newblock in: \bibinfo{booktitle}{Proceedings of the Conference on Empirical
  Methods in Natural Language Processing (EMNLP)}, \bibinfo{year}{2020}.
%Type = Inproceedings
\bibitem[{Zheng et~al.(2021)Zheng, Guha, Anderson, Henderson, and Ho}]{zheng}
\bibinfo{author}{L.~Zheng}, \bibinfo{author}{N.~Guha}, \bibinfo{author}{B.~R.
  Anderson}, \bibinfo{author}{P.~Henderson}, \bibinfo{author}{D.~E. Ho},
\newblock \bibinfo{title}{When does pretraining help? assessing self-supervised
  learning for law and the casehold dataset of 53,000+ legal holdings},
\newblock in: \bibinfo{booktitle}{Proceedings of the International Conference
  on Artificial Intelligence and Law}, \bibinfo{year}{2021}.
%Type = Inproceedings
\bibitem[{Khandve et~al.(2022)Khandve, Wagh, Wani, Joshi, and Joshi}]{khandve}
\bibinfo{author}{S.~I. Khandve}, \bibinfo{author}{V.~K. Wagh},
  \bibinfo{author}{A.~D. Wani}, \bibinfo{author}{I.~M. Joshi},
  \bibinfo{author}{R.~B. Joshi},
\newblock \bibinfo{title}{Hierarchical neural network approaches for long
  document classification},
\newblock in: \bibinfo{booktitle}{Proceedings of the International Conference
  on Machine Learning and Computing (ICMLC)}, \bibinfo{year}{2022}.
%Type = Article
\bibitem[{Cer et~al.(2018)Cer, Yang, Kong, Hua, Limtiaco, John, Constant,
  Guajardo-Cespedes, Yuan, and Tar}]{use}
\bibinfo{author}{D.~Cer}, \bibinfo{author}{Y.~Yang}, \bibinfo{author}{S.-Y.
  Kong}, \bibinfo{author}{N.~Hua}, \bibinfo{author}{N.~Limtiaco},
  \bibinfo{author}{R.~S. John}, \bibinfo{author}{N.~Constant},
  \bibinfo{author}{M.~Guajardo-Cespedes}, \bibinfo{author}{S.~Yuan},
  \bibinfo{author}{C.~e.~a. Tar},
\newblock \bibinfo{title}{Universal sentence encoder},
\newblock \bibinfo{journal}{arXiv} \bibinfo{volume}{arXiv:1803.11175}
  (\bibinfo{year}{2018}).
%Type = Inproceedings
\bibitem[{Wu et~al.(2021)Wu, Wu, Qi, and Huang}]{qi}
\bibinfo{author}{C.~Wu}, \bibinfo{author}{F.~Wu}, \bibinfo{author}{T.~Qi},
  \bibinfo{author}{Y.~Huang},
\newblock \bibinfo{title}{{Hi-Transformer}: Hierarchical interactive
  {Transformer} for efficient and effective long document modeling},
\newblock in: \bibinfo{booktitle}{Proceedings of the Annual Meeting of the
  Association for Computational Linguistics and the International Joint
  Conference on Natural Language Processing}, \bibinfo{year}{2021}.
%Type = Misc
\bibitem[{Liu et~al.(2022)Liu, Liu, Chen, Lu, Feng, Feng, Sun, Tian, Wu, and
  Wang}]{liu}
\bibinfo{author}{Y.~Liu}, \bibinfo{author}{J.~Liu}, \bibinfo{author}{L.~Chen},
  \bibinfo{author}{Y.~Lu}, \bibinfo{author}{S.~Feng},
  \bibinfo{author}{Z.~Feng}, \bibinfo{author}{Y.~Sun},
  \bibinfo{author}{H.~Tian}, \bibinfo{author}{H.~Wu},
  \bibinfo{author}{H.~Wang}, \bibinfo{title}{{ERNIE-SPARSE}: Learning
  hierarchical efficient {Transformer} through regularized self-attention},
  \bibinfo{year}{2022}.
%Type = Article
\bibitem[{Onan(2023)}]{onan_hier}
\bibinfo{author}{A.~Onan},
\newblock \bibinfo{title}{Hierarchical graph-based text classification
  framework with contextual node embedding and {BERT}-based dynamic fusion},
\newblock \bibinfo{journal}{Journal of King Saud University - Computer and
  Information Sciences} \bibinfo{volume}{35} (\bibinfo{year}{2023})
  \bibinfo{pages}{101610}.
%Type = Article
\bibitem[{Miller(1995)}]{wordnet}
\bibinfo{author}{G.~A. Miller},
\newblock \bibinfo{title}{{WordNet}: A lexical database for {English}},
\newblock \bibinfo{journal}{Communications of the ACM} \bibinfo{volume}{38}
  (\bibinfo{year}{1995}) \bibinfo{pages}{39–41}.
%Type = Article
\bibitem[{Bellman and Kalaba(1959)}]{dtw}
\bibinfo{author}{R.~Bellman}, \bibinfo{author}{R.~E. Kalaba},
\newblock \bibinfo{title}{On adaptive control processes},
\newblock \bibinfo{journal}{IRE Transactions on Automatic Control}
  \bibinfo{volume}{4} (\bibinfo{year}{1959}) \bibinfo{pages}{1--9}.
%Type = Article
\bibitem[{Zhang et~al.(2021)Zhang, Chang, Yu, and Dhillon}]{XR-Transformer}
\bibinfo{author}{J.~Zhang}, \bibinfo{author}{W.~Chang},
  \bibinfo{author}{H.~Yu}, \bibinfo{author}{I.~S. Dhillon},
\newblock \bibinfo{title}{Fast multi-resolution {Transformer} fine-tuning for
  extreme multi-label text classification},
\newblock \bibinfo{journal}{CoRR} \bibinfo{volume}{abs/2110.00685}
  (\bibinfo{year}{2021}). \URLprefix \url{https://arxiv.org/abs/2110.00685}.
  \href{http://arxiv.org/abs/2110.00685}{{\tt arXiv:2110.00685}}.
%Type = Inproceedings
\bibitem[{Ding et~al.(2021)Ding, Shang, Wang, Sun, Tian, Wu, and
  Wang}]{ernie-doc}
\bibinfo{author}{S.~Ding}, \bibinfo{author}{J.~Shang},
  \bibinfo{author}{S.~Wang}, \bibinfo{author}{Y.~Sun},
  \bibinfo{author}{H.~Tian}, \bibinfo{author}{H.~Wu},
  \bibinfo{author}{H.~Wang},
\newblock \bibinfo{title}{{ERNIE}-{D}oc: A retrospective long-document modeling
  transformer},
\newblock in: \bibinfo{booktitle}{Proceedings of the Annual Meeting of the
  Association for Computational Linguistics and the International Joint
  Conference on Natural Language Processing}, \bibinfo{year}{2021}.
%Type = Article
\bibitem[{Cho et~al.(2014)Cho, van Merrienboer, G{\"{u}}l{\c{c}}ehre, Bougares,
  Schwenk, and Bengio}]{cho}
\bibinfo{author}{K.~Cho}, \bibinfo{author}{B.~van Merrienboer},
  \bibinfo{author}{C.~G{\"{u}}l{\c{c}}ehre}, \bibinfo{author}{F.~Bougares},
  \bibinfo{author}{H.~Schwenk}, \bibinfo{author}{y.~Bengio},
\newblock \bibinfo{title}{Learning phrase representations using {RNN}
  encoder-decoder for statistical machine translation},
\newblock \bibinfo{journal}{Proceedings of the Conference on Empirical Methods
  in Natural Language Processing (EMNLP)}  (\bibinfo{year}{2014}).
%Type = Inproceedings
\bibitem[{Sutskever et~al.(2014)Sutskever, Vinyals, and Le}]{seq2seq}
\bibinfo{author}{I.~Sutskever}, \bibinfo{author}{O.~Vinyals},
  \bibinfo{author}{Q.~V. Le},
\newblock \bibinfo{title}{Sequence to sequence learning with neural networks},
\newblock in: \bibinfo{editor}{Z.~Ghahramani}, \bibinfo{editor}{M.~Welling},
  \bibinfo{editor}{C.~Cortes}, \bibinfo{editor}{N.~Lawrence},
  \bibinfo{editor}{K.~Weinberger} (Eds.), \bibinfo{booktitle}{Proceedings of
  the Advances in Neural Information Processing Systems (NIPS)},
  \bibinfo{year}{2014}.
%Type = Inproceedings
\bibitem[{Xiao and Carenini(2019)}]{xiao}
\bibinfo{author}{W.~Xiao}, \bibinfo{author}{G.~Carenini},
\newblock \bibinfo{title}{Extractive summarization of long documents by
  combining global and local context},
\newblock in: \bibinfo{booktitle}{Proceedings of the Conference on Empirical
  Methods in Natural Language Processing and the International Joint Conference
  on Natural Language Processing (EMNLP-IJCNLP)},
  \bibinfo{publisher}{Association for Computational Linguistics},
  \bibinfo{year}{2019}.
%Type = Inproceedings
\bibitem[{Lin(2004)}]{rouge}
\bibinfo{author}{C.-Y. Lin},
\newblock \bibinfo{title}{{ROUGE}: A package for automatic evaluation of
  summaries},
\newblock in: \bibinfo{booktitle}{Text Summarization Branches Out},
  \bibinfo{publisher}{Association for Computational Linguistics},
  \bibinfo{year}{2004}.
%Type = Inproceedings
\bibitem[{Akter et~al.(2022)Akter, Bansal, and Karmaker}]{akter}
\bibinfo{author}{M.~Akter}, \bibinfo{author}{N.~Bansal}, \bibinfo{author}{S.~K.
  Karmaker},
\newblock \bibinfo{title}{Revisiting automatic evaluation of extractive
  summarization task: Can we do better than {ROUGE}?},
\newblock in: \bibinfo{booktitle}{Proceedings of the Findings of the
  Association for Computational Linguistics (ACL)}, \bibinfo{year}{2022}.
%Type = Misc
\bibitem[{Zhong et~al.(2020)Zhong, Liu, Chen, Wang, Qiu, and Huang}]{matchsum}
\bibinfo{author}{M.~Zhong}, \bibinfo{author}{P.~Liu},
  \bibinfo{author}{Y.~Chen}, \bibinfo{author}{D.~Wang},
  \bibinfo{author}{X.~Qiu}, \bibinfo{author}{X.~Huang},
  \bibinfo{title}{Extractive summarization as text matching},
  \bibinfo{year}{2020}.
%Type = Article
\bibitem[{See et~al.(2017)See, Liu, and Manning}]{abigail}
\bibinfo{author}{A.~See}, \bibinfo{author}{P.~J. Liu}, \bibinfo{author}{C.~D.
  Manning},
\newblock \bibinfo{title}{Get to the point: Summarization with
  {Pointer-Generator Networks}},
\newblock \bibinfo{journal}{arXiv preprint arXiv:1704.04368}
  (\bibinfo{year}{2017}).
%Type = Article
\bibitem[{Suleiman and Awajan(2020)}]{suleiman}
\bibinfo{author}{D.~Suleiman}, \bibinfo{author}{A.~Awajan},
\newblock \bibinfo{title}{Deep learning-based abstractive text summarization:
  approaches, datasets, evaluation measures, and challenges},
\newblock \bibinfo{journal}{Mathematical problems in engineering}
  \bibinfo{volume}{2020} (\bibinfo{year}{2020}) \bibinfo{pages}{1--29}.
%Type = Inproceedings
\bibitem[{Wan et~al.(2023)Wan, Liu, McKeown, Dreyer, and
  Bansal}]{Wan2023faithfulness}
\bibinfo{author}{D.~Wan}, \bibinfo{author}{M.~Liu},
  \bibinfo{author}{K.~McKeown}, \bibinfo{author}{M.~Dreyer},
  \bibinfo{author}{M.~Bansal},
\newblock \bibinfo{title}{Faithfulness-aware decoding strategies for
  abstractive summarization},
\newblock in: \bibinfo{booktitle}{Proceedings of the Conference of the European
  Chapter of the Association for Computational Linguistics},
  \bibinfo{year}{2023}.
%Type = Inproceedings
\bibitem[{van~der Poel et~al.(2022)van~der Poel, Cotterell, and
  Meister}]{van2022mutual}
\bibinfo{author}{L.~van~der Poel}, \bibinfo{author}{R.~Cotterell},
  \bibinfo{author}{C.~Meister},
\newblock \bibinfo{title}{Mutual information alleviates hallucinations in
  abstractive summarization},
\newblock in: \bibinfo{booktitle}{Proceedings of the ACL Conference on
  Empirical Methods in Natural Language Processing}, \bibinfo{year}{2022}.
%Type = Inproceedings
\bibitem[{Graham(2015)}]{graham}
\bibinfo{author}{Y.~Graham},
\newblock \bibinfo{title}{Re-evaluating automatic summarization with {BLEU} and
  192 shades of {ROUGE}},
\newblock in: \bibinfo{booktitle}{Proceedings of the Conference on Empirical
  Methods in Natural Language Processing (EMNLP)}, \bibinfo{year}{2015}.
%Type = Inproceedings
\bibitem[{Pitler et~al.(2010)Pitler, Louis, and Nenkova}]{pitler}
\bibinfo{author}{E.~Pitler}, \bibinfo{author}{A.~Louis},
  \bibinfo{author}{A.~Nenkova},
\newblock \bibinfo{title}{Automatic evaluation of linguistic quality in
  multi-document summarization},
\newblock in: \bibinfo{booktitle}{Proceedings of the Annual Meeting of the
  Association for Computational Linguistics}, \bibinfo{year}{2010}.
%Type = Article
\bibitem[{Fabbri et~al.(2021)Fabbri, Kry{\'s}ci{\'n}ski, McCann, Xiong, Socher,
  and Radev}]{fabbri}
\bibinfo{author}{A.~R. Fabbri}, \bibinfo{author}{W.~Kry{\'s}ci{\'n}ski},
  \bibinfo{author}{B.~McCann}, \bibinfo{author}{C.~Xiong},
  \bibinfo{author}{R.~Socher}, \bibinfo{author}{D.~Radev},
\newblock \bibinfo{title}{{SummEval}: Re-evaluating summarization evaluation},
\newblock \bibinfo{journal}{Transactions of the Association for Computational
  Linguistics} \bibinfo{volume}{9} (\bibinfo{year}{2021})
  \bibinfo{pages}{391--409}.
%Type = Inproceedings
\bibitem[{Kryscinski et~al.(2020)Kryscinski, McCann, Xiong, and
  Socher}]{Kry2020}
\bibinfo{author}{W.~Kryscinski}, \bibinfo{author}{B.~McCann},
  \bibinfo{author}{C.~Xiong}, \bibinfo{author}{R.~Socher},
\newblock \bibinfo{title}{Evaluating the factual consistency of abstractive
  text summarization},
\newblock in: \bibinfo{booktitle}{Proceedings of the Conference on Empirical
  Methods in Natural Language Processing}, \bibinfo{year}{2020}.
%Type = Inproceedings
\bibitem[{Banerjee and Lavie(2005)}]{meteor}
\bibinfo{author}{S.~Banerjee}, \bibinfo{author}{A.~Lavie},
\newblock \bibinfo{title}{{METEOR}: An automatic metric for {MT} evaluation
  with improved correlation with human judgments},
\newblock in: \bibinfo{booktitle}{Proceedings of the {ACL} Workshop on
  Intrinsic and Extrinsic Evaluation Measures for Machine Translation and/or
  Summarization}, \bibinfo{publisher}{Association for Computational
  Linguistics}, \bibinfo{year}{2005}.
%Type = Inproceedings
\bibitem[{Zhang et~al.(2020)Zhang, Kishore, Wu, Weinberger, and
  Artzi}]{bert_score}
\bibinfo{author}{T.~Zhang}, \bibinfo{author}{V.~Kishore},
  \bibinfo{author}{F.~Wu}, \bibinfo{author}{K.~Q. Weinberger},
  \bibinfo{author}{Y.~Artzi},
\newblock \bibinfo{title}{{BERTScore}: Evaluating text generation with {BERT}},
\newblock in: \bibinfo{booktitle}{Proceedings of the International Conference
  on Learning Representations (ICLR)}, \bibinfo{year}{2020}.
%Type = Inproceedings
\bibitem[{Peyrard et~al.(2017)Peyrard, Botschen, and Gurevych}]{s3}
\bibinfo{author}{M.~Peyrard}, \bibinfo{author}{T.~Botschen},
  \bibinfo{author}{I.~Gurevych},
\newblock \bibinfo{title}{Learning to score system summaries for better content
  selection evaluation},
\newblock in: \bibinfo{booktitle}{Proceedings of the Workshop on New Frontiers
  in Summarization}, \bibinfo{year}{2017}.
%Type = Article
\bibitem[{Vasilyev et~al.(2020)Vasilyev, Dharnidharka, and Bohannon}]{blanc}
\bibinfo{author}{O.~Vasilyev}, \bibinfo{author}{V.~Dharnidharka},
  \bibinfo{author}{J.~Bohannon},
\newblock \bibinfo{title}{Fill in the {BLANC:} human-free quality estimation of
  document summaries}  (\bibinfo{year}{2020}).
%Type = Misc
\bibitem[{Yuan et~al.(2021)Yuan, Neubig, and Liu}]{bart_score}
\bibinfo{author}{W.~Yuan}, \bibinfo{author}{G.~Neubig},
  \bibinfo{author}{P.~Liu}, \bibinfo{title}{Bartscore: Evaluating generated
  text as text generation}, \bibinfo{year}{2021}.
  \href{http://arxiv.org/abs/2106.11520}{{\tt arXiv:2106.11520}}.
%Type = Article
\bibitem[{Rahimi et~al.(2023)Rahimi, Hoover, Mimno, Naacke, Constantin, and
  Amann}]{CTC}
\bibinfo{author}{H.~Rahimi}, \bibinfo{author}{J.~L. Hoover},
  \bibinfo{author}{D.~Mimno}, \bibinfo{author}{H.~Naacke},
  \bibinfo{author}{C.~Constantin}, \bibinfo{author}{B.~Amann},
\newblock \bibinfo{title}{Contextualized topic coherence metrics},
\newblock \bibinfo{journal}{arXiv preprint 2305.14587}  (\bibinfo{year}{2023}).
%Type = Article
\bibitem[{Laban et~al.(2022)Laban, Schnabel, Bennett, and
  Hearst}]{Laban2022summac}
\bibinfo{author}{P.~Laban}, \bibinfo{author}{T.~Schnabel},
  \bibinfo{author}{P.~N. Bennett}, \bibinfo{author}{M.~A. Hearst},
\newblock \bibinfo{title}{{SummaC}: Re-visiting {NLI}-based models for
  inconsistency detection in summarization},
\newblock \bibinfo{journal}{Transactions of the Association for Computational
  Linguistics} \bibinfo{volume}{10} (\bibinfo{year}{2022})
  \bibinfo{pages}{163--177}.
%Type = Inproceedings
\bibitem[{Jeretic et~al.(2020)Jeretic, Warstadt, Bhooshan, and
  Williams}]{Jeretic2020}
\bibinfo{author}{P.~Jeretic}, \bibinfo{author}{A.~Warstadt},
  \bibinfo{author}{S.~Bhooshan}, \bibinfo{author}{A.~Williams},
\newblock \bibinfo{title}{Are {Natural Language Inference} models {IMPPRESsive?
  Learning IMPlicature and PRESupposition}},
\newblock in: \bibinfo{booktitle}{Proceedings of the Annual Meeting of the
  Association for Computational Linguistics}, \bibinfo{year}{2020}.
%Type = Inproceedings
\bibitem[{Deng et~al.(2021)Deng, Tan, Liu, Xing, and Hu}]{Deng2021compression}
\bibinfo{author}{M.~Deng}, \bibinfo{author}{B.~Tan}, \bibinfo{author}{Z.~Liu},
  \bibinfo{author}{E.~P. Xing}, \bibinfo{author}{Z.~Hu},
\newblock \bibinfo{title}{Compression, transduction, and creation: A unified
  framework for evaluating natural language generation},
\newblock in: \bibinfo{booktitle}{Proceedings of the Conference on Empirical
  Methods in Natural Language Processing (EMNLP)}, \bibinfo{year}{2021}.
%Type = Inproceedings
\bibitem[{Cheng and Lapata(2016)}]{lapata}
\bibinfo{author}{J.~Cheng}, \bibinfo{author}{M.~Lapata},
\newblock \bibinfo{title}{Neural summarization by extracting sentences and
  words},
\newblock in: \bibinfo{booktitle}{Proceedings of the Annual Meeting of the
  Association for Computational Linguistics}, \bibinfo{publisher}{Association
  for Computational Linguistics}, \bibinfo{year}{2016}.
%Type = Inproceedings
\bibitem[{Nallapati et~al.(2016)Nallapati, Zhai, and Zhou}]{nallapati}
\bibinfo{author}{R.~Nallapati}, \bibinfo{author}{F.~Zhai},
  \bibinfo{author}{B.~Zhou},
\newblock \bibinfo{title}{{SummaRuNNer}: {A} {Recurrent Neural Network}-based
  sequence model for extractive summarization of documents},
\newblock in: \bibinfo{booktitle}{Proceedings of the AAAI Conference on
  Artificial Intelligence}, \bibinfo{year}{2016}.
%Type = Inproceedings
\bibitem[{Wang and Chang(2016)}]{wang}
\bibinfo{author}{W.~Wang}, \bibinfo{author}{B.~Chang},
\newblock \bibinfo{title}{Graph-based dependency parsing with bidirectional
  {LSTM}},
\newblock in: \bibinfo{booktitle}{Proceedings of the Annual Meeting of the
  Association for Computational Linguistics (Volume 1: Long Papers)},
  \bibinfo{publisher}{Association for Computational Linguistics},
  \bibinfo{year}{2016}.
%Type = Inproceedings
\bibitem[{Liu and Lapata(2019)}]{lapata_bert}
\bibinfo{author}{Y.~Liu}, \bibinfo{author}{M.~Lapata},
\newblock \bibinfo{title}{Text summarization with pretrained encoders},
\newblock in: \bibinfo{booktitle}{Proceedings of the Conference on Empirical
  Methods in Natural Language Processing and the International Joint Conference
  on Natural Language Processing (EMNLP-IJCNLP)}, \bibinfo{year}{2019}.
%Type = Article
\bibitem[{Ruan et~al.(2022)Ruan, Ostendorff, and Rehm}]{histruct}
\bibinfo{author}{Q.~Ruan}, \bibinfo{author}{M.~Ostendorff},
  \bibinfo{author}{G.~Rehm},
\newblock \bibinfo{title}{{HiStruct+}: Improving extractive text summarization
  with hierarchical structure information},
\newblock \bibinfo{journal}{arXiv} \bibinfo{volume}{arXiv:2203.09629}
  (\bibinfo{year}{2022}).
%Type = Inproceedings
\bibitem[{Grail et~al.(2021)Grail, Perez, and Gaussier}]{global_BERT}
\bibinfo{author}{Q.~Grail}, \bibinfo{author}{J.~Perez},
  \bibinfo{author}{E.~Gaussier},
\newblock \bibinfo{title}{Globalizing {BERT}-based {T}ransformer architectures
  for long document summarization},
\newblock in: \bibinfo{booktitle}{Proceedings of the Conference of the European
  Chapter of the Association for Computational Linguistics},
  \bibinfo{publisher}{Association for Computational Linguistics},
  \bibinfo{year}{2021}.
%Type = Inproceedings
\bibitem[{Zhang et~al.(2022)Zhang, Liu, and Zhang}]{hegel}
\bibinfo{author}{H.~Zhang}, \bibinfo{author}{X.~Liu},
  \bibinfo{author}{J.~Zhang},
\newblock \bibinfo{title}{{HEGEL: Hypergraph Transformer} for long document
  summarization},
\newblock in: \bibinfo{booktitle}{Proceedings of the Conference on Empirical
  Methods in Natural Language Processing (EMNLP)}, \bibinfo{year}{2022}.
%Type = Article
\bibitem[{Blei et~al.(2003)Blei, Ng, and Jordan}]{dirichet}
\bibinfo{author}{D.~M. Blei}, \bibinfo{author}{A.~Y. Ng},
  \bibinfo{author}{M.~I. Jordan},
\newblock \bibinfo{title}{{Latent Dirichlet Allocation}},
\newblock \bibinfo{journal}{Journal of Machine Learning Research}
  \bibinfo{volume}{3} (\bibinfo{year}{2003}) \bibinfo{pages}{993--1022}.
%Type = Misc
\bibitem[{Grootendorst(2020)}]{keybert}
\bibinfo{author}{M.~Grootendorst}, \bibinfo{title}{{KeyBERT}: Minimal keyword
  extraction with {BERT}}, \bibinfo{year}{2020}. \URLprefix
  \url{https://doi.org/10.5281/zenodo.4461265}.
  \DOIprefix\doi{10.5281/zenodo.4461265}.
%Type = Inproceedings
\bibitem[{Li et~al.(2018)Li, Xiao, Lyu, and Wang}]{wei2}
\bibinfo{author}{W.~Li}, \bibinfo{author}{X.~Xiao}, \bibinfo{author}{Y.~Lyu},
  \bibinfo{author}{Y.~Wang},
\newblock \bibinfo{title}{Improving neural abstractive document summarization
  with explicit information selection modeling},
\newblock in: \bibinfo{booktitle}{Proceedings of the Conference on Empirical
  Methods in Natural Language Processing (EMNLP)}, \bibinfo{year}{2018}.
%Type = Inproceedings
\bibitem[{Gehrmann et~al.(2018)Gehrmann, Deng, and Rush}]{gehrmann}
\bibinfo{author}{S.~Gehrmann}, \bibinfo{author}{Y.~Deng},
  \bibinfo{author}{A.~Rush},
\newblock \bibinfo{title}{Bottom-up abstractive summarization},
\newblock in: \bibinfo{booktitle}{Proceedings of the Conference on Empirical
  Methods in Natural Language Processing (EMNLP)}, \bibinfo{year}{2018}.
%Type = Inproceedings
\bibitem[{Zhang et~al.(2020)Zhang, Zhao, Saleh, and Liu}]{pegasus}
\bibinfo{author}{J.~Zhang}, \bibinfo{author}{Y.~Zhao},
  \bibinfo{author}{M.~Saleh}, \bibinfo{author}{P.~Liu},
\newblock \bibinfo{title}{{PEGASUS}: Pre-training with extracted gap-sentences
  for abstractive summarization},
\newblock in: \bibinfo{booktitle}{Proceedings of the International Conference
  on Machine Learning (ICML)}, \bibinfo{organization}{PMLR},
  \bibinfo{year}{2020}.
%Type = Inproceedings
\bibitem[{Guo et~al.(2022)Guo, Ainslie, Uthus, Ontanon, Ni, Sung, and
  Yang}]{Guo22LongT5}
\bibinfo{author}{M.~Guo}, \bibinfo{author}{J.~Ainslie},
  \bibinfo{author}{D.~Uthus}, \bibinfo{author}{S.~Ontanon},
  \bibinfo{author}{J.~Ni}, \bibinfo{author}{Y.-H. Sung},
  \bibinfo{author}{Y.~Yang},
\newblock \bibinfo{title}{{L}ong{T}5: {E}fficient text-to-text transformer for
  long sequences},
\newblock in: \bibinfo{booktitle}{Proceedings of the Annual Conference of the
  North American Chapter of the Association for Computational Linguistics
  (NAAC)}, \bibinfo{year}{2022}.
%Type = Article
\bibitem[{Moro et~al.(2023)Moro, Ragazzi, Valgimigli, Frisoni, Sartori, and
  Marfia}]{luca}
\bibinfo{author}{G.~Moro}, \bibinfo{author}{L.~Ragazzi},
  \bibinfo{author}{L.~Valgimigli}, \bibinfo{author}{G.~Frisoni},
  \bibinfo{author}{C.~Sartori}, \bibinfo{author}{G.~Marfia},
\newblock \bibinfo{title}{Efficient memory-enhanced {Transformer} for
  long-document summarization in low-resource regimes},
\newblock \bibinfo{journal}{Sensors} \bibinfo{volume}{23}
  (\bibinfo{year}{2023}).
%Type = Inproceedings
\bibitem[{Pang et~al.(2023)Pang, Nijkamp, Kry{\'s}ci{\'n}ski, Savarese, Zhou,
  and Xiong}]{pan_bottom}
\bibinfo{author}{B.~Pang}, \bibinfo{author}{E.~Nijkamp},
  \bibinfo{author}{W.~Kry{\'s}ci{\'n}ski}, \bibinfo{author}{S.~Savarese},
  \bibinfo{author}{Y.~Zhou}, \bibinfo{author}{C.~Xiong},
\newblock \bibinfo{title}{Long document summarization with top-down and
  bottom-up inference},
\newblock in: \bibinfo{booktitle}{Proceedings of the Findings of the
  Association for Computational Linguistics (ACL)}, \bibinfo{year}{2023}.
%Type = Inproceedings
\bibitem[{Victor et~al.(2022)Victor, Albert, Colin, Stephen, Lintang, Zaid,
  Antoine, Arnaud, Arun, and Manan}]{victor2022multitask}
\bibinfo{author}{S.~Victor}, \bibinfo{author}{W.~Albert},
  \bibinfo{author}{R.~Colin}, \bibinfo{author}{B.~Stephen},
  \bibinfo{author}{S.~Lintang}, \bibinfo{author}{A.~Zaid},
  \bibinfo{author}{C.~Antoine}, \bibinfo{author}{S.~Arnaud},
  \bibinfo{author}{R.~Arun}, \bibinfo{author}{D.~e.~a. Manan},
\newblock \bibinfo{title}{Multitask prompted training enables zero-shot task
  generalization},
\newblock in: \bibinfo{booktitle}{Proceedings of the International Conference
  on Learning Representations (ICLR)}, \bibinfo{year}{2022}.
%Type = Article
\bibitem[{Chung et~al.(2022)Chung, Hou, Longpre, Zoph, Tay, Fedus, Li, Wang,
  Dehghani, and Brahma}]{chung2022scaling}
\bibinfo{author}{H.~W. Chung}, \bibinfo{author}{L.~Hou},
  \bibinfo{author}{S.~Longpre}, \bibinfo{author}{B.~Zoph},
  \bibinfo{author}{Y.~Tay}, \bibinfo{author}{W.~Fedus},
  \bibinfo{author}{E.~Li}, \bibinfo{author}{X.~Wang},
  \bibinfo{author}{M.~Dehghani}, \bibinfo{author}{S.~e.~a. Brahma},
\newblock \bibinfo{title}{Scaling instruction-finetuned language models},
\newblock \bibinfo{journal}{arXiv preprint arXiv:2210.11416}
  (\bibinfo{year}{2022}).
%Type = Article
\bibitem[{Goyal et~al.(2022)Goyal, Li, and Durrett}]{goyal2022news}
\bibinfo{author}{T.~Goyal}, \bibinfo{author}{J.~J. Li},
  \bibinfo{author}{G.~Durrett},
\newblock \bibinfo{title}{News summarization and evaluation in the era of
  {GPT}-3},
\newblock \bibinfo{journal}{arXiv preprint arXiv:2209.12356}
  (\bibinfo{year}{2022}).
%Type = Article
\bibitem[{Köksal et~al.(2023)Köksal, Schick, Korhonen, and
  Schütze}]{Longform}
\bibinfo{author}{A.~Köksal}, \bibinfo{author}{T.~Schick},
  \bibinfo{author}{A.~Korhonen}, \bibinfo{author}{H.~Schütze},
\newblock \bibinfo{title}{{LongForm}: Optimizing instruction tuning for long
  text generation with corpus extraction},
\newblock \bibinfo{journal}{arXiv preprint 2304.08460}  (\bibinfo{year}{2023}).
%Type = Inproceedings
\bibitem[{Liu and Liu(2021)}]{Liu2021SimCLS}
\bibinfo{author}{Y.~Liu}, \bibinfo{author}{P.~Liu},
\newblock \bibinfo{title}{{SimCLS}: A simple framework for contrastive learning
  of abstractive summarization},
\newblock in: \bibinfo{booktitle}{Proceedings of the Annual Meeting of the
  Association for Computational Linguistics and the International Joint
  Conference on Natural Language Processing}, \bibinfo{year}{2021}.
%Type = Inproceedings
\bibitem[{Ladhak et~al.(2022)Ladhak, Durmus, He, Cardie, and
  McKeown}]{Ladhak2022}
\bibinfo{author}{F.~Ladhak}, \bibinfo{author}{E.~Durmus},
  \bibinfo{author}{H.~He}, \bibinfo{author}{C.~Cardie},
  \bibinfo{author}{K.~McKeown},
\newblock \bibinfo{title}{Faithful or extractive? {O}n mitigating the
  faithfulness-abstractiveness trade-off in abstractive summarization},
\newblock in: \bibinfo{booktitle}{Proceedings of the Annual Meeting of the
  Association for Computational Linguistics}, \bibinfo{year}{2022}.
%Type = Inproceedings
\bibitem[{Nan et~al.(2021)Nan, Nallapati, Wang, Santos, Zhu, Zhang, McKeown,
  and Xiang}]{nan2021entity}
\bibinfo{author}{F.~Nan}, \bibinfo{author}{R.~Nallapati},
  \bibinfo{author}{Z.~Wang}, \bibinfo{author}{C.~N. Santos},
  \bibinfo{author}{H.~Zhu}, \bibinfo{author}{D.~Zhang},
  \bibinfo{author}{K.~McKeown}, \bibinfo{author}{B.~Xiang},
\newblock \bibinfo{title}{Entity-level factual consistency of abstractive text
  summarization},
\newblock in: \bibinfo{booktitle}{Proceedings of the Conference of the European
  Chapter of the Association for Computational Linguistics},
  \bibinfo{year}{2021}.
%Type = Inproceedings
\bibitem[{Goyal and Durrett(2021)}]{goyal2021annotating}
\bibinfo{author}{T.~Goyal}, \bibinfo{author}{G.~Durrett},
\newblock \bibinfo{title}{Annotating and modeling fine-grained factuality in
  summarization},
\newblock in: \bibinfo{booktitle}{Proceedings of the Conference of the North
  American Chapter of the Association for Computational Linguistics: Human
  Language Technologies}, \bibinfo{year}{2021}.
%Type = Inproceedings
\bibitem[{Wan and Bansal(2022)}]{Wan2022Fact}
\bibinfo{author}{D.~Wan}, \bibinfo{author}{M.~Bansal},
\newblock \bibinfo{title}{{FactTPEGASUS}: Factuality-aware pre-training and
  fine-tuning for abstractive summarization},
\newblock in: \bibinfo{booktitle}{Proceedings of the Conference of the North
  American Chapter of the Association for Computational Linguistics: Human
  Language Technologies}, \bibinfo{year}{2022}.
%Type = Inproceedings
\bibitem[{Chen et~al.(2020)Chen, Kornblith, Norouzi, and
  Hinton}]{Chen2020simple}
\bibinfo{author}{T.~Chen}, \bibinfo{author}{S.~Kornblith},
  \bibinfo{author}{M.~Norouzi}, \bibinfo{author}{G.~Hinton},
\newblock \bibinfo{title}{A simple framework for contrastive learning of visual
  representations},
\newblock in: \bibinfo{booktitle}{Proceedings of the International Conference
  on Machine Learning (ICML)}, \bibinfo{organization}{PMLR},
  \bibinfo{year}{2020}.
%Type = Inproceedings
\bibitem[{Zhang et~al.(2022)Zhang, Yavuz, Kryscinski, Hashimoto, and
  Zhou}]{zhang2022improving}
\bibinfo{author}{H.~Zhang}, \bibinfo{author}{S.~Yavuz},
  \bibinfo{author}{W.~Kryscinski}, \bibinfo{author}{K.~Hashimoto},
  \bibinfo{author}{Y.~Zhou},
\newblock \bibinfo{title}{Improving the faithfulness of abstractive
  summarization via entity coverage control},
\newblock in: \bibinfo{booktitle}{Findings of the Association for Computational
  Linguistics}, \bibinfo{year}{2022}.
%Type = Inproceedings
\bibitem[{Xiao and Carenini(2022)}]{Xiao2022entity}
\bibinfo{author}{W.~Xiao}, \bibinfo{author}{G.~Carenini},
\newblock \bibinfo{title}{Entity-based {SpanCopy} for abstractive summarization
  to improve the factual consistency},
\newblock in: \bibinfo{booktitle}{Proceedings of the Workshop on Computational
  Approaches to Discourse (CODI)}, \bibinfo{year}{2022}.
%Type = Inproceedings
\bibitem[{Holtzman et~al.(2019)Holtzman, Buys, Du, Forbes, and
  Choi}]{holtzman2019curious}
\bibinfo{author}{A.~Holtzman}, \bibinfo{author}{J.~Buys},
  \bibinfo{author}{L.~Du}, \bibinfo{author}{M.~Forbes},
  \bibinfo{author}{Y.~Choi},
\newblock \bibinfo{title}{The curious case of neural text degeneration},
\newblock in: \bibinfo{booktitle}{Proceedings of the International Conference
  on Learning Representations (ICLR)}, \bibinfo{year}{2019}.
%Type = Article
\bibitem[{Koh et~al.(2022)Koh, Ju, Liu, and Pan}]{koh}
\bibinfo{author}{H.~Y. Koh}, \bibinfo{author}{J.~Ju}, \bibinfo{author}{M.~Liu},
  \bibinfo{author}{S.~Pan},
\newblock \bibinfo{title}{An empirical survey on long document summarization:
  Datasets, models, and metrics},
\newblock \bibinfo{journal}{ACM Computing Surveys} \bibinfo{volume}{55}
  (\bibinfo{year}{2022}) \bibinfo{pages}{1--35}.
%Type = Article
\bibitem[{Ahmad and Edalati(2022)}]{pakistan}
\bibinfo{author}{W.~Ahmad}, \bibinfo{author}{M.~Edalati},
\newblock \bibinfo{title}{Urdu speech and text based sentiment analyzer},
\newblock \bibinfo{journal}{arXiv preprint arXiv:2207.09163}
  (\bibinfo{year}{2022}).
%Type = Article
\bibitem[{Barreto et~al.(2023)Barreto, Moura, Carvalho, Paes, and
  Plastino}]{sergio}
\bibinfo{author}{S.~Barreto}, \bibinfo{author}{R.~Moura},
  \bibinfo{author}{J.~Carvalho}, \bibinfo{author}{A.~Paes},
  \bibinfo{author}{A.~Plastino},
\newblock \bibinfo{title}{Sentiment analysis in tweets: an assessment study
  from classical to modern word representation models},
\newblock \bibinfo{journal}{Data Mining and Knowledge Discovery}
  \bibinfo{volume}{37} (\bibinfo{year}{2023}) \bibinfo{pages}{318--380}.
%Type = Article
\bibitem[{Plutchik(2001)}]{Plutchik2001}
\bibinfo{author}{R.~Plutchik},
\newblock \bibinfo{title}{The nature of emotions: Human emotions have deep
  evolutionary roots, a fact that may explain their complexity and provide
  tools for clinical practice},
\newblock \bibinfo{journal}{American scientist} \bibinfo{volume}{89}
  (\bibinfo{year}{2001}) \bibinfo{pages}{344--350}.
%Type = Article
\bibitem[{Rambocas and Pacheco(2018)}]{sa_example_1}
\bibinfo{author}{M.~Rambocas}, \bibinfo{author}{B.~G. Pacheco},
\newblock \bibinfo{title}{Online sentiment analysis in marketing research: a
  review},
\newblock \bibinfo{journal}{Journal of Research in Interactive Marketing}
  (\bibinfo{year}{2018}).
%Type = Techreport
\bibitem[{Rambocas et~al.(2013)Rambocas, Gama et~al.}]{sa_example_2}
\bibinfo{author}{M.~Rambocas}, \bibinfo{author}{J.~Gama}, et~al.,
  \bibinfo{title}{Marketing research: The role of sentiment analysis},
  \bibinfo{type}{Technical Report}, Universidade do Porto, Faculdade de
  Economia do Porto, \bibinfo{year}{2013}.
%Type = Article
\bibitem[{Micu et~al.(2017)Micu, Micu, Geru, and Lixandroiu}]{sa_example_3}
\bibinfo{author}{A.~Micu}, \bibinfo{author}{A.~E. Micu},
  \bibinfo{author}{M.~Geru}, \bibinfo{author}{R.~C. Lixandroiu},
\newblock \bibinfo{title}{Analyzing user sentiment in social media:
  Implications for online marketing strategy},
\newblock \bibinfo{journal}{Psychology \& Marketing} \bibinfo{volume}{34}
  (\bibinfo{year}{2017}) \bibinfo{pages}{1094--1100}.
%Type = Inproceedings
\bibitem[{Ramteke et~al.(2016)Ramteke, Shah, Godhia, and Shaikh}]{sa_example_4}
\bibinfo{author}{J.~Ramteke}, \bibinfo{author}{S.~Shah},
  \bibinfo{author}{D.~Godhia}, \bibinfo{author}{A.~Shaikh},
\newblock \bibinfo{title}{Election result prediction using {Twitter} sentiment
  analysis},
\newblock in: \bibinfo{booktitle}{Proceedings of the International Conference
  on Inventive Computation Technologies (ICICT)}, \bibinfo{year}{2016}.
%Type = Inproceedings
\bibitem[{Kaya et~al.(2012)Kaya, Fidan, and Toroslu}]{sa_example_5}
\bibinfo{author}{M.~Kaya}, \bibinfo{author}{G.~Fidan}, \bibinfo{author}{I.~H.
  Toroslu},
\newblock \bibinfo{title}{Sentiment analysis of {Turkish} political news},
\newblock in: \bibinfo{booktitle}{Proceedings of the IEEE/WIC/ACM International
  Conferences on Web Intelligence and Intelligent Agent Technology},
  \bibinfo{year}{2012}.
%Type = Article
\bibitem[{Karamouzas et~al.(2022)Karamouzas, Mademlis, and
  Pitas}]{karamouzas2022SNAM}
\bibinfo{author}{D.~Karamouzas}, \bibinfo{author}{I.~Mademlis},
  \bibinfo{author}{I.~Pitas},
\newblock \bibinfo{title}{Public opinion monitoring through collective semantic
  analysis of tweets},
\newblock \bibinfo{journal}{Social Network Analysis and Mining}
  \bibinfo{volume}{12} (\bibinfo{year}{2022}) \bibinfo{pages}{91}.
%Type = Article
\bibitem[{Drus and Khalid(2019)}]{sa_applications}
\bibinfo{author}{Z.~Drus}, \bibinfo{author}{H.~Khalid},
\newblock \bibinfo{title}{Sentiment analysis in social media and its
  application: Systematic literature review},
\newblock \bibinfo{journal}{Procedia Computer Science} \bibinfo{volume}{161}
  (\bibinfo{year}{2019}) \bibinfo{pages}{707--714}.
%Type = Article
\bibitem[{Mori et~al.(2022)Mori, Yamane, Mukuta, and Harada}]{omori}
\bibinfo{author}{Y.~Mori}, \bibinfo{author}{H.~Yamane},
  \bibinfo{author}{Y.~Mukuta}, \bibinfo{author}{T.~Harada},
\newblock \bibinfo{title}{Computational storytelling and emotions: A survey},
\newblock \bibinfo{journal}{arXiv} \bibinfo{volume}{arXiv:2003.01200}
  (\bibinfo{year}{2022}).
%Type = Article
\bibitem[{Mehndiratta et~al.(2017)Mehndiratta, Sachdeva, and Soni}]{shelly}
\bibinfo{author}{P.~Mehndiratta}, \bibinfo{author}{S.~Sachdeva},
  \bibinfo{author}{D.~Soni},
\newblock \bibinfo{title}{Detection of sarcasm in text data using deep
  convolutional neural networks},
\newblock \bibinfo{journal}{Scalable Computing: Practice and Experience}
  \bibinfo{volume}{18} (\bibinfo{year}{2017}) \bibinfo{pages}{219--228}.
%Type = Article
\bibitem[{Onan and Korukoglu(2017)}]{onan_2017_sa}
\bibinfo{author}{A.~Onan}, \bibinfo{author}{S.~Korukoglu},
\newblock \bibinfo{title}{A feature selection model based on genetic rank
  aggregation for text sentiment classification},
\newblock \bibinfo{journal}{Journal of Information Science}
  \bibinfo{volume}{43} (\bibinfo{year}{2017}) \bibinfo{pages}{25--38}.
%Type = Inproceedings
\bibitem[{Pang et~al.(2002)Pang, Lee, and Vaithyanathan}]{pang2}
\bibinfo{author}{B.~Pang}, \bibinfo{author}{L.~Lee},
  \bibinfo{author}{S.~Vaithyanathan},
\newblock \bibinfo{title}{Thumbs up? sentiment classification using machine
  learning techniques},
\newblock in: \bibinfo{booktitle}{Proceedings of the Conference on Empirical
  Methods in Natural Language Processing ({EMNLP})},
  \bibinfo{publisher}{Association for Computational Linguistics},
  \bibinfo{year}{2002}.
%Type = Inproceedings
\bibitem[{Pang and Lee(2004)}]{pang}
\bibinfo{author}{B.~Pang}, \bibinfo{author}{L.~Lee},
\newblock \bibinfo{title}{A sentimental education: Sentiment analysis using
  subjectivity summarization based on minimum cuts},
\newblock in: \bibinfo{booktitle}{Proceedings of the Annual Meeting on
  Association for Computational Linguistics {(ACL '04)}},
  \bibinfo{publisher}{Association for Computational Linguistics},
  \bibinfo{year}{2004}.
%Type = Article
\bibitem[{Onan(2022)}]{onan2022_sa}
\bibinfo{author}{A.~Onan},
\newblock \bibinfo{title}{Bidirectional convolutional {Recurrent Neural
  Network} architecture with group-wise enhancement mechanism for text
  sentiment classification},
\newblock \bibinfo{journal}{Journal of King Saud University - Computer and
  Information Sciences} \bibinfo{volume}{34} (\bibinfo{year}{2022})
  \bibinfo{pages}{2098--2117}.
%Type = Article
\bibitem[{Liu and Guo(2019)}]{liu_2019_attention}
\bibinfo{author}{G.~Liu}, \bibinfo{author}{J.~Guo},
\newblock \bibinfo{title}{Bidirectional {LSTM} with attention mechanism and
  convolutional layer for text classification},
\newblock \bibinfo{journal}{Neurocomputing} \bibinfo{volume}{337}
  (\bibinfo{year}{2019}) \bibinfo{pages}{325--338}.
%Type = Article
\bibitem[{Li et~al.(2017)Li, Peng, Yao, Cui, Hu, You, and Chi}]{li2017}
\bibinfo{author}{X.~Li}, \bibinfo{author}{L.~Peng}, \bibinfo{author}{X.~Yao},
  \bibinfo{author}{S.~Cui}, \bibinfo{author}{Y.~Hu}, \bibinfo{author}{C.~You},
  \bibinfo{author}{T.~Chi},
\newblock \bibinfo{title}{{Long Short-Term Memory} neural network for air
  pollutant concentration predictions: Method development and evaluation},
\newblock \bibinfo{journal}{Environmental Pollution} \bibinfo{volume}{231}
  (\bibinfo{year}{2017}) \bibinfo{pages}{997--1004}.
%Type = Inproceedings
\bibitem[{Onan(2019)}]{onan_2019_sarcasm}
\bibinfo{author}{A.~Onan},
\newblock \bibinfo{title}{Topic-enriched word embeddings for sarcasm
  identification},
\newblock in: \bibinfo{booktitle}{Proceedings of the Computer Science On-line
  Conference}, \bibinfo{year}{2019}.
%Type = Article
\bibitem[{Onan and Toçoğlu(2021)}]{onan_sarcasm_survey}
\bibinfo{author}{A.~Onan}, \bibinfo{author}{M.~A. Toçoğlu},
\newblock \bibinfo{title}{A term-weighted neural language model and stacked
  bidirectional {LSTM}-based framework for sarcasm identification},
\newblock \bibinfo{journal}{IEEE Access} \bibinfo{volume}{9}
  (\bibinfo{year}{2021}) \bibinfo{pages}{7701--7722}.
%Type = Article
\bibitem[{Onan(2021)}]{onan_2021}
\bibinfo{author}{A.~Onan},
\newblock \bibinfo{title}{Sentiment analysis on product reviews based on
  weighted word embeddings and {Deep Neural Networks}},
\newblock \bibinfo{journal}{Concurrency and Computation: Practice and
  Experience} \bibinfo{volume}{33} (\bibinfo{year}{2021})
  \bibinfo{pages}{e5909}.
%Type = Inproceedings
\bibitem[{Davidov et~al.(2010)Davidov, Tsur, and Rappoport}]{dmitry}
\bibinfo{author}{D.~Davidov}, \bibinfo{author}{O.~Tsur},
  \bibinfo{author}{A.~Rappoport},
\newblock \bibinfo{title}{Semi-supervised recognition of sarcasm in {Twitter
  and Amazon}},
\newblock in: \bibinfo{booktitle}{Proceedings of the Conference on
  Computational Natural Language Learning}, \bibinfo{year}{2010}.
%Type = Inproceedings
\bibitem[{Tang et~al.(2015)Tang, Qin, and Liu}]{tang}
\bibinfo{author}{D.~Tang}, \bibinfo{author}{B.~Qin}, \bibinfo{author}{T.~Liu},
\newblock \bibinfo{title}{Document modeling with {Gated Recurrent Neural
  Network} for sentiment classification},
\newblock in: \bibinfo{booktitle}{Proceedings of the Conference on Empirical
  Methods in Natural Language Processing}, \bibinfo{publisher}{Association for
  Computational Linguistics}, \bibinfo{year}{2015}.
%Type = Article
\bibitem[{Pelletier(1994)}]{francis}
\bibinfo{author}{F.~Pelletier},
\newblock \bibinfo{title}{The principle of semantic compositionality},
\newblock \bibinfo{journal}{Topoi} \bibinfo{volume}{13} (\bibinfo{year}{1994})
  \bibinfo{pages}{11--24}.
%Type = Article
\bibitem[{Rhanoui et~al.(2019)Rhanoui, Mikram, Yousfi, and Barzali}]{morocco}
\bibinfo{author}{M.~Rhanoui}, \bibinfo{author}{M.~Mikram},
  \bibinfo{author}{S.~Yousfi}, \bibinfo{author}{S.~Barzali},
\newblock \bibinfo{title}{A {CNN-BiLSTM} model for document-level sentiment
  analysis},
\newblock \bibinfo{journal}{Machine Learning and Knowledge Extraction}
  \bibinfo{volume}{1} (\bibinfo{year}{2019}) \bibinfo{pages}{832--847}.
%Type = Inproceedings
\bibitem[{Le and Mikolov(2014)}]{doc2vec}
\bibinfo{author}{Q.~Le}, \bibinfo{author}{T.~Mikolov},
\newblock \bibinfo{title}{Distributed representations of sentences and
  documents},
\newblock in: \bibinfo{booktitle}{Proceedings of the International Conference
  on Machine Learning}, \bibinfo{organization}{PMLR}, \bibinfo{year}{2014}.
%Type = Article
\bibitem[{Mao et~al.(2022)Mao, Zhang, Jiao, and Zhang}]{mao2022document}
\bibinfo{author}{Y.~Mao}, \bibinfo{author}{Y.~Zhang},
  \bibinfo{author}{L.~Jiao}, \bibinfo{author}{H.~Zhang},
\newblock \bibinfo{title}{Document-level sentiment analysis using
  attention-based bi-directional {Long Short-Term Memory} network and
  two-dimensional {Convolutional Neural Network}},
\newblock \bibinfo{journal}{Electronics} \bibinfo{volume}{11}
  (\bibinfo{year}{2022}) \bibinfo{pages}{1906}.
%Type = Article
\bibitem[{Bhuvaneshwari et~al.(2022)Bhuvaneshwari, Rao, Robinson, and
  Thippeswamy}]{bhuvaneshwari2022sentiment}
\bibinfo{author}{P.~Bhuvaneshwari}, \bibinfo{author}{A.~N. Rao},
  \bibinfo{author}{Y.~H. Robinson}, \bibinfo{author}{M.~Thippeswamy},
\newblock \bibinfo{title}{Sentiment analysis for user reviews using {Bi-LSTM
  self-attention-based CNN} model},
\newblock \bibinfo{journal}{Multimedia Tools and Applications}
  \bibinfo{volume}{81} (\bibinfo{year}{2022}) \bibinfo{pages}{12405--12419}.
%Type = Inproceedings
\bibitem[{Zhao et~al.(2020)Zhao, Li, Zheng, and Zhang}]{xinzhao}
\bibinfo{author}{L.~Zhao}, \bibinfo{author}{L.~Li}, \bibinfo{author}{X.~Zheng},
  \bibinfo{author}{J.~Zhang},
\newblock \bibinfo{title}{A {BERT}-based sentiment analysis and key entity
  detection approach for on-line financial texts},
\newblock in: \bibinfo{booktitle}{Proceedings of the IEEE International
  Conference on Computer Supported Cooperative Work in Design (CSCWD)},
  \bibinfo{year}{2020}.
%Type = Inproceedings
\bibitem[{Nallapati et~al.(2016)Nallapati, Zhou, dos Santos, Gulcehre, and
  Xiang}]{nallapati2}
\bibinfo{author}{R.~Nallapati}, \bibinfo{author}{B.~Zhou},
  \bibinfo{author}{C.~dos Santos}, \bibinfo{author}{C.~Gulcehre},
  \bibinfo{author}{B.~Xiang},
\newblock \bibinfo{title}{Abstractive text summarization using
  sequence-to-sequence {RNN}s and beyond},
\newblock in: \bibinfo{booktitle}{Proceedings of the {SIGNLL} Conference on
  Computational Natural Language Learning}, \bibinfo{year}{2016}.
%Type = Misc
\bibitem[{tf_(date)}]{tf_cnn_dailymail}
\bibinfo{title}{{TensorFlow CNN-DailyMail Dataset}},
  \bibinfo{howpublished}{\url{https://www.tensorflow.org/datasets/catalog/cnn_dailymail}},
  \bibinfo{year}{nodate}. \bibinfo{note}{Accessed: 2023-05-30}.
%Type = Misc
\bibitem[{kag(date)}]{kaggle_cnn_dailymail}
\bibinfo{title}{{Kaggle CNN-DailyMail Dataset}},
  \bibinfo{howpublished}{\url{https://www.kaggle.com/datasets/gowrishankarp/newspaper-text-summarization-cnn-dailymail}},
  \bibinfo{year}{nodate}. \bibinfo{note}{Accessed: 2023-05-30}.
%Type = Article
\bibitem[{Gong et~al.(2016)Gong, Chen, Qiu, and Huang}]{gong}
\bibinfo{author}{J.~Gong}, \bibinfo{author}{X.~Chen}, \bibinfo{author}{X.~Qiu},
  \bibinfo{author}{X.~Huang},
\newblock \bibinfo{title}{End-to-end neural sentence ordering using pointer
  network},
\newblock \bibinfo{journal}{arXiv} \bibinfo{volume}{arXiv:1611.04953}
  (\bibinfo{year}{2016}).
%Type = Misc
\bibitem[{tf-datasets(date)}]{tf_datasets}
tf-datasets, \bibinfo{title}{{TensorFlow} scientific papers dataset},
  \bibinfo{howpublished}{\url{https://www.tensorflow.org/datasets/catalog/scientific_papers}},
  \bibinfo{year}{nodate}. \bibinfo{note}{Accessed: 2023-04-22}.
%Type = Misc
\bibitem[{Bozsolik(2020)}]{kaggle_arxiv}
\bibinfo{author}{T.~Bozsolik}, \bibinfo{title}{{Kaggle arXiv} dataset},
  \bibinfo{howpublished}{\url{https://www.tensorflow.org/datasets/catalog/scientific_papers}},
  \bibinfo{year}{2020}. \bibinfo{note}{Accessed: 2023-04-22}.
%Type = Misc
\bibitem[{arxiv-hugging-face(date)}]{hugging_face_arxiv}
arxiv-hugging-face, \bibinfo{title}{{Hugging Face library arXiv} dataset},
  \bibinfo{howpublished}{\url{https://www.tensorflow.org/datasets/catalog/scientific_papers}},
  \bibinfo{year}{nodate}. \bibinfo{note}{Accessed: 2023-04-22}.
%Type = Misc
\bibitem[{pubmed-hugging-face(date)}]{hugging_face_pubmed}
pubmed-hugging-face, \bibinfo{title}{{Hugging Face library arXiv} dataset},
  \bibinfo{howpublished}{\url{https://huggingface.co/datasets/pubmed}},
  \bibinfo{year}{nodate}. \bibinfo{note}{Accessed: 2023-04-22}.
%Type = Article
\bibitem[{Vishnubhotla et~al.(2022)Vishnubhotla, Hammond, and
  Hirst}]{vishnubhotla2022project}
\bibinfo{author}{K.~Vishnubhotla}, \bibinfo{author}{A.~Hammond},
  \bibinfo{author}{G.~Hirst},
\newblock \bibinfo{title}{{The Project Dialogism Novel Corpus: A dataset for
  quotation attribution in literary texts}},
\newblock \bibinfo{journal}{arXiv preprint arXiv:2204.05836}
  (\bibinfo{year}{2022}).
%Type = Inproceedings
\bibitem[{Bamman et~al.(2020)Bamman, Lewke, and Mansoor}]{bamman}
\bibinfo{author}{D.~Bamman}, \bibinfo{author}{O.~Lewke},
  \bibinfo{author}{A.~Mansoor},
\newblock \bibinfo{title}{An annotated dataset of coreference in {E}nglish
  literature},
\newblock in: \bibinfo{booktitle}{Proceedings of the International Conference
  on Language Resources and Evaluation ({LREC})}, \bibinfo{publisher}{European
  Language Resources Association}, \bibinfo{year}{2020}.
%Type = Inproceedings
\bibitem[{Chen et~al.(2021)Chen, Chu, Wiseman, and Gimpel}]{summset}
\bibinfo{author}{M.~Chen}, \bibinfo{author}{Z.~Chu},
  \bibinfo{author}{S.~Wiseman}, \bibinfo{author}{K.~Gimpel},
\newblock \bibinfo{title}{{SummScreen}: {A} dataset for abstractive screenplay
  summarization},
\newblock \bibinfo{year}{2021}.
%Type = Inproceedings
\bibitem[{Tiedemann(2012)}]{tiedemann}
\bibinfo{author}{J.~Tiedemann},
\newblock \bibinfo{title}{Parallel data, tools and interfaces in {OPUS}},
\newblock in: \bibinfo{booktitle}{Proceedings of the International Conference
  on Language Resources and Evaluation ({LREC})}, \bibinfo{publisher}{European
  Language Resources Association (ELRA)}, \bibinfo{year}{2012}.
%Type = Inproceedings
\bibitem[{Koreeda and Manning(2021)}]{koreeda}
\bibinfo{author}{Y.~Koreeda}, \bibinfo{author}{C.~Manning},
\newblock \bibinfo{title}{{C}ontract{NLI}: A dataset for document-level natural
  language inference for contracts},
\newblock in: \bibinfo{booktitle}{Proceedings of the Findings of the
  Association for Computational Linguistics (EMNLP)}, \bibinfo{year}{2021}.
%Type = Article
\bibitem[{Clement et~al.(2019)Clement, Bierbaum, O'Keeffe, and
  Alemi}]{arxiv_size}
\bibinfo{author}{C.~B. Clement}, \bibinfo{author}{M.~Bierbaum},
  \bibinfo{author}{K.~P. O'Keeffe}, \bibinfo{author}{A.~A. Alemi},
\newblock \bibinfo{title}{On the use of {arXiv} as a dataset},
\newblock \bibinfo{journal}{arXiv preprint arXiv:1905.00075}
  (\bibinfo{year}{2019}).
%Type = Inproceedings
\bibitem[{Wagh et~al.(2021)Wagh, Khandve, Joshi, Wani, Kale, and Joshi}]{wagh}
\bibinfo{author}{V.~Wagh}, \bibinfo{author}{S.~Khandve},
  \bibinfo{author}{I.~Joshi}, \bibinfo{author}{A.~Wani},
  \bibinfo{author}{G.~Kale}, \bibinfo{author}{R.~Joshi},
\newblock \bibinfo{title}{Comparative study of long document classification},
\newblock in: \bibinfo{booktitle}{Proceedings of the IEEE Region 10 Conference
  (TENCON)}, \bibinfo{year}{2021}.
%Type = Article
\bibitem[{Fields et~al.(2024)Fields, Chovanec, and Madiraju}]{Fields2024}
\bibinfo{author}{J.~Fields}, \bibinfo{author}{K.~Chovanec},
  \bibinfo{author}{P.~Madiraju},
\newblock \bibinfo{title}{A survey of text classification with transformers:
  How wide? how large? how long? how accurate? how expensive? how safe?},
\newblock \bibinfo{journal}{IEEE Access}  (\bibinfo{year}{2024}).
%Type = Article
\bibitem[{Uppalapati et~al.(2023)Uppalapati, Dabbiru, and Rao}]{Uppalapati2023}
\bibinfo{author}{P.~J. Uppalapati}, \bibinfo{author}{M.~Dabbiru},
  \bibinfo{author}{K.~V. Rao},
\newblock \bibinfo{title}{A comprehensive survey on summarization techniques},
\newblock \bibinfo{journal}{SN Computer Science} \bibinfo{volume}{4}
  (\bibinfo{year}{2023}) \bibinfo{pages}{560}.
%Type = Article
\bibitem[{Tay et~al.(2022)Tay, Dehghani, Bahri, and Metzler}]{Tay2022survey}
\bibinfo{author}{Y.~Tay}, \bibinfo{author}{M.~Dehghani},
  \bibinfo{author}{D.~Bahri}, \bibinfo{author}{D.~Metzler},
\newblock \bibinfo{title}{Efficient {Transformers}: A survey},
\newblock \bibinfo{journal}{ACM Computing Surveys} \bibinfo{volume}{55}
  (\bibinfo{year}{2022}).
%Type = Article
\bibitem[{Fournier et~al.(2023)Fournier, Caron, and Aloise}]{Fournier2023}
\bibinfo{author}{Q.~Fournier}, \bibinfo{author}{G.~M. Caron},
  \bibinfo{author}{D.~Aloise},
\newblock \bibinfo{title}{A practical survey on faster and lighter
  {Transformers}},
\newblock \bibinfo{journal}{ACM Computing Surveys} \bibinfo{volume}{55}
  (\bibinfo{year}{2023}) \bibinfo{pages}{1--40}.
%Type = Inproceedings
\bibitem[{Elouargui et~al.(2023)Elouargui, Zyate, Sassioui, Chergui, El~Kamili,
  and Ouzzif}]{Elouargui2023}
\bibinfo{author}{Y.~Elouargui}, \bibinfo{author}{M.~Zyate},
  \bibinfo{author}{A.~Sassioui}, \bibinfo{author}{M.~Chergui},
  \bibinfo{author}{M.~El~Kamili}, \bibinfo{author}{M.~Ouzzif},
\newblock \bibinfo{title}{A comprehensive survey on efficient transformers},
\newblock in: \bibinfo{booktitle}{Proceedings of the IEEE International
  Conference on Wireless Networks and Mobile Communications (WINCOM)},
  \bibinfo{year}{2023}.
%Type = Article
\bibitem[{Treviso et~al.(2023)Treviso, Lee, Ji, Aken, Cao, Ciosici, Hassid,
  Heafield, Hooker, Raffel et~al.}]{Treviso2023}
\bibinfo{author}{M.~Treviso}, \bibinfo{author}{J.-U. Lee},
  \bibinfo{author}{T.~Ji}, \bibinfo{author}{B.~Aken}, \bibinfo{author}{Q.~Cao},
  \bibinfo{author}{M.~R. Ciosici}, \bibinfo{author}{M.~Hassid},
  \bibinfo{author}{K.~Heafield}, \bibinfo{author}{S.~Hooker},
  \bibinfo{author}{C.~Raffel}, et~al.,
\newblock \bibinfo{title}{Efficient methods for {Natural Language Processing}:
  A survey},
\newblock \bibinfo{journal}{Transactions of the Association for Computational
  Linguistics} \bibinfo{volume}{11} (\bibinfo{year}{2023})
  \bibinfo{pages}{826--860}.

\end{thebibliography}
\newpage
\section*{Appendix}
\textbf{Relevant Previous Review/Survey Articles}\\
\begin{table}[H]
        \centering
         \begin{tabular}
             { |p{4cm}|p{0.6cm}|p{0.6cm}|p{0.6cm}|p{0.6cm}|p{0.6cm}|p{0.6cm}|p{0.6cm}|p{0.6cm}|p{0.6cm}|p{2cm}|}
             \hline
             \textbf{Review}&\cite{wagh}&\cite{koh}&\cite{omori}&\cite{Fields2024}&\cite{Uppalapati2023}&\cite{Tay2022survey}&\cite{Fournier2023}&\cite{Elouargui2023}&\cite{Treviso2023}&This article\\
             \hline
             Document Classification&\checkmark&\xmark&\xmark&\checkmark&\xmark&\xmark&\xmark&\checkmark&\xmark&\checkmark\\
             \hline
             Document Summarization&\xmark&\checkmark&\xmark&\xmark&\checkmark&\xmark&\xmark&\xmark&\xmark&\checkmark\\
             \hline
             Sentiment Analysis&\xmark&\xmark&\checkmark&\checkmark&\xmark&\xmark&\xmark&\xmark&\xmark&\checkmark\\
             \hline
             Relevant Neural Networks&\xmark&\xmark&\xmark&\xmark&\xmark&\checkmark&\checkmark&\checkmark&\checkmark&\checkmark\\
             \hline
             Issue Discussion&\checkmark&\checkmark&\xmark&\xmark&\xmark&\xmark&\xmark&\xmark&\xmark&\checkmark\\
             \hline
             Issue Aggregation&\xmark&\xmark&\xmark&\xmark&\xmark&\xmark&\xmark&\xmark&\xmark&\checkmark\\
             \hline
             Long Document Datasets&\xmark&\checkmark&\xmark&\xmark&\checkmark&\xmark&\xmark&\xmark&\xmark&\checkmark\\
             \hline
             LLMs&\xmark&\xmark&\xmark&\checkmark&\xmark&\xmark&\xmark&\xmark&\checkmark&\xmark\\
             \hline
        \end{tabular}
    \caption{Comparisons between this article and notable recent review/survey publications pertaining to long document classification or summarization.}
    \label{tab::review_table}
\end{table}

Despite the existence of many overview/survey articles pertaining to NLP or relevant neural architectures (e.g., efficient Transformers), there is a distinct scarcity of recent such publications that specifically concern long document analysis and overlap with this article (i.e., focus on classification or summarization). This section briefly presents a selection of existing overlapping surveys and highlights the points of differentiation from this article.

In \cite{wagh}, the authors present an experimental comparative evaluation of traditional text classification methods and deep neural approaches, in the context of long document classification. However, they do not cover recent neural architectures such as efficient Transformers, public long document datasets, document summarization or sentiment analysis. Thus, they do not synthesize their findings across different NLP tasks and their article is removed from today's SoA. In \cite{koh}, a thorough survey of SoA models, datasets and metrics for document summarization is offered, which does cover a number of early efficient Transformer architectures and systematically extracts insights regarding open issues and potential solutions. However, it does not correlate its findings with long document classification and does not elaborate on how the presented neural architectures operate, while its coverage is slightly outdated as of 2024. The survey in \cite{Uppalapati2023} is more up-to-date regarding long document summarization while also presenting relevant metrics and datasets in brief, but otherwise it also does not extend to document classification and does not detail the employed neural architectures.

Sentiment analysis, also known as \textit{opinion mining}, typically concerns short documents, such as reviews, comments and social media posts. However, a review of modern approaches to a specific subtask of sentiment analysis is conducted in \cite{omori}, where the goal is to analyze entire literary works with respect to their story and emotions. Due to its particular scope, at the intersection of NLP and literary analysis with a strong emphasis on emotion recognition, this survey does not cover generic document classification or document summarization, does not elaborate on the operation of the discussed neural architectures and does not present public long document datasets. In contrast, \cite{Fields2024} surveys text classification in general across multiple datasets and classification subtasks (such as sentiment analysis), including in long document datasets, focusing on recent Transformer architectures. However, it does not specialize to the peculiarities of long texts, it does not cover document summarization and does not present the discussed neural architectures in-depth, as its focus is on identifying the currently optimal method per dataset. Finally, \cite{Fields2024} focuses on Large Language Models (LLMs) in contrast to this article that explicitly does not, emphasizing instead task-specific efficient Transformer architectures.

A number of recent surveys/reviews focus on efficient Transformer neural architectures, regardless of the application domain, and thus are overlapping with this article from a different perspective, given that such DNNs can better analyze lengthy input sequences. For instance, the survey in \cite{Tay2022survey} covers efficient Transformers extensively and in-depth, but does not specialize in any long document analysis task and, therefore, does not present relevant datasets or domain-specific issues. The situation is similar for \cite{Fournier2023} and \cite{Elouargui2023}, although the latter's coverage of efficient Transformers is much less extensive. Moreover, \cite{Elouargui2023} does present benchmark datasets for lengthy input sequences, but only minimally touches upon document classification. Finally, significant overlap exists with the neural architectures covered in \cite{Treviso2023}, but that survey's emphasis is on efficient Transformers for NLP with limited computational resources (not on long document analysis).

While all of the above surveys/reviews overlap with this article, there is no previous effort for aggregating and comparing the issues and challenges specifically posed by long documents across both classification and summarization, while focusing exclusively on modern DNNs. This article attempts to expose common issues and encourage the sharing of tried-and-tested solutions across tasks. Additionally, due to the in-depth coverage of the presented neural architectures, it possesses a tutorial value which is lacking from the majority of related published surveys. In summary, a simplified comparison between this article and notable recent review publications pertaining to long document classification or summarization can be found in Table \ref{tab::review_table}. The row ``Issue Discussion" indicates that the corresponding paper specifically discusses open issues with long document analysis, while the row ``Issue Aggregation" indicates that issues are aggregated and compared across different NLP tasks. Only officially published articles have been included in this comparison, i.e., no arXiv-uploaded preprints.
\end{document}